%% file: IJCV.tex
\documentclass[10pt,twocolumn,letterpaper]{article}

\usepackage{graphicx}
\usepackage{color}
\usepackage{amssymb}
\usepackage{amsmath}

\usepackage{booktabs}
\usepackage{tabulary}
\usepackage{array, multirow}
\usepackage{comment}
\usepackage{rotating}

\usepackage{cite}
\usepackage{url}

\usepackage[utf8]{inputenc}
\usepackage[british]{babel}
\usepackage[hyphenbreaks]{breakurl}
\usepackage{lipsum}
\usepackage{multirow}
\usepackage[table,xcdraw]{xcolor}
\usepackage{enumitem}
\setlist{topsep=0pt, leftmargin=*}

\newcommand*{\cc}[1]{\textcolor{black}{#1}}
\newcommand*{\ccc}[1]{\textcolor{black}{#1}}
\newcommand*{\nc}[1]{\textcolor{black}{#1}}
\newcommand*{\crv}[1]{\textcolor{black}{#1}}

\newcommand{\task}{\textcolor{black}{{paradigm}}}
\newcommand{\tasks}{\textcolor{black}{{paradigms}}}
\newcommand{\Task}{\textcolor{black}{{Paradigm}}}
\newcommand{\Tasks}{\textcolor{black}{{Paradigms }}}

\newcommand{\widthfivefigs}{0.19}

\usepackage{xspace}

\newcommand*{\eg}{\emph{e.g.}\@\xspace}
\newcommand*{\ie}{\emph{i.e.}\@\xspace}

\newcommand{\PAR}[1]{\vskip4pt \noindent{\bf #1.}}
\newcommand{\PARR}[1]{\vskip4pt \noindent{\bf #1}}

\newcommand{\degree}{$^{\circ}$}

\newcommand{\myparagraph}[1]{\PAR{#1}}

\begin{document}

\title{\textbf{Investigating the Role of Image Retrieval for Visual Localization - An exhaustive benchmark}\thanks{This preprint has not undergone peer review (when applicable) or any post-submission improvements or corrections. The Version of Record of this article is published in the International Journal of Computer Vision (2022), and is available online at https://doi.org/10.1007/s11263-022-01615-7.}}

\author{Martin Humenberger$^1$ \quad
        Yohann Cabon$^1$ \quad 
        Noé Pion$^4$ \quad
        Philippe Weinzaepfel$^1$\\ \quad
        Donghwan Lee$^2$ \quad
        Nicolas Guérin$^1$ \quad
        Torsten Sattler$^3$ \quad
        Gabriela Csurka$^1$\\ \\
        $^1$\small{NAVER LABS Europe, France} \\
        $^2$\small{NAVER LABS, Republic of Korea} \\
        $^3$\small{Czech Institute of Informatics, Robotics and Cybernetics, Czech Technical University in Prague} \\
        $^4$\small{The work has been done during an appointment at NAVER LABS Europe.}}

\date{}

\maketitle

\begin{abstract}
\emph{Visual localization, \ie, camera pose estimation in a known scene, is a core component of technologies such as autonomous driving and augmented reality. State-of-the-art localization approaches often rely on image retrieval techniques for one of two \crv{purposes:} (1) provide an approximate pose estimate or (2) determine which parts of the scene are potentially visible in a given query image. It is common practice to use state-of-the-art image retrieval algorithms for \crv{both of them}. These algorithms are often trained for the goal of retrieving the same landmark under a large range of viewpoint changes which often differs from the requirements of visual localization. \crv{In order to investigate the consequences for visual localization}, this paper focuses on understanding the role of image retrieval for multiple visual localization \tasks. First, we introduce a novel benchmark setup and compare state-of-the-art retrieval representations on multiple datasets using localization performance as metric. Second, we investigate several definitions of ``ground truth" for image retrieval. Using these definitions as upper bounds for the visual localization \tasks, we show that there is still significant room for improvement. Third, using these tools and in-depth analysis, we show that retrieval performance on classical landmark retrieval or place recognition tasks correlates only for some but not all \tasks~to localization performance. Finally, we analyze the effects of blur and dynamic scenes in the images. We conclude that there is a need for retrieval approaches specifically designed for localization \tasks.
Our benchmark and evaluation protocols are available at} \textbf{https://github.com/naver/kapture-localization}.
\end{abstract}



\section{Introduction}
\label{sec:introduction}

Visual localization is the problem of estimating the exact camera pose for a given image in a known scene, \ie, the exact position and orientation from which the image was taken. 
Localization algorithms are core components of systems such as self-driving cars~\cite{HengICRA19ProjectAutoVisionLocalization3DAutonomousVehicle}, autonomous robots~\cite{LimIJRR15RealTimeMonocularImageBased6DoFLocalization}, and mixed reality applications~\cite{ArthISMAR09WideAreaLocalizationMobilePhones,MiddelbergECCV14Scalable6DOFLocalization,VenturaTVCG1Gl4obalLocalizationMonocularSLAM,LynenRSSC15GetOutVisualInertialLocalization,CastleISWC08VideoRateLocalizationMultipleMapsAR}. 

\begin{figure*}[th!]
    \centering
    \includegraphics[width=\textwidth]{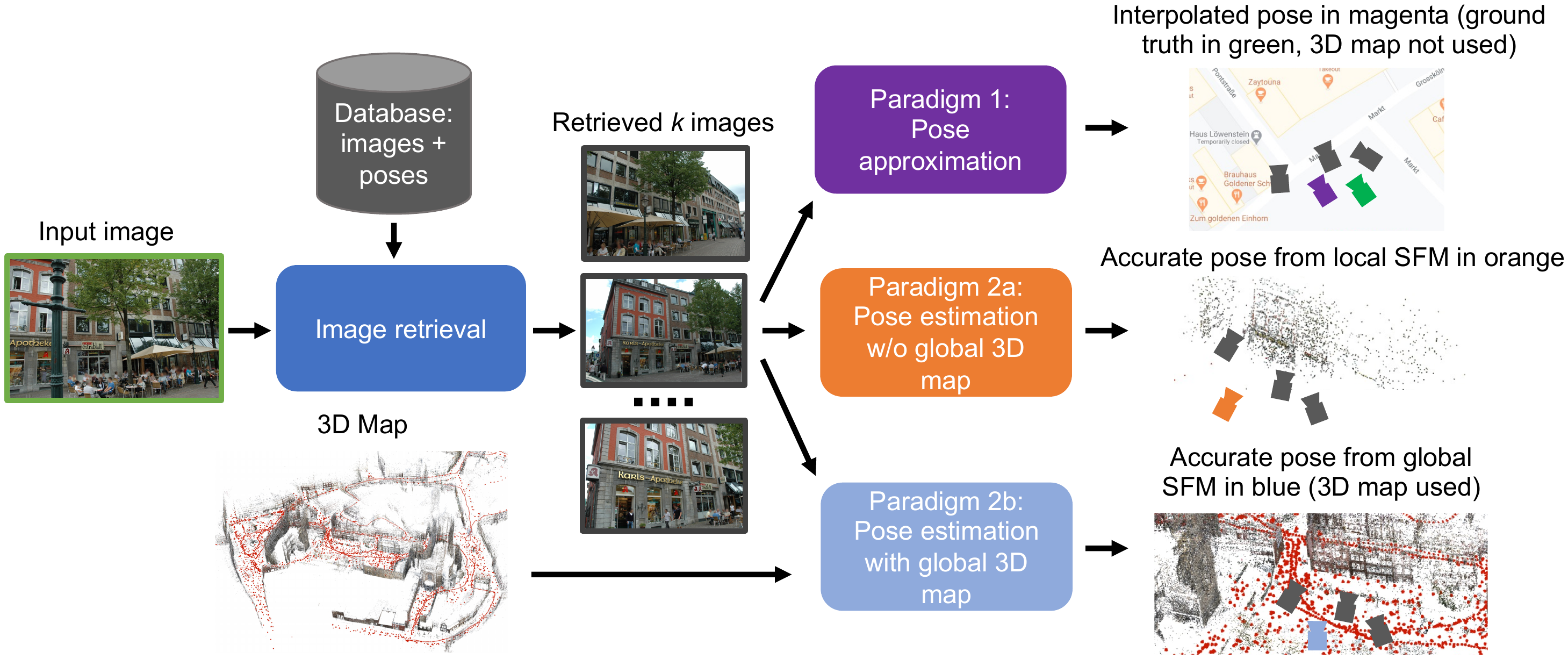}
     \caption{This paper analyzes the role of image retrieval in three visual localization \tasks~through extensive experiments.}
    \label{fig:overview}
\end{figure*}

Traditionally, visual localization algorithms rely on a 3D scene representation of the target area~\cite{SeIROS02GlobalLocalizationDistinctiveVisualFeatures,LiECCV12WorldwidePoseEst,IrscharaCVPR09FromSFMLocationRecognition,LiECCV10LocationRecPriorFeatureMatching,SattlerPAMI17EfficientPrioritizedMatching}, constructed from reference or database images with known camera poses (further referred to as poses). 
They use 2D-3D matches between a query image and the 3D representation for pose estimation. 
This representation can be an explicit 3D model, often obtained via Structure-from-Motion (SFM)~\cite{SchonbergerCVPR16StructureFromMotionRevisited,SnavelyIJCV08ModelingTheWorldFromInternetPhotoCollections,HeinlyCVPR15ReconstructingTheWorldSixDays} using local features for 2D-3D matching, or an implicit representation through a machine learning algorithm~\cite{MassicetiICRA17RandomForestVersusNNCamLoc,BrachmannCVPR18LearningLessIsMore6DLocalization,ShottonCVPR13SceneCoordinateRegression}. 
In the latter case, the learning algorithm is trained to regress 2D-3D matches.
These structure-based methods can be scaled to large scenes through an intermediate image retrieval step~\cite{TairaPAMI19InLocIndoorVisualLocalization,SarlinCVPR19FromCoarsetoFineHierarchicalLocalization,SattlerBMVC12ImRetLocalizationRevisited,SattlerICCV15HyperpointsFineVocabulariesLocRecogn,Germain3DV19SparseToDenseHypercolumnMatchingVisLoc,BrachmannICCV19ExpertSampleConsensusReLocalization,Cui3DV17GraphMatchEfficientGraphSFM}.
In image retrieval, global image representations (global features) are used to compute similarity between a query image and the database images which can then be used for ranking.
The intuition is that the top retrieved images provide hypotheses about which parts of the scene are likely visible in a query image. 2D-3D matching can then be restricted to these parts. 

The pre-processing step of building a 3D scene representation is not strictly necessary. 
Instead, the camera pose of a query image can directly be computed using the known poses of the top database images found, again using image retrieval. 
This can be achieved via relative pose estimation between query and retrieved images~\cite{ZhouICRA20ToLearnLocalizationFromEssentialMatrices,ZhangS3DPVT06ImageBasedLocUrbanEnvironments}, by estimating the absolute pose from 2D-2D matches~\cite{ZhengICCV15StructureFromMotionStructureLessResection}, via relative pose regression~\cite{BalntasECCV18RelocNetMetricLearningRelocalisation,DingICCV19CamNetRetrievalForReLocalization} or by building local 3D models on demand~\cite{ToriiPAMI19AreLargeScale3DModelsNecessaryforVisLoc}. 
If high pose accuracy is not required, the query pose can be approximated very efficiently via a combination of the poses of the top retrieved database images~\cite{ToriiICCVWS11VisualLocalizationByLinearCombination,ZamirECCV10AccurateImageLocalization,ToriiPAMI19AreLargeScale3DModelsNecessaryforVisLoc}. 

As illustrated in Fig.~\ref{fig:overview}, there are various roles that image retrieval can play in visual localization systems:  
\textbf{Efficient pose approximation} by representing the pose of a query image by a (linear) combination of the poses of retrieved database images~\cite{ToriiICCVWS11VisualLocalizationByLinearCombination,ZamirECCV10AccurateImageLocalization,ToriiPAMI19AreLargeScale3DModelsNecessaryforVisLoc} \textbf{(\Task~1)}. 
\textbf{Accurate pose estimation without a global 3D map} by computing the pose of the query image relative to the known poses of retrieved database images~\cite{ZhangS3DPVT06ImageBasedLocUrbanEnvironments,ZhouICRA20ToLearnLocalizationFromEssentialMatrices,LaskarICCVWS17CameraRelocalizationRelativePosesCNN,BalntasECCV18RelocNetMetricLearningRelocalisation,DingICCV19CamNetRetrievalForReLocalization,ToriiPAMI19AreLargeScale3DModelsNecessaryforVisLoc} \textbf{(\Task~2a)}. 
\textbf{Accurate pose estimation with a global 3D map} by estimating 2D-3D matches between features in a query image and the 3D points visible in the retrieved images~\cite{IrscharaCVPR09FromSFMLocationRecognition,CaoCVPR13GraphBasedLocationRecognition,SattlerBMVC12ImRetLocalizationRevisited,SarlinCVPR19FromCoarsetoFineHierarchicalLocalization,Germain3DV19SparseToDenseHypercolumnMatchingVisLoc,TairaPAMI19InLocIndoorVisualLocalization} \textbf{(\Task~2b)}.

These three \tasks \, serve different application scenarios: \Task~1 is useful if no accurate camera pose is required, such as a scenario where it is enough to roughly understand the surrounding of the device.
If a very accurate position is needed, such as for augmented reality or precise robot navigation, \Tasks 2a and 2b can be used.
While \Task~2b often produces the most accurate pose, it also comes with the burden of creating and maintaining a 3D map.

\crv{Considering memory consumption and processing time, pose approximation approaches (\Task~1), such as the ones considered in this paper, have the most lightweight scene representation (in our case: a set of image descriptors and corresponding camera poses) and the pose can be computed efficiently with comparably little computation power.
Localization without a map (\Task~2a) has a similar lightweight scene representation but requires more computation time for pose estimation due to the need of constructing a local map. Localization with a global map (\Task~2b) has a significantly larger memory footprint due to the need of storing the 3D map and the associations between the map and the reference images (2D-3D correspondences between 2D features and 3D points). In addition, building the global map adds offline processing requirements.}

In terms of algorithms, these three \tasks \, also have differing requirements on the results of the retrieval stage: 
\Task~1 requires the retrieval step to find images taken from poses as similar as possible to the query, \ie, the image representation should not be too robust or invariant to changes in viewpoint. 
\Tasks 2a and 2b require the retrieval stage to find images depicting the same part of the scene as the query image. 
However, the retrieved images do not need to be taken from a similar pose as the query as long as local feature matching succeeds.
In fact, \Task~2a usually requires retrieving multiple images from a diverse set of viewpoints that differ from the query pose because the local 3D map needs to be created by triangulation of 3D points~\cite{LaskarICCVWS17CameraRelocalizationRelativePosesCNN,ZhouICRA20ToLearnLocalizationFromEssentialMatrices}.  
\Task~2b benefits from retrieving images of high visual overlap with the query image and (in theory) requires only one relevant database image.

Despite differing requirements, modern localization methods~\cite{ToriiPAMI19AreLargeScale3DModelsNecessaryforVisLoc,TairaPAMI19InLocIndoorVisualLocalization,SarlinCVPR19FromCoarsetoFineHierarchicalLocalization,DusmanuCVPR19D2NetDeepLocalFeatures,Germain3DV19SparseToDenseHypercolumnMatchingVisLoc,ZhouICRA20ToLearnLocalizationFromEssentialMatrices} indiscriminately use the same representations based on compact image-level descriptors~\cite{ToriiPAMI18247PlaceRecognitionViewSynthesis,ArandjelovicCVPR16NetVLADPlaceRecognition}. 
These descriptors are typically trained for landmark retrieval and place recognition with the goal to produce similar descriptors for images showing the same building or place independently of the pose or other viewing conditions~\cite{LiuICCV19StochasticAttractionRepulsionEmbedding,RadenovicPAMI19FineTuningCNNImRet}. 
Interestingly, to the best of our knowledge, there is no work analyzing the suitability of such descriptors on the three visual localization \tasks.

In order to close this gap in the literature, this paper investigates the role of image retrieval for visual localization. 
We design a benchmark to measure the correlation between localization and retrieval/recognition performance for each \task using four global image features. 
Our benchmark enables a fair comparison of different retrieval approaches by fixing the remaining parts of the localization pipeline.

Our main contributions are the benchmark, a set of extensive experiments, and the conclusions we draw from them: 
(\textbf{1}) There is a strong correlation between landmark retrieval performance and \Task~1, and only NetVLAD~\cite{ArandjelovicCVPR16NetVLADPlaceRecognition} improves pose accuracy by interpolation when retrieving more than one image (with a drop when $k$ is high).
(\textbf{2}) Similarly, \Task~2b correlates with the classical place recognition \task, but when compared with retrieval ground truth (upper bound), there is significant room for improvement. 
(\textbf{3}) Landmark retrieval and place recognition performance are not good indicators for performance on \Task~2a.
(\textbf{4}) In summary, our results clearly show that there is a need to design image retrieval approaches specifically tailored to the requirements of the localization \tasks.
To foster such research, our benchmark and evaluation protocols are publicly available at \url{https://github.com/naver/kapture-localization}. 

This work is an extended version of our paper published at 3DV 2020~\cite{Pion3DV20Benchmarking}. 
It extends the 3DV paper through the following additional contributions: 
\begin{itemize}
    \item We rewrote the experimental results section from scratch in order to provide a more thorough analysis.
    \item We analyze the correlation between the localization and retrieval \tasks~in more detail using three correlation measures compared to the single measure used in~\cite{Pion3DV20Benchmarking}:
    \begin{itemize}[label=\textbullet]
        \item per query image linear correlation (Pearson) using a range of top $k$ retrieved images,
        \item same as above but per dataset (percentage of images localized), 
        \item and per dataset rank correlation (Spearman) to analyze how two metrics rank different global feature types.
    \end{itemize}
    \item We analyze 3 strategies to define image retrieval ``ground truth" in the context of visual localization:  frustum-overlap, relative pose, and co-observations. 
    Using the top $k$ images obtained via these strategies, instead of the images retrieved by a particular method, we show that there is considerable room for improvement for each localization \task. 
    At the same time, these definitions and their performance hint at how training data for image retrieval in the context of visual localization should be obtained. 
    \item We analyze the effects of blur and dynamic scenes on image retrieval performance.
    \item We use two additional datasets to strengthen our findings: InLoc~\cite{TairaPAMI19InLocIndoorVisualLocalization} and our own GangnamStation\_B2~\ccc{\cite{LeeCVPR21LargeScaleLocalizationDatasetsCrowdedIndoorSpaces}} dataset.
\end{itemize}



\section{Related work}

\PAR{Landmark retrieval}
Landmark retrieval is the task of identifying all relevant database images depicting the same landmark as a query image. 
Early methods relying on global image statistics were significantly outperformed by methods based on aggregating local features, most notably the bag of visual words representations for images~\cite{SivicICCV03VideoGoogle,CsurkaECCVWS04VisualCategorizationBagsKeypoints} and its extensions such as Fisher Vectors~\cite{PerronninCVPR07FisherKernels} and the Vector of Locally Aggregated Descriptors (VLAD)~\cite{JegouCVPR10AggregatingLocalDescriptors}.
More recently, deep representation learning has led to further improvements. 
They apply various pooling mechanisms~\cite{RazavianTMTA15VisualInstanceRetCNN,ToliasICLR16ParticularObjRetIntegralMaxpoolingCNN,BabenkoICCV15AggregatingDeepConvolutionalFeature,KalantidisECCVWS16CrossDimensionalWeightingAggregatedDeepFeatures,ToliasICLR16ParticularObjRetIntegralMaxpoolingCNN,RadenovicPAMI19FineTuningCNNImRet,ArandjelovicCVPR16NetVLADPlaceRecognition} on activations in the last convolutional feature map of CNNs in order to construct a global image descriptor. 
They learn the similarity metric by using ranking losses such as contrastive, triplet, or average precision (AP)~\cite{BabenkoECCV14NeuralCodesImRet,RadenovicPAMI19FineTuningCNNImRet,GordoIJCV17EndToEndDeepVisualReprImRet,RevaudICCV19LearningwithAPTrainingImgRetrievalListwiseLoss}. 

Several benchmark papers compare such image representations on the task of instance-level landmark retrieval~\cite{PhilbinCVPR08LostInQuantization,ArandjelovicCVPR13AllAboutVLAD,ZhengX16GoodPracticeCNNFeatureTransfer,NohICCV17LargeScaleAttentiveDeepLocalFeatures,RadenovicCVPR18RevisitingOxfordParisImRetBenchmarking,WeyandCVPR20GoogleLandmarksDatasetv2}.
In contrast, this paper explores how state-of-the-art landmark retrieval approaches perform in the context of visual localization.

\PAR{Visual localization}
Traditionally, structure-based methods establish 2D-3D correspondences between a query image and a 3D map, typically via matching local feature descriptors~\cite{SchoenbergerCVPR17ComparativeEvaluationLocalFeatures,CsurkaX18FromHandcraftedToDeepLocalFeatures} and use them to compute the camera pose by solving a perspective-n-point (PNP)  problem~\cite{KneipCVPR11ANovelParametrizationAbsoluteCamPose,KukelovaICCV13RealTimeSolutionAbsolutePoseProblem,LarssonICCV17MakingMinimalSolversAbsPoseEstimation} robustly inside a RANSAC~\cite{FischlerCACM81RandomSampleConsensus,ChumPAMI08OptimalRandomizedRANSAC,LebedaBMVC12FixingTheLocallyOptimizedRANSAC} loop.  

More recently, scene coordinate regression techniques determine these correspondences using random forests~\cite{MassicetiICRA17RandomForestVersusNNCamLoc,ShottonCVPR13SceneCoordinateRegression} or CNNs~\cite{BrachmannCVPR18LearningLessIsMore6DLocalization,MassicetiICRA17RandomForestVersusNNCamLoc,BrachmannICCV19ExpertSampleConsensusReLocalization}.
Earlier methods trained a regressor specifically for each scene while recent models are able to adapt the trained model on-the-fly to new scenes~\cite{CavallariPAMI19RealTimeRGBDCamPoseEstimation,Cavallari3DV17LetsTakeThisOnlineSceneCoordinateRegression}. \ccc{More recently proposed dense coordinate regression networks, such as SANet~\cite{YangICCV19SANetSceneAgnosticLocalization} or  DSM~\cite{TangCVPR21LearningCamLocDenseSceneMatching} have been shown to be able to perform in scene-agnostic manner.}
Even if scene coordinate regression methods achieve high pose accuracy on small datasets, they currently do not scale up well to larger and more complex scenes~\cite{TairaPAMI19InLocIndoorVisualLocalization,LiECCV10LocationRecPriorFeatureMatching,SattlerCVPR18Benchmarking6DoFOutdoorLoc,WeinzaepfelCVPR19VisualLocObjectsOfInterestDenseMatchRegression,BrachmannICCV19ExpertSampleConsensusReLocalization}.
To overcome this, \cite{LiCVPR2020HierarchicalSceneCoordinateClassification} proposes a coarse-to-fine strategy within the neural network to increase the size of environments scene point regression can be successfully used.
For large-scale datasets however, \cite{LiCVPR2020HierarchicalSceneCoordinateClassification} suggests to use image retrieval as additional conditioning.
\ccc{The scene agnostic neural network introduced by \cite{SarlinCVPR21BackToTheFeatureLearningRobustCameraLocPixelsToPose}, learning strong
data priors by end-to-end training from pixels to pose separating model parameters and scene geometry, can localize in large environments by aligning the image to an explicit 3D model of the scene based on dense features.}
In this work, however, we focus on feature-based localization methods that use image retrieval to cope with the problem of large scenes~\cite{RevaudX19R2D2ReliableRepeatableDetectorsDescriptors,SarlinCVPR19FromCoarsetoFineHierarchicalLocalization,DusmanuCVPR19D2NetDeepLocalFeatures}.

Absolute pose regression methods forego 2D-3D matching and train a CNN to directly predict the full camera pose from an image for a given scene~\cite{KendallICCV15PoseNetCameraRelocalization,KendallCVPR17GeometricLossCameraPoseRegression,WalchICCV17ImagebasedLocalizationUsingLSTMs,BrahmbhattCVPR18GeometryAwareLocalization}. 
However, they are significantly less accurate than structure-based methods~\cite{SattlerCVPR19UnderstandingLimitationsPoseRegression} and currently not (significantly) more accurate than simple retrieval baselines~\cite{SattlerCVPR19UnderstandingLimitationsPoseRegression} but significantly less scalable~\cite{SattlerCVPR18Benchmarking6DoFOutdoorLoc}.
This is why we focus on image retrieval for efficient and scalable pose approximation instead. 

Accurate real-world visual localization needs to be robust to a variety of conditions, including day-night, weather and seasonal variations.
\cite{SattlerCVPR18Benchmarking6DoFOutdoorLoc} introduces several benchmark datasets specifically designed for analyzing the impact of such factors on visual localization, using query and training images taken under varying conditions.
For our benchmark, we use the following datasets: Aachen Day-Night-v1.1~\cite{SattlerCVPR18Benchmarking6DoFOutdoorLoc,ZhangIJCV20ReferencePoseGenerationVisLoc}, RobotCar Seasons~\cite{MaddernIJRR171YearOxfordRobotCarDataset,SattlerCVPR18Benchmarking6DoFOutdoorLoc}, Baidu shopping mall~\cite{SunCVPR17DatasetBenchmarkingLocalization}, InLoc~\cite{TairaPAMI19InLocIndoorVisualLocalization} and GangnamStation\_B2.
The latter is a new dataset captured in the busy and crowded Gangnam metro station in Seoul (see Sec.~\ref{sec:framework:datasets} for more details).
For more details on aspects of visual localization, please see recent survey papers~\cite{GarciaFidalgoRAS15VisionBasedTopologicalMapLocSurvey,LowryTROB16VisualPlaceRecognitionSurvey,ZamirB16LargeScaleVisualGeoLocalization,BrejchaPAA17StateSOAVisualGeoLocalization,PiascoPR18ASurveyVisualBasedLocHeterogeneousData} and benchmarks~\cite{ToriiPAMI19AreLargeScale3DModelsNecessaryforVisLoc,SattlerCVPR18Benchmarking6DoFOutdoorLoc,SattlerCVPR19UnderstandingLimitationsPoseRegression}.

\PAR{Place recognition}
Place recognition, also called visual geo-localization~\cite{ZamirB16LargeScaleVisualGeoLocalization}, lies between landmark retrieval and visual localization.
While, similar to the latter, its goal is to estimate the camera location, a coarse position of the image is considered sufficient~\cite{SchindlerCVPR07CityScaleLocationRecognition,HaysCVPR08IM2GPSGeographicInformation,ZamirECCV10AccurateImageLocalization,VoICCV17RevisitingIM2GPS}.
It is often important to explicitly handle confusing~\cite{SchindlerCVPR07CityScaleLocationRecognition,KnoppECCV10AvoidingConfusingFeatures} and repetitive scene elements~\cite{ToriiPAMI15VisualPlaceRecognRepetitiveStructures,SattlerCVPR16LargeScaleLocationRecognitionGeometricBurstiness,ArandjelovicACCV14DislocationDistinctivenessForLocation}, especially in large urban scenes. 
To improve scalability, a popular strategy is to perform visual and geo-clustering~\cite{CrandallWWW09MappingTheWorldsPhotos,LIICCV09LandmarkClassificationLargeScaleImageCollections,CaoCVPR13GraphBasedLocationRecognition,AvrithisACMMM10RetrievingLandmarkCommunityPhoto,KalantidisMTA11VIRaLVisualImgRetLocalization}.

As image matching and retrieval are key ingredients of place recognition, several papers proposed improved image representations using GPS and geometric information as a form of weak supervision~\cite{ArandjelovicCVPR16NetVLADPlaceRecognition,VoICCV17RevisitingIM2GPS,KimCVPR17LearnedContextualFeatureReweightingGeoLoc,RadenovicPAMI19FineTuningCNNImRet}.
In this paper, a.o., we use NetVLAD~\cite{ArandjelovicCVPR16NetVLADPlaceRecognition}, which is probably the most popular representation trained this way, as well as DenseVLAD~\cite{ToriiPAMI18247PlaceRecognitionViewSynthesis}, its handcrafted counterpart.  \ccc{Recently, several extensions were proposed to improve NetVLAD for place recognition, such as adding attention layers~\cite{KimCVPR17LearnedContextualFeatureReweightingGeoLoc} or 
multi-scale fusion of patch-level features from NetVLAD residuals~\cite{HauslerCVPR21PatchNetVLADMultiScaleFusionLocallyGlobalDescPlaceRec}
to increase the robustness of the model to photometric and geometric changes. 
Note that our publicly available benchmark pipeline allows to easily compare image representations within the proposed visual localization paradigms.}



\section{The proposed benchmark}
\label{sec:framework}

Modern localization algorithms often use state-of-the-art landmark retrieval and place recognition representations. 
However, different localization \tasks \, have different requirements on the retrieved images and thus on the used retrieval representations. 
In this paper, we are interested in understanding how landmark retrieval and place recognition performance relates to visual localization performance. 
In particular, we are interested in determining whether current state-of-the-art retrieval/recognition representations are sufficient or whether specialized (\task-dependent) representations for localization are needed. 

This is why, in this section, we present an evaluation framework designed to answer this question. 
Our framework enables a fair comparison of different retrieval approaches for each of the three localization \tasks \, by fixing the remaining parts of the localization pipeline.
On the one hand, it enables measuring localization performance for the three \tasks \, (\Task~1, \Task~2a, and \Task~2b) identified above, and on the other hand it enables measuring landmark retrieval and place recognition performance (also referred to as \Tasks 3a and 3b) on the same datasets.

\subsection{\Task~1: Pose approximation} 
\label{sec:task1}

Pose approximation methods are inspired by place recognition~\cite{ZamirECCV10AccurateImageLocalization,ToriiPAMI19AreLargeScale3DModelsNecessaryforVisLoc,ToriiICCVWS11VisualLocalizationByLinearCombination,SattlerCVPR19UnderstandingLimitationsPoseRegression} and aim to efficiently approximate the query pose from the poses of the top $k$ retrieved database images.
We represent a camera pose as a tuple $\mathbf{P} = (\mathbf{c},\mathbf{q})$. 
Here, $\mathbf{c}\in \mathbb{R}^3$ is the position of the camera in the global coordinate system of the scene and $\mathbf{q}\in\mathbb{R}^4$ is the rotation of the camera parameterized as a unit quaternion. 
We compute the pose of the query image as a weighted linear combination $\mathbf{P}_q = \sum_{i = 1}^k w_i \mathbf{P}_i$, where $\mathbf{P}_i$ is the pose of the top $i$ retrieved image and $w_i$ is a corresponding weight\footnote{$\mathbf{q}_q = \sum_i w_i \mathbf{q}_i$ is re-normalized to be a unit quaternion.}.
As a consequence, for $k=1$ we directly use the pose of the top retrieved image.

We consider three variants: 
\textbf{Equal weighted barycenter} (\textbf{EWB}) assigns the same weight to all of the top $k$ retrieved images with $w_i=1/k$.
\textbf{Barycentric descriptor interpolation} (\textbf{BDI})~\cite{ToriiICCVWS11VisualLocalizationByLinearCombination,SattlerCVPR19UnderstandingLimitationsPoseRegression} estimates $w_i$ as the best barycentric approximation of the query descriptor via the database descriptors by minimizing   
\begin{equation}
    \left\| \mathbf{d}_q - \sum_{i = 1}^k w_i \mathbf{d}_i \right\|_2 \enspace \text{subject to} \enspace \sum_{i=1}^k w_i = 1 \enspace.
    \label{eq:BDI}
\end{equation}
Here, $\mathbf{d}_q$ and $\mathbf{d}_i$ are the global image-level descriptors of the query image and the top $i$ retrieved database image, respectively. 
In the third approach, $w_i$ is based on the \textbf{cosine similarity} (\textbf{CSI}) between L2 normalized descriptors as
\begin{equation}
    w_i = \frac{1}{z_i} \left(\mathbf{d}_q^T \mathbf{d}_i\right)^\alpha  \enspace \text{, with} \enspace  z_i= \sum_{j = 1}^k \left(\mathbf{d}_q^T \mathbf{d}_j\right)^\alpha \enspace. 
    \label{eq:CalphaI}
\end{equation}
Setting $\alpha = 0$ reduces this method to \textbf{EWB}.
At the opposite, as $\alpha \to \infty$, the resulting pose becomes the one of the image with the highest similarity.
We fix $\alpha=8$ based on preliminary results on the Cambridge Landmarks~\cite{KendallICCV15PoseNetCameraRelocalization} dataset.
\Task~1 requires images taken from poses as similar as possible to the query, \ie, the image representation should not be too robust or invariant to changes in viewpoint. 

\subsection{\Task~2a: Pose estimation without a global map} 
\label{sec:task2a}

In theory, using the top 1 retrieved image would be sufficient for this \task as long as the relative pose between the query and this image could be estimated accurately, including the scale of the translation~\cite{BalntasECCV18RelocNetMetricLearningRelocalisation,DingICCV19CamNetRetrievalForReLocalization}.
In practice, retrieving $k>1$ images improves the accuracy because \emph{k} relative poses can be considered~\cite{ZhangS3DPVT06ImageBasedLocUrbanEnvironments,ZhouICRA20ToLearnLocalizationFromEssentialMatrices,LaskarICCVWS17CameraRelocalizationRelativePosesCNN,ToriiPAMI19AreLargeScale3DModelsNecessaryforVisLoc}. 
Once the relative poses between query and database images are estimated, triangulation can be used to compute the absolute pose~\cite{ZhangS3DPVT06ImageBasedLocUrbanEnvironments,ZhouICRA20ToLearnLocalizationFromEssentialMatrices,LaskarICCVWS17CameraRelocalizationRelativePosesCNN}. 
However, pose triangulation fails if the query pose is co-linear with the poses of the database images, which is often the case in autonomous driving scenarios.

Therefore, we follow~\cite{ToriiPAMI19AreLargeScale3DModelsNecessaryforVisLoc} where the retrieved database images with known poses are used to build the 3D map of the scene on-the-fly\footnote{Note that compared to the query pose, the 3D points are very seldom co-linear with the reference poses and can thus be accurately triangulated.}. 
This model is then used to register the query image 
using PNP and RANSAC.
Similar to pose triangulation, this \textbf{local SFM} approach fails if
(i) less than two images among the top $k$ database images depict the same place as the query image,
(ii) the viewpoint change among the retrieved images and/or between the query and the retrieved images is too large to be handled by local feature matching,  
or (iii) the baseline (distance) between the retrieved database images is not large enough to allow stable triangulation of enough 3D points.
Thus, this approach requires retrieving a diverse set of images depicting the same scene as the query image from a variety of viewpoints. 
As such, methods for \Task~2a benefit from image representations that are robust but not invariant to viewpoint changes.

\subsection{\Task~2b: Pose estimation with a global map} 
\label{sec:task2b}

In contrast to \Task~2a, this \task~uses a pre-built global 3D model of the scene rather than reconstructing it locally on-the-fly.
We follow a standard local feature-based approach from the literature ~\cite{IrscharaCVPR09FromSFMLocationRecognition,SattlerBMVC12ImRetLocalizationRevisited,SarlinCVPR19FromCoarsetoFineHierarchicalLocalization,Germain3DV19SparseToDenseHypercolumnMatchingVisLoc} where an SFM model of the scene provides the correspondences between local features in the database images and 3D points in the map.
Establishing 2D-2D matches between the query image and the top ranked database images yields a set of 2D-3D matches which are then used for pose estimation via PNP and RANSAC. 

In theory, retrieving a single relevant image among the top $k$ is sufficient as long as the viewpoint change between the query and this image can be handled by the local features. 
However, retrieving more relevant images increases the chance for accurate pose estimation.
Still, for two reasons, $k$ should be as small as possible: 
As local feature matching is often the bottleneck in terms of processing time, a small $k$ improves efficiency.
As accurate pose estimation depends on robust outlier rejection using RANSAC, a large $k$ might also increases noise due finding matches with irrelevant images.
Overall, we expect this \task~to benefit from retrieval representations that are moderately robust to viewpoint changes while still allowing reliable local feature matching.

\subsection{\Task~3a: Landmark retrieval}
\label{sec:task3a}

This is an instance retrieval task where all images containing the main object of interest shown in the query image are to be retrieved from a large database of images.
Thus, image representations should be as robust as possible to viewpoint and viewing condition changes in order to identify all relevant images. 
In order to determine whether a retrieved image is relevant for a query, \cite{RadenovicPAMI19FineTuningCNNImRet} proposes a 3D model-based definition where the similarity of two images is computed as the number of 3D scene points observed by both images in an SFM model. 
Among other methods (see Sec.~\ref{sec:gt}), we follow this definition in this paper.

\subsection{\Task~3b: Place recognition} 
\label{sec:task3b}

This task aims to approximately determine the place from which a given query image was taken. 
Since the place is defined by the location of the retrieved images, this requires at least one relevant reference image amongst the top $k$ retrieved ones. 
A database image is typically considered relevant if it was taken within a spatial neighborhood of the query image~\cite{ToriiPAMI18247PlaceRecognitionViewSynthesis,ToriiPAMI15VisualPlaceRecognRepetitiveStructures,ArandjelovicACCV14DislocationDistinctivenessForLocation,SattlerCVPR16LargeScaleLocationRecognitionGeometricBurstiness}. 
If camera poses or geo-tags are available, as is the case in this paper, the selection can be done using a distance threshold. 
We thus follow this common definition from the literature to define relevant images for the place recognition task.

\subsection{Metrics}

In this section, we describe the metrics we use for evaluation of visual localization, landmark retrieval, and place recognition performance.

\subsubsection{Visual localization metrics}
\label{sec:locmetric}

To measure localization performance, we follow common practice from the literature~\cite{KendallICCV15PoseNetCameraRelocalization,ShottonCVPR13SceneCoordinateRegression,SattlerCVPR18Benchmarking6DoFOutdoorLoc}.
Let $\mathtt{R}\in \mathbb{R}^{3\times 3}$ be the camera rotation and $\mathbf{c}\in\mathbb{R}^3$ be the camera position, \ie, a 3D point $\mathbf{X}_g$ in world coordinates is mapped to local camera coordinates as $\mathbf{X}_l = \mathtt{R}(\mathbf{X}_w - \mathbf{c})$. 
Following~\cite{SattlerCVPR18Benchmarking6DoFOutdoorLoc}, the position and rotation errors between an estimated pose and the reference pose are defined as
\begin{eqnarray}
c_\text{error} & = & \lVert \mathbf{c}_\text{estimated} - \mathbf{c}_\text{reference} \rVert_2  \label{eq:position_error} \\
R_\text{error} & = & \arccos\left(\frac{\text{trace}\left(\mathtt{R}_\text{estimated}^{-1} \cdot \mathtt{R}_\text{reference}\right) - 1}{2} \right)  \label{eq:orientation_error}
\end{eqnarray}
where $R_\text{error}$ is the angle of the smallest rotation aligning $\mathtt{R}_\text{estimated}$ and $\mathtt{R}_\text{reference}$.
For evaluation, we use the percentage of images localized within a given error threshold $(X\text{m}, Y^\circ)$~\cite{ShottonCVPR13SceneCoordinateRegression,SattlerCVPR18Benchmarking6DoFOutdoorLoc}, \ie, the percentage of query images for which $c_\text{error} < X$ and $R_\text{error}<Y$.
Following~\cite{SattlerCVPR18Benchmarking6DoFOutdoorLoc}, for all datasets we use three different threshold pairs evaluating \textbf{low} (5m, 10$^\circ$), \textbf{medium} (0.5m, 5$^\circ$), and \textbf{high} (0.25m, 2$^\circ$) accuracy localization.

\subsubsection{Retrieval metrics }
\label{sec:retmetric}

In classical image or instance retrieval and, in particular, landmark retrieval, most common metrics used to evaluate retrieval performance are \emph{Precision@$k$} (P@$k$), \emph{Recall@$k$} (R@$k$), and \emph{mean Average Precision} (mAP).
All of them require binary (\ie is relevant or not) relevance scores. 
In landmark retrieval, such a binary relevance score is defined by whether or not a landmark is visible in the image.
In place recognition, the distance between query and mapping images can be used to compute a relevance score.
In visual localization, the definition of relevance is less well defined and it depends on the \task. Nevertheless, all \tasks~share at least one requirement: The two images need to see partially the same scene.

\crv{In order to select suitable relevance scores, we roughly divide the methods in three groups using (i) camera position, (ii) overlapping views, and (iii) image content. 
These groups cover different aspects of visual localization. 
While using image content intuitively is the best choice, depending on the method used to compare images, some bias is introduced (\eg~when using local features). 
Using the camera position as criterion is easy and fast to compute but it does not guarantee maximizing the view overlaps (which is beneficial for visual localization). 
Comparing and intersecting camera frusta maximizes view overlap but is heavy to compute, the camera orientations need to be known, and it is, same as for using camera position, not robust to occlusions (fursta could overlap even if the images do not share the same content, \eg~when a wall is between the two images).
In order to provide more insights to these aspects as part of the benchmark, we selected one candidate/method of each group, being distance~\cite{ArandjelovicACCV14DislocationDistinctivenessForLocation,SattlerCVPR16LargeScaleLocationRecognitionGeometricBurstiness,ToriiPAMI18247PlaceRecognitionViewSynthesis,ToriiPAMI15VisualPlaceRecognRepetitiveStructures} (note that in the literature, sometimes only position/translation is used), frustum~\cite{BalntasECCV18RelocNetMetricLearningRelocalisation}, and co-observations~\cite{RadenovicPAMI19FineTuningCNNImRet}.}

While the first two methods only require the camera poses, for the last method an SFM reconstruction is necessary.
This additional effort, however, enables to also take the image content into account and not only camera pose. 
As detailed below, we use all three definitions for our experiments. 

\crv{Note that there are of course other ways of taking image content into account, such as any kind of intensity-based or feature-based~\cite{SarlinCVPR21BackToTheFeatureLearningRobustCameraLocPixelsToPose} similarity estimation. 
The challenge with such methods in visual localization is that the test images are often taken under very different viewing conditions (day-night, weather, seasonal changes) and thus, intensities (and even geometry) might not be directly comparable in all parts of the images.
Recent work on analysing ground truth bias~\cite{brachmannICCV2021limits} also shows that the way of scoring in ground truth generation can have a significant impact when used for benchmarking algorithms.
That is why we selected three simple ways of computing our ground truth rankings. 
They are only effected by the accuracy of the provided camera poses (distance and frustum) and the ability of computing robust local feature matches (co-observations).}

\subsubsection{Retrieval relevance scores between images }
\label{sec:gt}

In this section, we assess three methods to compute ground truth (GT) relevance scores in the context of retrieval/recognition and visual localization. 
In order to assess which GT is best suited for the localization \tasks, we propose the following protocol. 
For each method $M_j$, we define a score between two images $s_j(q,t)$, where $q$ is a query image and $t$ is an image from the reference (or training) set.
These non-binary scores can be used to rank the training images for all queries.
Therefore, to evaluate them in various localization \tasks, we will use them similarly to the global image representations used for retrieval: 
We consider the top training images $t_k$ according to the scores $s_j(q,t_k)$ and use them in our localization pipeline (\Task~1, \Task~2a, \Task~2b) described in Sec.~\ref{sec:framework}.
Then, we use the metrics described in Sec.~\ref{sec:locmetric} to evaluate the localization performance. 
In this way, we can assess which GT retrieval method provides the best upper bound for the localization \task.
The best methods will then be used to analyze the correlation between landmark retrieval, place recognition, and visual localization.
Table~\ref{tab:GT} reports the basic statistics about the resulting retrieval GT on the datasets used in this paper\footnote{Note that only datasets with publicly available ground truth are used to generate these results. For Aachen Day-Night and InLoc, no GT poses were available (see Sec.~\ref{sec:framework:datasets}).}.
Since frustum-overlap and relative camera pose do not take image content into account, they might produce more images per query than co-observations (see Tab.~\ref{tab:GT}), but some of them might actually be irrelevant (\eg because their frusta overlap but there is a wall between the cameras, thus, they do not see the same scene).
We only consider top $k$ up to 50 in this paper, that is why we also reduce the GT rankings to top 50.

\PAR{Relative camera pose}
One possibility to compute if a retrieved image is relevant to a query, often used in place recognition  ~\cite{ToriiCVPR15PlaceRecognitionByViewSynthesis,ToriiPAMI15VisualPlaceRecognRepetitiveStructures,ArandjelovicACCV14DislocationDistinctivenessForLocation,SattlerCVPR16LargeScaleLocationRecognitionGeometricBurstiness}, is to consider the relative camera pose between them. 
Images are considered relevant if they were taken within a given radius around the query image (\eg 25 meters) and look in the same direction (\eg less than 45\degree apart).
We modified this relevance score to get a non-binary measure $s_{rcp}(q,t)= \frac{c_\text{diff}}{\tau_c} + \frac{R_\text{diff}}{\tau_R}$, where $c_\text{diff}$ and $R_\text{diff}$, similar to (\ref{eq:position_error}) and (\ref{eq:orientation_error}), are computed between $(\mathbf{c}_\text{q},\mathtt{R}_\text{q})$ and $(\mathbf{c}_\text{t},\mathtt{R}_\text{t})$.
The role of $\tau_c$ and $\tau_R$ is to normalize distance and angle, making them more comparable.
We use $\tau_c=25m$ and $\tau_R=45^\circ$ in our experiments. 

\PAR{Overlap between the viewing frusta}
Another possibility is to consider the overlap of the camera frusta~\cite{BalntasECCV18RelocNetMetricLearningRelocalisation}. 
The frustum is typically obtained by a truncation of the pyramid of vision with two parallel planes.
The relevance score $s_{fr}(q,t)$ is defined as the radius of the maximum size sphere that can be inscribed in the intersection of the viewing frusta.
This score is related to the distance to the far plane that truncates the frusta.
This distance depends on the application scenario and the datasets used.
In our experiments, we set it to 25m or 50m, depending on the dataset (see the legends of the figures in Sec.~\ref{sec:experiments}).

\PAR{Shared 3D observations}
The third method defines the relevance score for each $(q,t)$ pair as the number of shared 3D observations in a joint SFM map.
In this paper, we directly use this number as $s_{obs}(q,t)$, \ie, a database image is considered as relevant if it shares at least one observed 3D point with the query image~\cite{RadenovicPAMI19FineTuningCNNImRet}.

\begin{table}[t!]
\center
\caption{Statistics for retrieval ground truth methods: \emph{avg k} is the average number of relevant database images for all queries, \emph{missing} indicates the portion of queries (\%) without any relevant database image.}
\label{tab:GT}
\resizebox{\linewidth}{!}{
\input{IJCV_plots/table_GT}
}
\end{table}

\subsubsection{Correlation between metrics}
\label{sec:corr}

\myparagraph{Pearson correlation coefficient}
In order to assess the correlation between two metrics $A$ and $B$ (\eg, recall and localization accuracy), we use the Pearson correlation coefficient (PCC)~\cite{PearsonPRSL85NotesRegressionInheritanceTwoParents} that seeks a linear relationship between the measures.
Let $\{(a^j_k,b^j_k)\}_{j,k}$ be a set of paired results obtained with the metrics $A$ and $B$ for a query image $q$. 
$k$ corresponds to the results obtained with top $k$ retrieved images used for localization respectively to compute the retrieval metric (precision or recall). 
$j$ denotes a given global image representation used for retrieval. 
To compute the correlation between A and B, for each global representation $j$, the Pearson correlation score is defined as follows:
\begin{eqnarray}
\begin{aligned}
\rho_P(q,j)= & \frac{\sum_{k} (a^j_k-\mu^j_a) (b^j_k-\mu^j_b)}{\sqrt{\sum_{k}{(a^j_k-\mu^j_a)^2}} \sqrt{\sum_{k}{(b^j_k-\mu^j_b)^2}}} \\              & \in [-1,1],
\end{aligned}
\label{eq:pearson}
\end{eqnarray}
where $\mu^j_a$ is the mean of $a^j_k$ and $\mu^j_b$ the mean of $b^j_k$. 

\myparagraph{Spearman correlation coefficient}
In addition to linear correlation, we would like to assess how similarly two measures (\eg, recall and localization accuracy) rank different global feature types with respect to each other.
For this we use the Spearman rank correlation coefficient (SRC).
The SRC coefficient is defined as the Pearson correlation coefficient between the rank variables~\cite{MyersB03ResearchDesignStatisticalAnalysis}, \ie, the covariance of the two variables divided by the product of their standard deviations.
While PCC aims to analyze a linear relationship between two metrics, the SRC scores high when two metrics are monotonically related, even if their relationship is non-linear.
Furthermore, as it analyzes ranks, it is less sensitive to outliers.

Let $\alpha^j_k$ and $\beta^j_k$ be the ranks obtained for the global feature $j$ with metrics $A$ and $B$ for a given $k$, \ie the rankings corresponding to $a^j_k$ and $b^j_k$, when $k$ is fixed. 
The SRC is defined as
\begin{eqnarray}
\begin{aligned}
\rho_S(q,k)= & \frac{\sum_{j} (\alpha^j_k-\eta^k_\alpha) (\beta^j_k-\eta^k_\beta)}{\sqrt{\sum_{j}{(\beta^j_k-\eta^k_\beta)^2}} \sqrt{\sum_{j}{(\beta^j_k-\eta^k_\beta)^2}}} \\
             & \in [-1,1],
\end{aligned}
\label{eq:spearman}
\end{eqnarray}
where $\eta^k_\alpha$ is the mean rank of $\alpha^j_k$ and $\eta^k_\beta$ the mean rank of $\beta^j_k$.
We compute the coefficients for every individual query image using the localization accuracy (PCC only), as well as for the entire dataset (PCC and SRC) using the percentage of localized images (both are defined in Sec.~\ref{sec:locmetric}).

In summary, here PCC evaluates if two metrics are correlated for a specific global feature type.
For example, it will be used to analyse if high recall also means high localization accuracy for a specific global feature type.
In addition, here SRC evaluates the ranking across different features types.
For example, it will be used to analyse if global features types that rank high for recall, also rank high for localization.
Essentially, in this example, both, PCC and SRC answer the question if recall needs to be optimized to maximize localization accuracy.
The way we compute PCC (looking for linear correlation as a function of $k$) answers this question for each feature type, the way we compute SRC (assessing monotonic relationships) answers it more generally considering how recall ranks global feature types.

Both correlation coefficients, Pearson and Spearman, range from 1, meaning very high correlation, to -1, meaning inverse correlation.



\section{Experimental setup}
\label{sec:experiments}

In this section, we describe our experimental setup.
We first introduce the datasets we selected for our benchmark (Sec.~\ref{sec:framework:datasets}), followed by the global features we compare (Sec.~\ref{sec:global_features}), and the SFM pipeline we used (Sec.\ref{sec:SFM}).

\subsection{Datasets}
\label{sec:framework:datasets}

To evaluate the role of image retrieval in visual localization, we selected multiple datasets aimed at benchmarking visual localization: \textbf{Aachen Day-Night-v1.1}~\cite{SattlerCVPR18Benchmarking6DoFOutdoorLoc,SattlerBMVC12ImRetLocalizationRevisited,ZhangIJCV20ReferencePoseGenerationVisLoc}, \textbf{RobotCar Seasons}~\cite{MaddernIJRR171YearOxfordRobotCarDataset,SattlerCVPR18Benchmarking6DoFOutdoorLoc}, \textbf{Baidu Mall}~\cite{SunCVPR17DatasetBenchmarkingLocalization}, \textbf{InLoc}~\cite{TairaCVPR18InLocndoorVisualLocalization}, and our new dataset \textbf{GangnamStation\_B2}~\ccc{\cite{LeeCVPR21LargeScaleLocalizationDatasetsCrowdedIndoorSpaces}}.

Altogether, the selected datasets cover a variety of application scenarios: large-scale outdoor handheld localization under varying conditions (Aachen Day-Night), large-scale autonomous driving (RobotCar Seasons), small-scale indoor handheld localization with occlusions, reflective and transparent surfaces (Baidu Mall), weakly textured and repetitive structures (InLoc), and crowded scenes with moving objects, as well as symmetric structures (GangnamStation\_B2).

\PARR{The Aachen Day-Night-v1.1~\cite{SattlerCVPR18Benchmarking6DoFOutdoorLoc,SattlerBMVC12ImRetLocalizationRevisited,ZhangIJCV20ReferencePoseGenerationVisLoc} dataset} contains 6,697 high-quality training and 1015 test images from the old inner city of Aachen, Germany. 
The database images are taken under daytime conditions using handheld cameras. 
The query images are captured with three mobile phones at day and at night. 
This dataset represents a handheld scenario similar to augmented or mixed reality applications in city-scale environments. 
The dataset consists of two conditions, \emph{day} and \emph{night}, while no night images are provided for training. 

\PARR{The RobotCar Seasons~\cite{MaddernIJRR171YearOxfordRobotCarDataset,SattlerCVPR18Benchmarking6DoFOutdoorLoc} dataset} is based on a subset of the RobotCar dataset~\cite{MaddernIJRR171YearOxfordRobotCarDataset}, captured in the city of Oxford, UK. 
The training sequences (26,121 images) are captured during daytime, the query images (11,934) are captured during different traversals and under changing weather, time of the day, and seasonal conditions. 
In contrast to the other datasets used, the RobotCar dataset contains multiple synchronized cameras and the images are provided in sequences. 
However, in our benchmark we did not use this additional information.
Following the online benchmark \emph{visuallocalization.net}~\cite{SattlerCVPR18Benchmarking6DoFOutdoorLoc}, we report results for all daytime images and for all nighttime images separately, \ie we do not consider more fine-grained conditions such as night-rain.

\PARR{The Baidu Mall~\cite{SunCVPR17DatasetBenchmarkingLocalization} dataset} was captured in a modern indoor shopping mall in China. It contains 689 training images captured with high resolution cameras in the empty mall and about 2,300 mobile phone query images taken a few months later while the mall was open.
The images were semi-manually registered into a LIDAR scan in order to obtain the ground truth poses.
The query images are of much poorer quality compared to the database images.
In contrary to the latter, where all images were taken in parallel or perpendicular with respect to the main corridor of the mall, query images were taken from more varying viewpoints.
Furthermore, the images contain reflective and transparent surfaces, moving people, and repetitive structures, which are all important challenges for visual localization and image retrieval.

\begin{figure}[t]
    \centering
    \includegraphics[width=\linewidth]{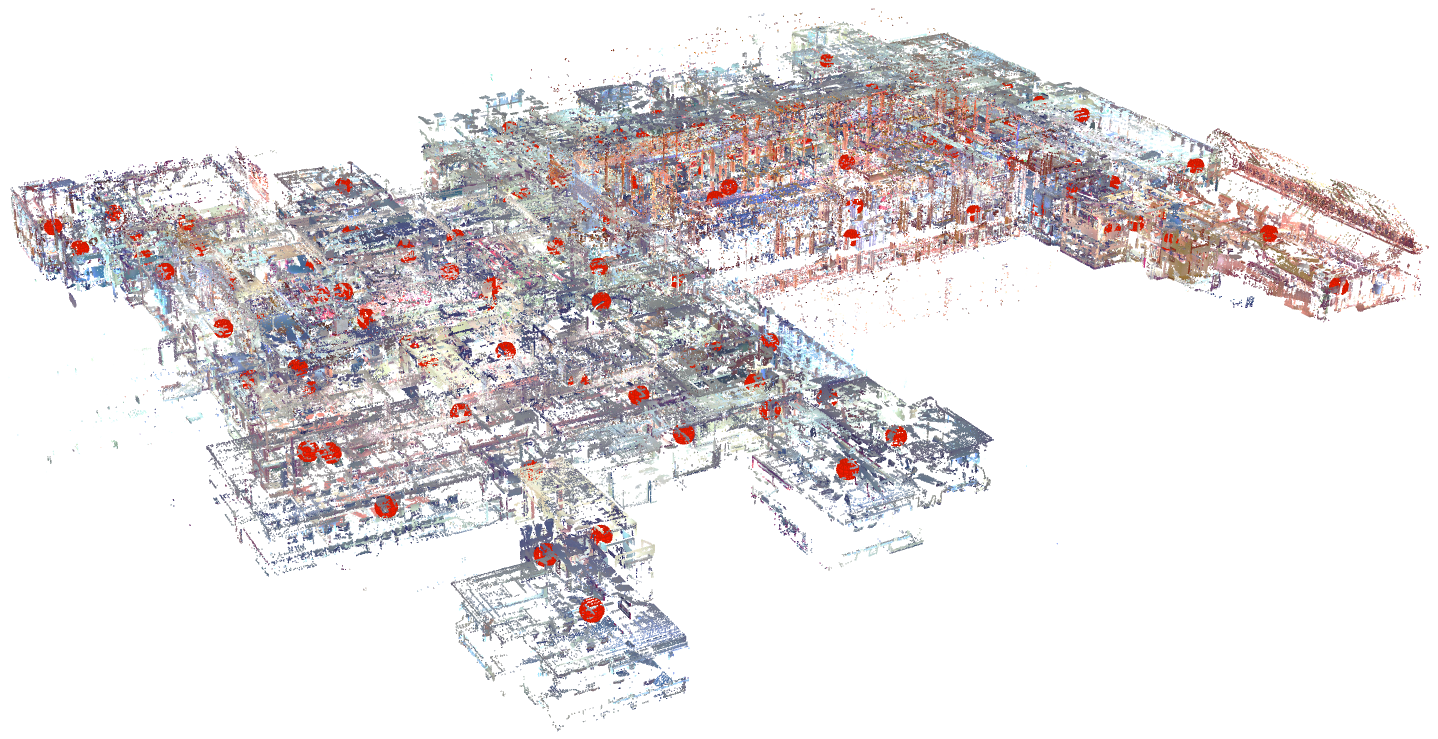}
    \caption{InLoc 3D map generated by assigning a 3D point to each local feature in the training images (viewed in COLMAP).}
    \label{fig:inloc}
\end{figure}

\PARR{The InLoc~\cite{TairaPAMI19InLocIndoorVisualLocalization,WijmansCVPR17Exploiting2DFloorplanBuildingScalePanoramaRGBDAlignment} dataset} represents an indoor handheld camera scenario with large viewpoint changes, occlusions, people, and movable furniture captured in various university buildings (only two, DUC1 and DUC2, are used for evaluation). 
There is only little visual overlap between the training images of this dataset which results in an SFM model which is, according to our experience, too sparse for the proposed visual localization pipeline. 
Furthermore, the InLoc environment is very challenging for global and local features because it contains large textureless areas and many repetitive structures. 
Contrary to the other datasets, InLoc provides 3D scan data for each training image and the original InLoc localization method~\cite{TairaPAMI19InLocIndoorVisualLocalization} introduced various dense matching and pose verification steps which make use of it to still achieve good localization performance.
For the experiments in this paper, we modified the SFM mapping pipeline (see Sec.~\ref{sec:SFM}) to use the provided 3D data. 
In detail, we constructed our SFM map using the provided 3D data and the camera poses by assigning a 3D point to each local feature in the training images. 
This results in a very dense 3D map (Fig.~\ref{fig:inloc}), where each 3D point is associated with a local descriptor and can, thus, be used in our benchmark.

\begin{figure}[t!]
    \centering
    \includegraphics[width=\linewidth]{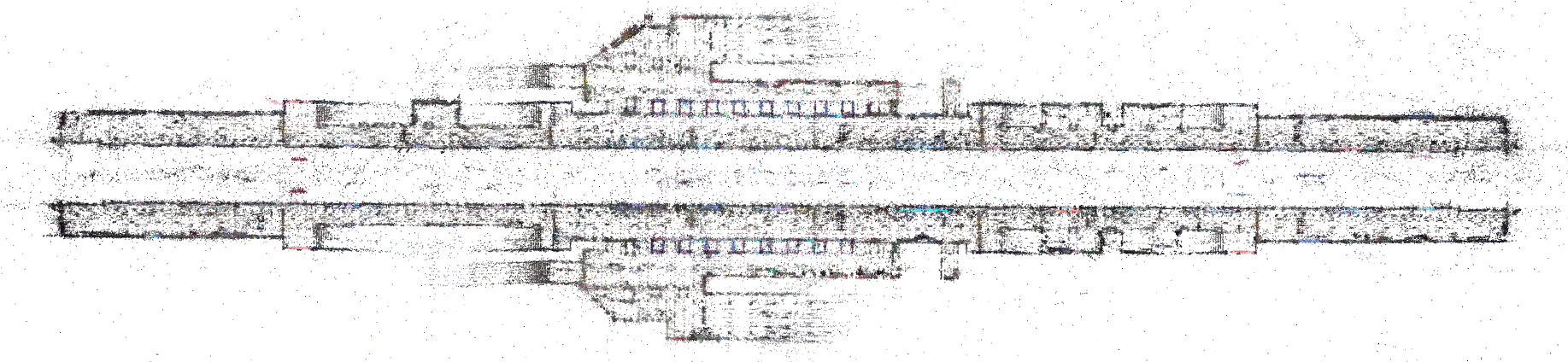} \\
    \caption{Top view of the 3D reconstruction of GangnamStation\_B2.}
    \label{fig:Gangnam}
\end{figure}

\PARR{The GangnamStation\_B2 dataset\ccc{\cite{LeeCVPR21LargeScaleLocalizationDatasetsCrowdedIndoorSpaces}}} was collected in one of the most crowded metro stations in Seoul. 
It is used to measure the robustness of visual retrieval and localization algorithms in scenes with many moving objects and people.
It was captured at the platforms of the metro station that were built with highly similar designs and internal 
structures, as shown in Figure~\ref{fig:Gangnam}. 
This introduces additional challenges for image retrieval and localization such as repetitive and symmetric scenes.
In addition, the dataset contains a lot of digital signage and platform screen doors that change appearance over time.
The datasets were recorded using a dedicated mapping device consisting of 10 cameras (6 industrial and 4 mobile phones) and 2 laser scanners.
The GT poses were obtained using SFM guided by the poses estimated using the platform's wheel odometry and LIDAR SLAM (based on pose-graph optimization~\cite{LuAR1997GloballyConsistent}).

\subsection{Global image representations for retrieval}
\label{sec:global_features}

Our benchmark compares four popular image representations using the best pre-trained models provided by the authors. 
Even if the impact of the dataset, used for training the representations, on visual localization performance is an interesting research question, a perfectly fair comparison would also require to equalize other parameters such as backbone and data augmentation.
Since this would distract from the core topic of this paper, we decided to use the pre-trained models.
Note that using pre-trained models is common practice in visual localization literature~\cite{DusmanuCVPR19D2NetDeepLocalFeatures,Germain3DV19SparseToDenseHypercolumnMatchingVisLoc,SattlerPAMI17EfficientPrioritizedMatching,HumenbergerX20RobustImageRetrievalBasedVisLocKapture}.

\PAR{DenseVLAD\footnote{Code available at~\url{http://www.ok.ctrl.titech.ac.jp/~torii/project/247/}.} \cite{ToriiCVPR15PlaceRecognitionByViewSynthesis}}
To obtain the DenseVLAD representation for an image, first RootSIFT~\cite{ArandjelovicCVPR12ThreeThingsEveryoneObjRet,LoweIJCV04DistinctiveImageFeaturesScaleInvariantKeypoints} descriptors are extracted on a multi-scale (we used four different scales corresponding to region widths of 16, 24, 32 and 40 pixels), regular, densely sampled grid, and then aggregated into an intra-normalized~VLAD~\cite{JegouCVPR10AggregatingLocalDescriptors} descriptor followed by PCA (principle component analysis) compression, whitening, and L2 normalization~\cite{JegouECCV12NegativeEvidencesCoOccurrences}.

\PAR{NetVLAD\footnote{Matlab code and pretrained models are available at~\url{https://github.com/Relja/netvlad}. We used the VGG-16-based NetVLAD model trained on Pitts30k~\cite{ArandjelovicCVPR16NetVLADPlaceRecognition}.} \cite{ArandjelovicCVPR16NetVLADPlaceRecognition}}
The main component of the NetVLAD architecture is a generalized VLAD layer that aggregates mid-level convolutional features extracted from the entire image into a compact single vector representation for efficient indexing similar to  VLAD~\cite{JegouCVPR10AggregatingLocalDescriptors}.
The resulting aggregated representation is then compressed using PCA to obtain a final compact descriptor of the image.
NetVLAD is trained with geo-tagged image sets consisting of groups of images taken from the same locations at different times and seasons. 
This allows the network to discover which features are useful or distracting and what changes should the image representation be robust against.
Furthermore, NetVLAD has been used in state-of-the-art localization pipelines~\cite{SarlinCVPR19FromCoarsetoFineHierarchicalLocalization,Germain3DV19SparseToDenseHypercolumnMatchingVisLoc} and in combination with D2-Net~\cite{DusmanuCVPR19D2NetDeepLocalFeatures} local features.

\PAR{AP-GeM\footnote{Pytorch implementation and models are available at \url{https://europe.naverlabs.com/Research/Computer-Vision/Learning-Visual-Representations/Deep-Image-Retrieval/}.} \cite{RevaudICCV19LearningwithAPTrainingImgRetrievalListwiseLoss}}
This image representation, similar to \cite{RadenovicPAMI19FineTuningCNNImRet}, uses a generalized-mean pooling layer (GeM) to aggregate CNN-based descriptors of several image regions at different scales. 
Instead of the contrastive loss used in \cite{RadenovicPAMI19FineTuningCNNImRet}, AP-GeM directly optimizes the Average Precision (AP). 
For training, AP is approximated by histogram binning to make it differentiable.
It is one of the state-of-the art image representation on popular landmark retrieval benchmarks (\eg, $\mathcal{R}$Oxford and $\mathcal{R}$Paris~\cite{RadenovicCVPR18RevisitingOxfordParisImRetBenchmarking}).
The model we used was trained on the Google Landmarks v1 dataset (GLD)~\cite{NohICCV17LargeScaleAttentiveDeepLocalFeatures}, where each training image has a class label based on the landmark shown in the image.
AP-GeM has been successfully used for visual localization in~\cite{HumenbergerX20RobustImageRetrievalBasedVisLocKapture}.

\PAR{DELG-GLDv2\footnote{We used the TensorFlow code publicly available at \url{https://github.com/tensorflow/models/tree/master/research/delf/delf/python/delg}.}~\cite{CaoECCV20UnifyingDeepLocalGlobalFeatures}} DELG is designed to extract local and global features using a single CNN.
After a common backbone, the model is split into two parts (heads) one to detect relevant local features and one which describes the global content of the image as a compact descriptor.
The two networks are jointly trained in an end-to-end manner using the ArcFace~\cite{DengCVPR19ArcFaceAdditiveAngularMarginLossFacRec} loss for the compact descriptor.
The model\footnote{Note that this is different from the model used in our 3DV paper~\cite{Pion3DV20Benchmarking}, where 
we used a model with ResNet50 backbone trained on GLD v1.} we use has a ResNet101 backbone and was trained on the Google Landmark v2~\cite{WeyandCVPR20GoogleLandmarksDatasetv2} dataset.
Initial experiments with the DELG local features showed that they perform significantly worse than R2D2~\cite{RevaudNIPS19R2D2ReliableRepeatableDetectorsDescriptors} or D2-Net~\cite{DusmanuCVPR19D2NetDeepLocalFeatures}, that is why we do not use them in our SFM pipeline and only focus on the global features.

\PAR{Discussion} Our choice of VLAD with densely extracted features (DenseVLAD, NetVLAD) is based on~\cite{TairaCVPR18InLocndoorVisualLocalization,ToriiPAMI18247PlaceRecognitionViewSynthesis}. 
It is shown that DenseVLAD and DenseFV (Fisher Vectors) significantly outperform SparseVLAD and SparseFV (based on local features) under strong illumination changes. 
This is due to the fact that using densely extracted features eliminates potential repeatability problems of feature detectors. 
Furthermore, \cite{ToriiPAMI18247PlaceRecognitionViewSynthesis} reports that DenseVLAD performs on par with advanced sparse bag of visual words representations. 
At the same time, both DenseVLAD and NetVLAD are used in state-of-the-art localization pipelines~\cite{SarlinCVPR19FromCoarsetoFineHierarchicalLocalization,TairaCVPR18InLocndoorVisualLocalization,TairaICCV19IsThisTheRightPlaceGeometricSemanticPoseVerification}.
Our choice of AP-GeM and DELG-GLDv2 is based on the fact that both perform well on standard image retrieval benchmarks (see Tab.~\ref{tab:RoxfordRparis}).

\subsection{SFM pipeline with and without a global model}
\label{sec:SFM}

For our experiments, we considered three types of local features: R2D2~\cite{RevaudNIPS19R2D2ReliableRepeatableDetectorsDescriptors}, D2-Net~\cite{DusmanuCVPR19D2NetDeepLocalFeatures}, and SIFT~\cite{LoweIJCV04DistinctiveImageFeaturesScaleInvariantKeypoints}.
Since \cite{Pion3DV20Benchmarking} shows that the same conclusions can be drawn when applying the benchmark with each of the three types, we only report results obtained with R2D2.

For global SFM (\Task~2b), for each dataset we created an SFM model by triangulating the 3D points from the local feature matches using the provided camera poses of the training images.
To not introduce a bias towards one specific global feature type and because matching all possible training image pairs would potentially introduce noise and would require large computational resources, we selected the image pairs to match by the overlap of the corresponding viewing frusta (see Sec.~\ref{sec:gt}).
For Aachen Day-Night and Baidu, we used all image pairs with an overlapping-sphere-radius of 10m or more.
For RobotCar and GangnamStation\_B2, this threshold results in too many image pairs to process, thus we only used the 50 most overlapping pairs.
In order to localize an image within this map, we match the query images with the top $k$ retrieved database images and use PNP and RANSAC to register them within the map.

Our local SFM experiments are inspired by the SFM-on-the-fly approach from~\cite{ToriiPAMI19AreLargeScale3DModelsNecessaryforVisLoc}, where the retrieved database images are used to create a small SFM map on-the-fly and the query images are registered within this map using PNP and RANSAC.
The SFM pipeline is the same as described above with the difference that the database image pairs to match are generated using all possible pairwise combinations of the retrieved images. 
For point triangulation and image registration (global and local SFM), we use COLMAP~\cite{SchonbergerCVPR16StructureFromMotionRevisited}.

\section{Experimental results}
\label{sec:experiments}

In this section, we present our experimental results and the findings we draw from them.
First, Sec.~\ref{sec:locExp} presents the results obtained using the proposed benchmark.
Second, in Sec.~\ref{sec:CorrExp} we evaluate the correlation of retrieval and localization metrics. 
Finally, in Sec.~\ref{sec:BlurrExp} we study how image retrieval for visual localization is effected by characteristics such as image blur or by the amount of dynamic objects such as persons or cars. 

\subsection{Evaluating the image retrieval component on the localization \tasks.}
\label{sec:locExp}

In order to evaluate the image retrieval component of the considered visual localization \tasks, in this section we present a comparison of the four selected global image features (see Sec.~\ref{sec:global_features}).
Furthermore, we compare different methods for obtaining GT for image retrieval (see Sec.~\ref{sec:gt}) and analyze how suited they are to act as an upper-bound on the performance for each \task.

\begin{figure*}[t]
\begin{center}
 \includegraphics[width=0.15\textwidth]{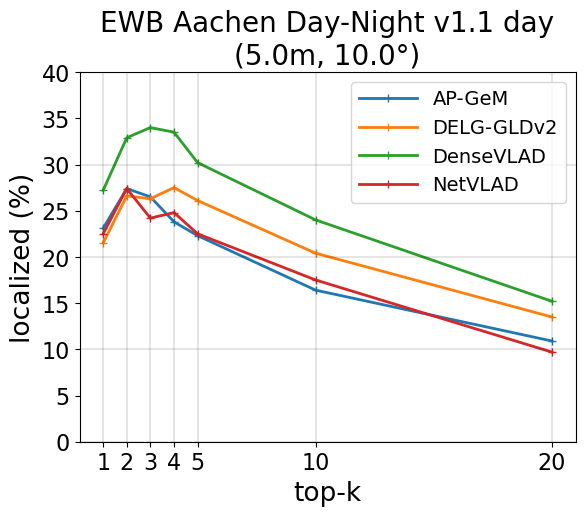}
 \includegraphics[width=0.15\textwidth]{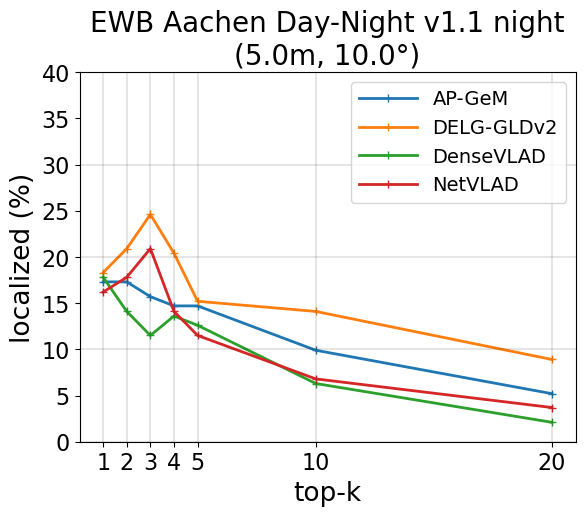}
 \includegraphics[width=0.15\textwidth]{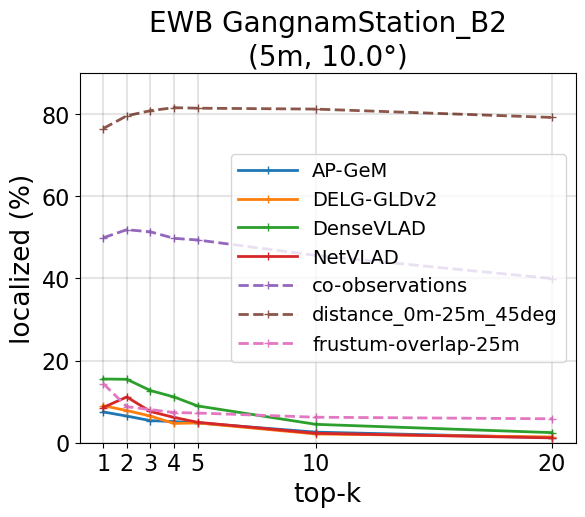}
 \includegraphics[width=0.155\textwidth]{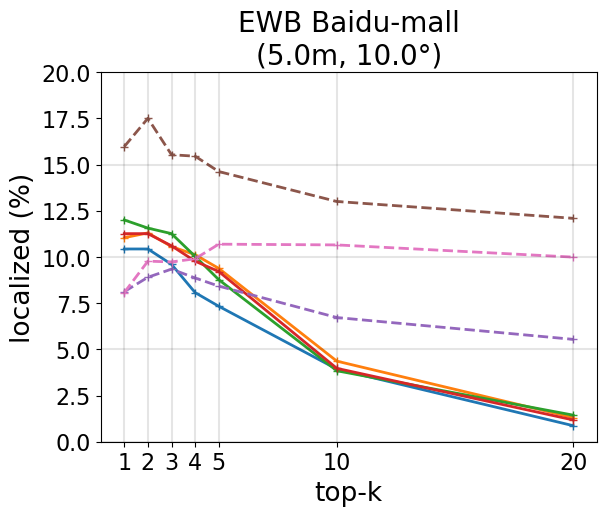}
 \includegraphics[width=0.155\textwidth]{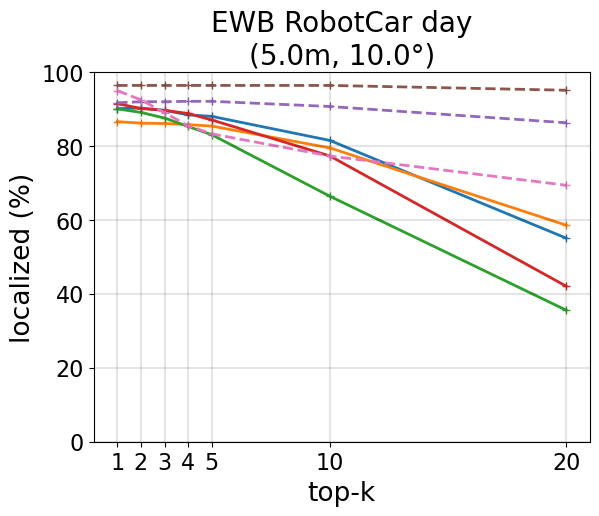}
 \includegraphics[width=0.155\textwidth]{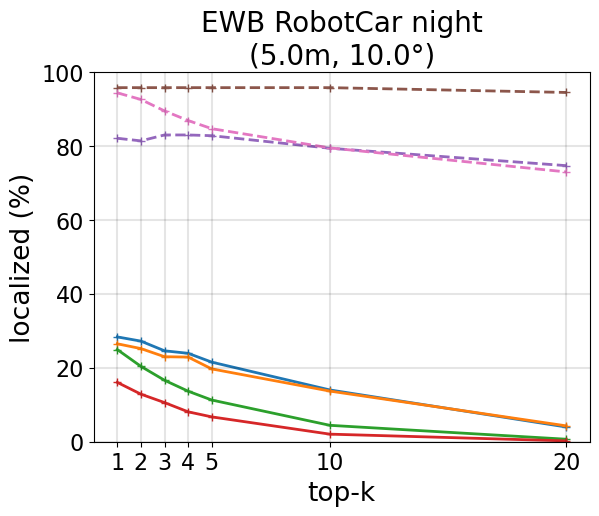}
 \\
 \includegraphics[width=0.15\textwidth]{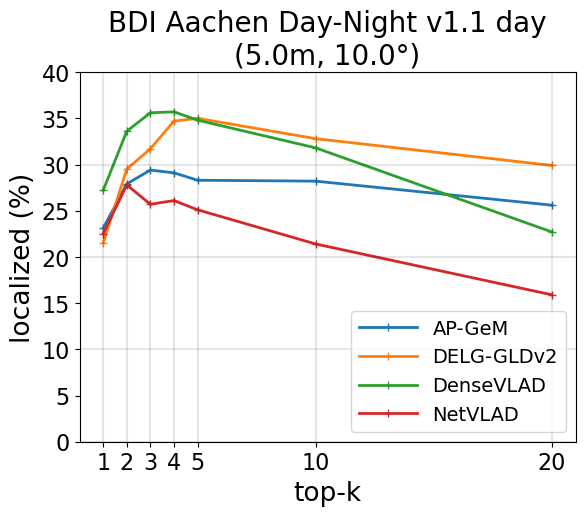}
 \includegraphics[width=0.15\textwidth]{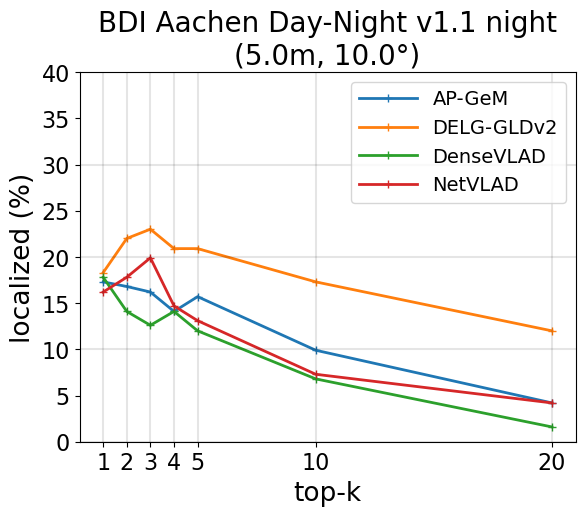}
 \includegraphics[width=0.15\textwidth]{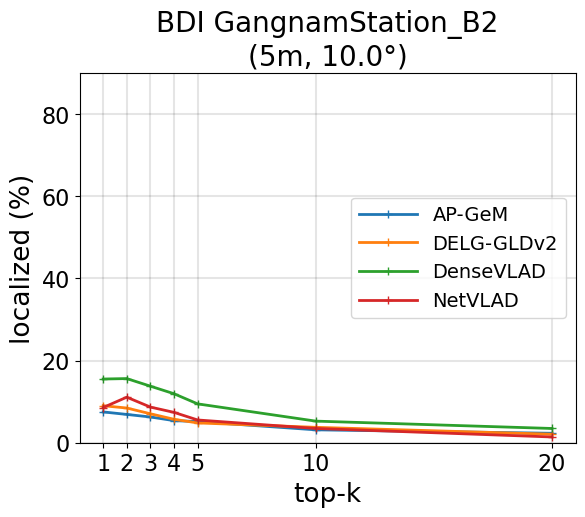}
 \includegraphics[width=0.155\textwidth]{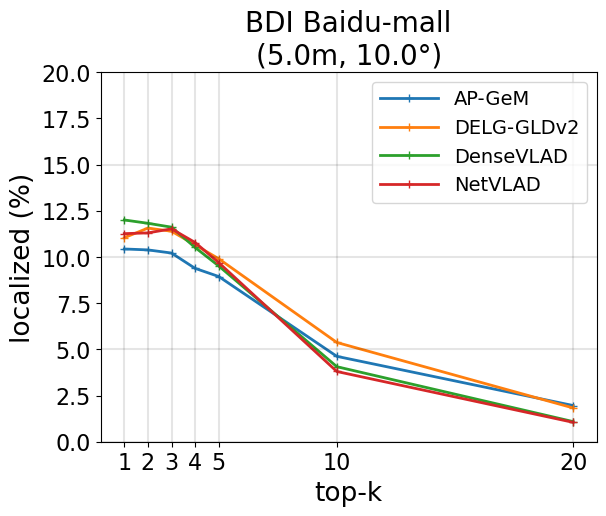}
 \includegraphics[width=0.155\textwidth]{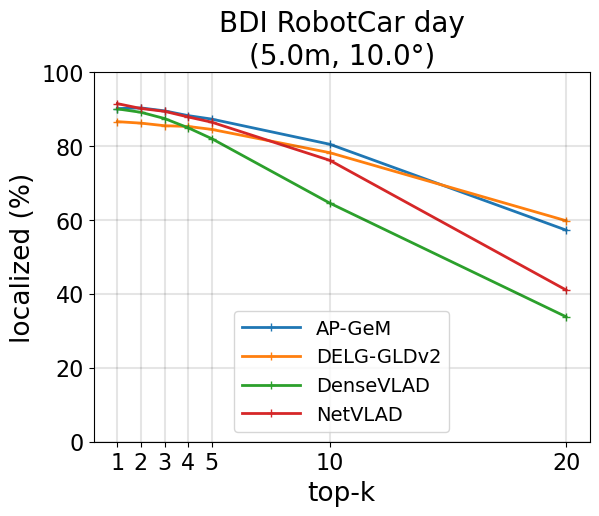}
 \includegraphics[width=0.155\textwidth]{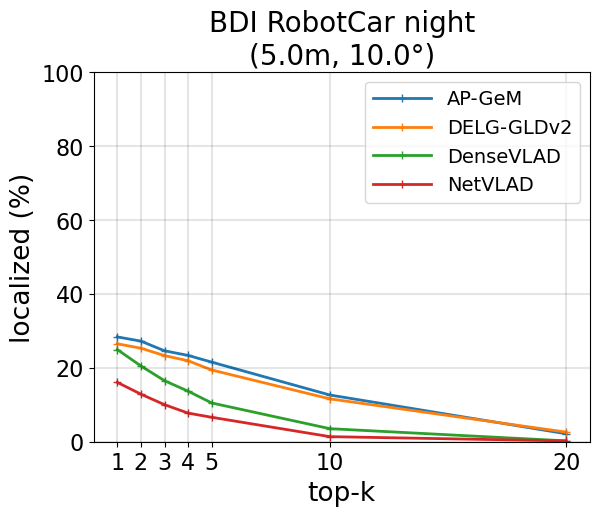}
 \\
 \includegraphics[width=0.15\textwidth]{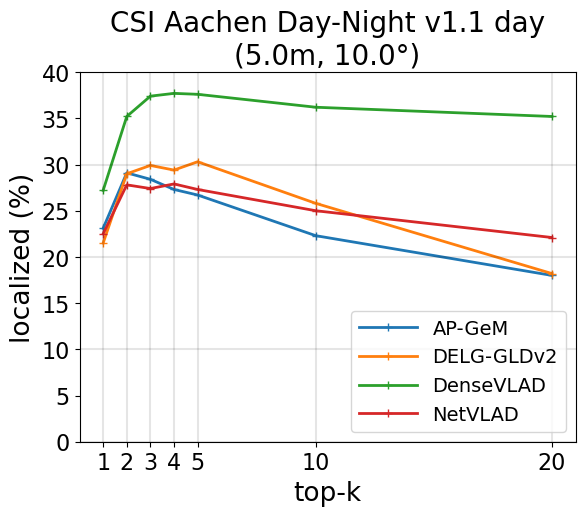}
 \includegraphics[width=0.15\textwidth]{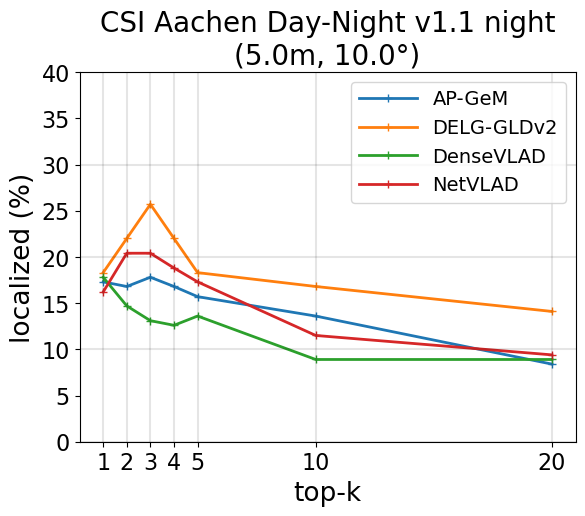}
 \includegraphics[width=0.15\textwidth]{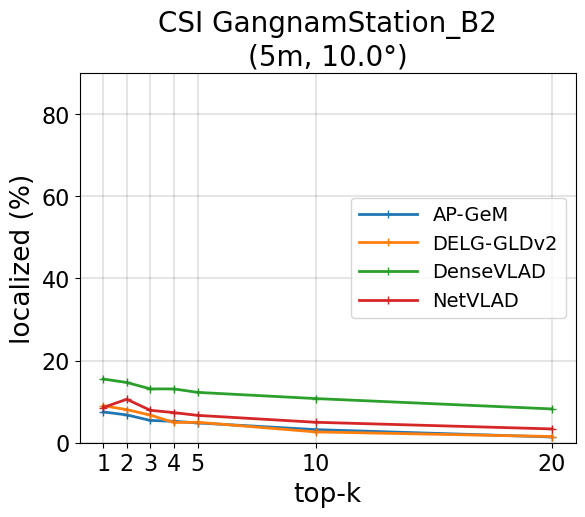}
 \includegraphics[width=0.155\textwidth]{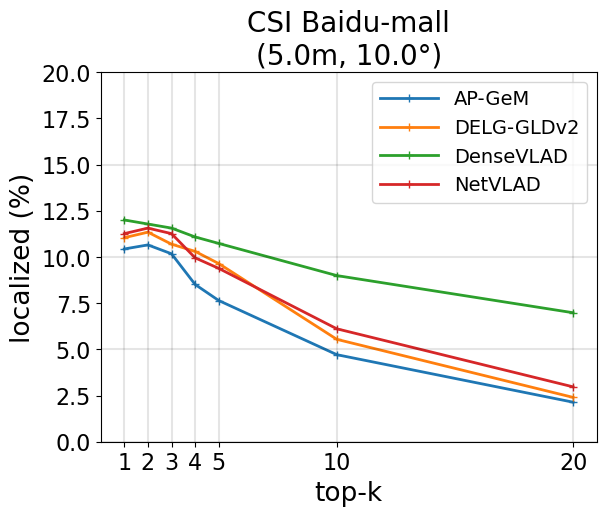}
 \includegraphics[width=0.155\textwidth]{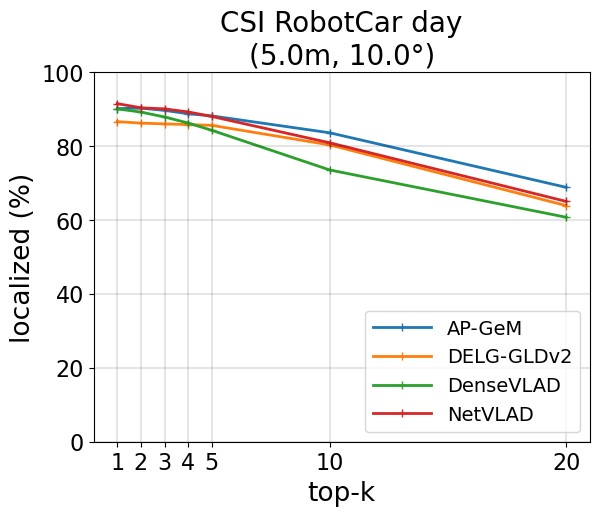}
 \includegraphics[width=0.155\textwidth]{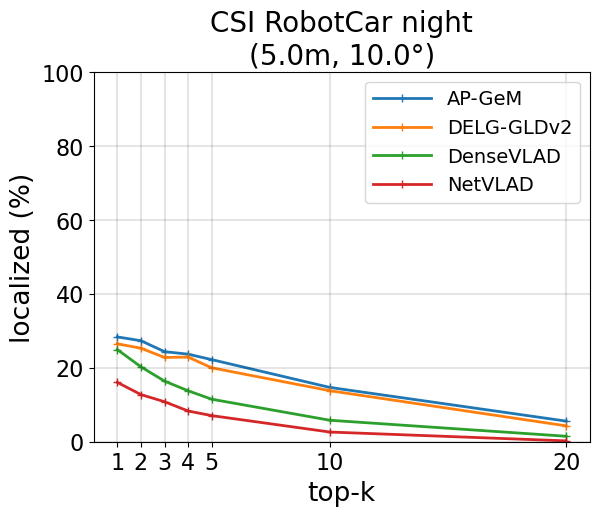}
\end{center}
   \caption{\textbf{\Task~1 (pose approximation)}. Results obtained with pose interpolation methods where the weights are obtained using EWB, BDI, and CSI (rows) for different datasets (columns). For datasets with available retrieval GT (see Sec.~\ref{sec:gt}), we show results obtained using the GT rankings (dashed lines) with EWB weighting scheme. These results can be understood as upper bounds on the localization performance.
   The best upper bound can be obtained with the distance-based ranking. The best pose approximation results are obtained with CSI and simply using the top-retrieved pose works best in many cases (except for NetVLAD and for all on Aachen). There is no clear winning global representation for all weighting schemes.}
\label{fig:exp:irbench:task1}
\end{figure*}

\subsubsection{\Task~1: Pose approximation}
\label{sec:expTask1}

Figure~\ref{fig:exp:irbench:task1} compares the four global feature types on pose approximation. 
The pose is approximated as a weighted interpolation of the top $k$ retrieved images.
As described in Sec.~\ref{sec:task1}, the weights were obtained with EWB, BDI, and CSI. 
We show the percentage of query images localized within a given error threshold with respect to the ground truth poses as a function of the number $k$ of retrieved images used for pose approximation. 
No method is able to provide highly accurate pose estimates. 
Thus, we show results for the low accuracy threshold (5m, 10$^\circ$) only.
If retrieval GT (Sec.~\ref{sec:gt}) is available, we use GT rankings to obtain upper bounds on the localization performance. 
For simplicity, we only use EWB for this because its weight computation is independent of the global feature type. 

\PAR{Comparison between interpolation methods}
As we can observe, BDI and CSI often perform slightly better than EWB (except for $k=1$ where all methods simply infer the pose of the top retrieved database image).
This is because EWB uses the same weight for each of the top $k$ retrieved images, while  BDI and CSI give more weight to higher ranked images. 
They thus assume certain correlation between the descriptor and pose similarity, an assumption that often holds in our experiments.
Surprisingly, the simpler CSI method, where the weights rely only on the similarity score between the features, performs slightly better than the more complex BDI scheme.

\PAR{Comparison between global representations} 
Interestingly, retrieving $k>1$ (up to $k=4$) images systematically improves the performance only on the Aachen dataset, most likely caused by the larger pose difference between query and reference images for this dataset (compared to the others). 
For the other datasets we observe such gain only occasionally for some of the features (especially NetVLAD) and in general there is no clear winning global representation for these methods.
DenseVLAD performs best on Aachen day and GangnamStation\_B2 since it relies on SIFT features and, thus, tends to retrieve more images taken from similar poses (SIFT is less robust to viewpoint changes than learned descriptors). 
However, DenseVLAD is also less robust to illumination changes and, as a consequence, performs poorly on the nighttime queries of RobotCar and Aachen.
DELG-GLDv2 outperforms the others on Aachen night and Baidu-mall (except for CSI) while AP-GeM performs best on RobotCar (day and night).
We attribute this to the fact that both descriptors are trained with large outdoor datasets that contain day and night images.

\PAR{Comparison between GT upper bounds}
Figure~\ref{fig:exp:irbench:task1} shows that different GT definitions yield different upper bounds on the localization performance. 
Since the EWB scheme is used, this suggests that the top ranked images differ  significantly between the rankings.
For this \task, the best upper bound can be obtained with the distance-based GT ranking. 
This was expected because, by definition of the ranking, the training image ranked first is the best possible image (\ie, the image with the pose closest to the query) for inferring its pose. 
Furthermore, we can see that the co-observation-based method provides a better upper bound than using the frustrum-overlap for pose approximation.
This can be explained by the fact that it also ranks images that are close to the query higher than frustum-overlap does.
Finally, we can see that in Baidu-mall only about 15\% of query images have a corresponding database image within the accuracy threshold, making this dataset extremely challenging for pose interpolation methods. 

\PAR{Discussion} Increasing the number of retrieved images improves the localization accuracy only until about $k = 3$. When further increasing $k$, less relevant images are added and the localization accuracy most often drops due to the increased risk of adding non-relevant poses to the interpolation scheme. 
This drop is most critical for EWB because all images are treated equally, thus it is very sensitive to outliers (even a single one). 
BDI and CSI can better cope with unrelated images (lower drop in Fig.~\ref{fig:exp:irbench:task1}), still the accuracy decreases with $k > 3$. 
We can first conclude that $k = 3$ is the sweet spot for the datasets used for our experiments. 
The evaluation of pose approximation with our ground truth rankings in Fig.~\ref{fig:exp:irbench:task1} shows that the localization accuracy increases, stays the same or only slightly decreases for $k > 3$.
Since this is not the case when using global features, the second conclusion is 
that the ranking produced by the chosen global features is not well-suited for interpolation-based pose approximation.
This shows that learning representations tailored to pose approximation, instead of using off-the-shelf methods trained for landmark retrieval, is an interesting research direction. 
In particular, our results with different GT ranking schemes show that there is room for considerable improvement.

\subsubsection{\Task~2a: Accurate pose estimation with local SFM}
\label{sec:expTask2a}

\begin{figure}[t!]
\begin{center}
 \includegraphics[width=0.23\textwidth]{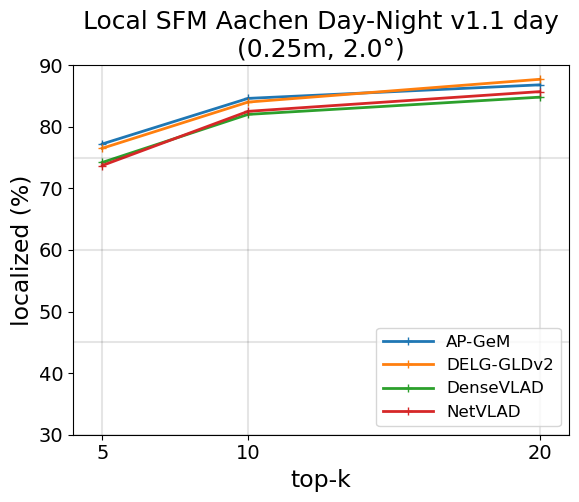}
 \includegraphics[width=0.23\textwidth]{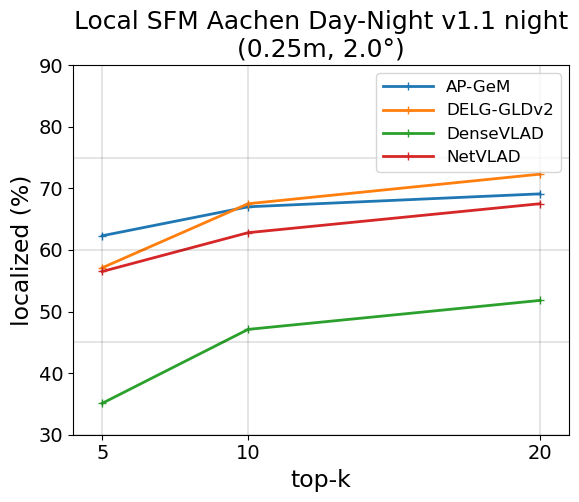} \\
 \includegraphics[width=0.23\textwidth]{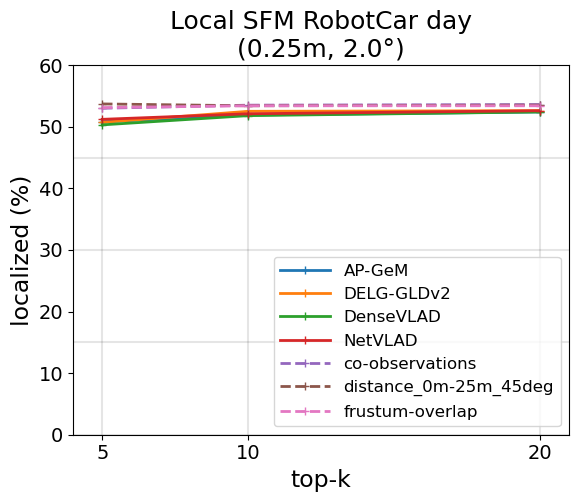}
 \includegraphics[width=0.23\textwidth]{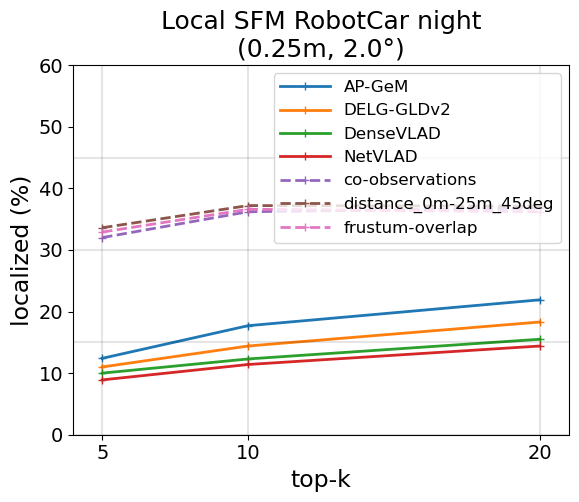} \\
 \includegraphics[width=0.23\textwidth]{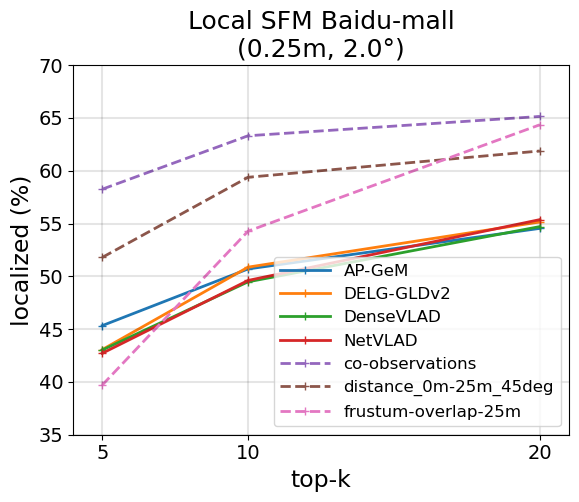}
 \includegraphics[width=0.23\textwidth]{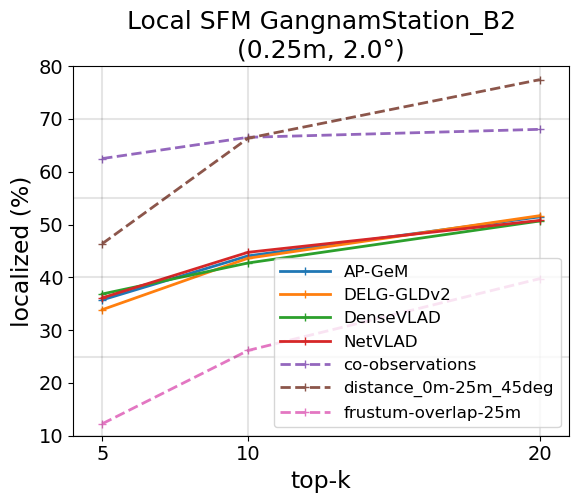} 
\end{center}
 \vspace{-0.3cm}
 \caption{\textbf{\Task~2a (pose estimation without a global map)}. We show the percentage of images localized within the high accuracy threshold as a function of $k$ retrieved images. For datasets with available retrieval GT (see Sec.~\ref{sec:gt}), we show it as upper bound (dashed lines). Best performance is obtained with the co-observation based GT ranking. The global features perform similarly on GangnamStation\_B2, Baidu-mall, and RobotCar day, while on RobotCar night and Aachen (day and night), AP-GeM and DELG-GLDv2 (due to their extensive training data) outperform NetVLAD and DenseVLAD.}
\label{fig:exp:irbench:task2a}
\end{figure}

In Figure~\ref{fig:exp:irbench:task2a}, we compare the four global feature types on accurate pose estimation by constructing an SFM map on-the-fly using the top $k$ (5, 10, 20) retrieved training images, as described in Sec.~\ref{sec:task2a}.
Since the goal of \Task~2a and \Task~2b is accurate pose estimation, we show the percentage of query images localized within the high accuracy threshold (0.25m, 2$^\circ$).
As for \Task~1, in order to provide an upper bound, we show the localization results obtained with the local map built from the top $k$ images retrieved from our GT rankings. 

\PAR{Comparison between global representations}
We observe that AP-GeM typically performs best on this \task, followed by DELG-GLDv2.
We attribute this to the fact that AP-GeM was trained with data augmentation simulating strong illumination changes.
Furthermore, we can see that learned descriptors can better cope with day-night changes of RobotCar and Aachen than DenseVLAD (except NetVLAD on RobotCar night).

\PAR{Comparison between GT upper bounds} 
In general the co-observation-based GT ranking gives a slightly better upper bound then the distance-based ranking and both perform better than the frustrum-overlap-based ranking.
Co-observations of 3D points is an important criteria for successful SFM, thus, this ranking is most suited for local SFM (essential component of this \task).
The GT rankings perform similar on RobotCar because there are no large viewpoint changes in this dataset.
This means that the frustum overlap is dominantly defined by the distance between the images (which is also the main criteria for the distance-based ranking).
Finally, co-observations (from an SFM map) are only possible if the images are close enough for successful local feature matching.

The big gaps between the results obtained with these GT rankings and the ones obtained with global features suggest that off-the-shelf methods trained for landmark retrieval are sub-optimal solutions for local SFM-based accurate pose estimation. 
Hence, this opens possibilities for research to design improved representations for this \task. 
Moreover, our results suggests that both co-observation-based and distance-based definitions could be used to train retrieval representations specifically for visual localization via approaches that learn to reproduce these rankings.

\PAR{Discussion} Since the retrieved images are used to first create a local SFM map where the query image is then registered, retrieving more images (as long as there are enough relevant ones among them) leads to a better local map and, thus, to higher localization accuracy. 
While this finding is expected, it is not obvious how many images are needed to maximise the localization accuracy or at least to approach the localization accuracy achievable when using a global map.
For local SFM, it is important to find a sweet spot for $k$ because using too many images, first, increases the chance of adding non-relevant images, and second, leads to an increase of processing time. Note however that, while the increase of the number of image pairs to be considered is quadratic, the matching between the map images can be pre-computed and only the triangulation to obtain the 3D map and the matching between query and retrieved map images is necessary to be done on-the-fly.
Furthermore, matching of the map images can even be skipped as those matches can be derived from the query-map pairs. In detail, if pixel $q1$ from the query image matches with pixels $m1$ and $m2$ from different map images, \ie there exist pixel pairs $(q1,m1)$ and $(q1,m2)$, the pixel pair $(m1,m2)$ is a natural consequence.   
While the sweet spot for $k$ depends on the dataset and the density of the map images, we found that $k = 10$ is a good overall compromise between accuracy and computation cost.

\subsubsection{\Task~2b: Accurate pose estimation with global SFM}
\label{sec:expTask2b}

\begin{figure*}[t!]
\begin{center}
 \includegraphics[width=0.24\textwidth]{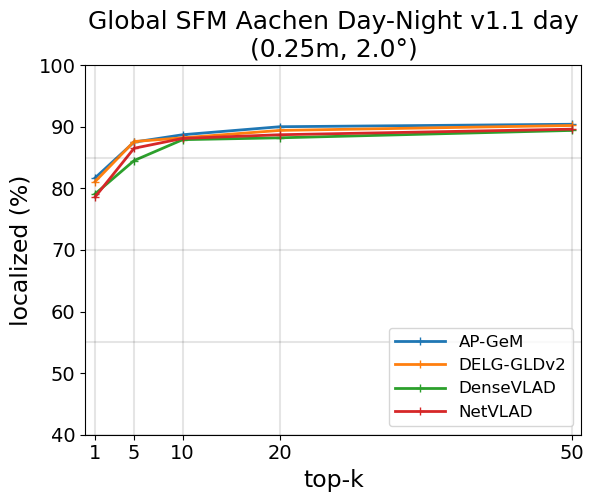}
 \includegraphics[width=0.24\textwidth]{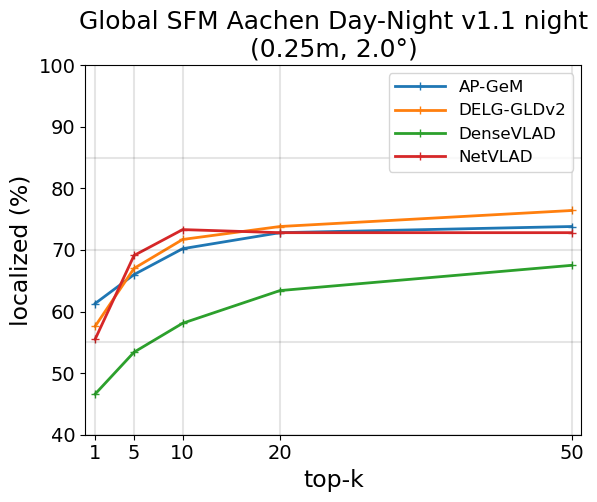}
 \includegraphics[width=0.24\textwidth]{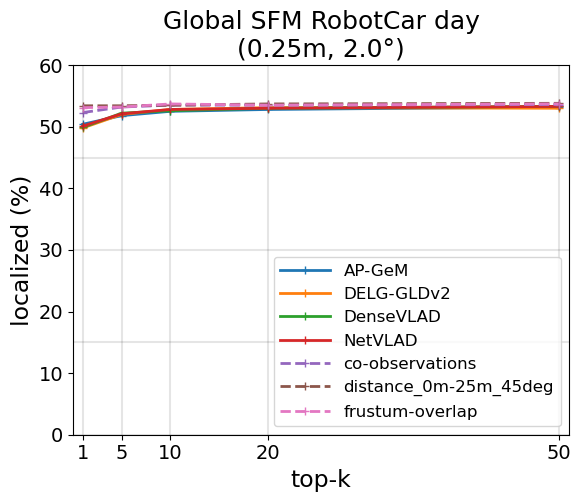}
 \includegraphics[width=0.24\textwidth]{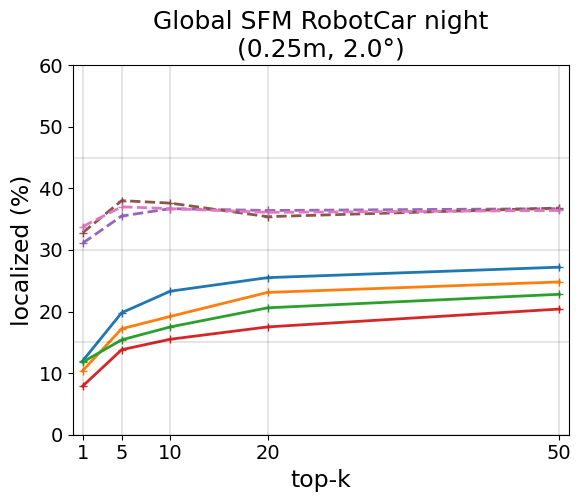}
 \\
 \includegraphics[width=0.24\textwidth]{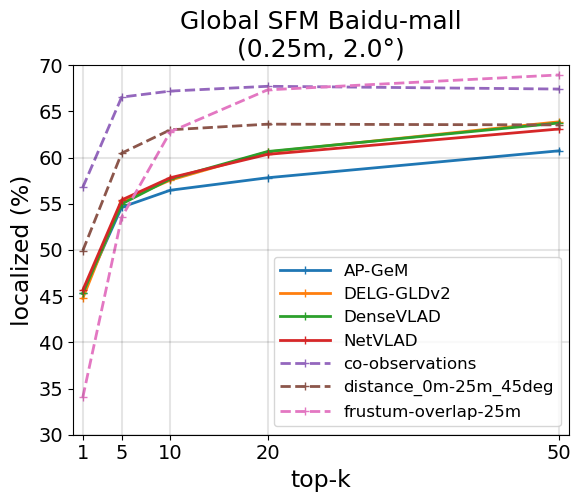}
 \includegraphics[width=0.24\textwidth]{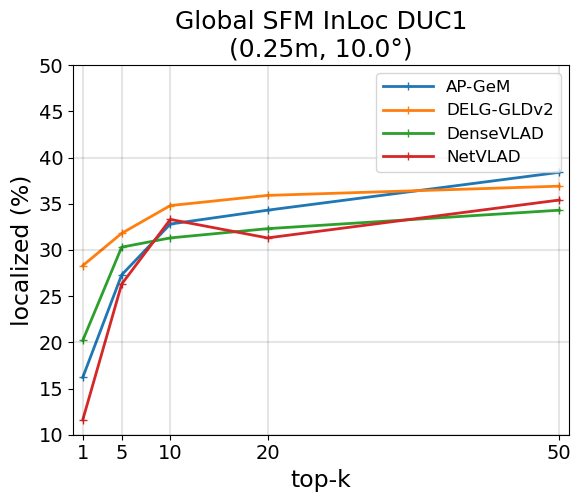}
 \includegraphics[width=0.24\textwidth]{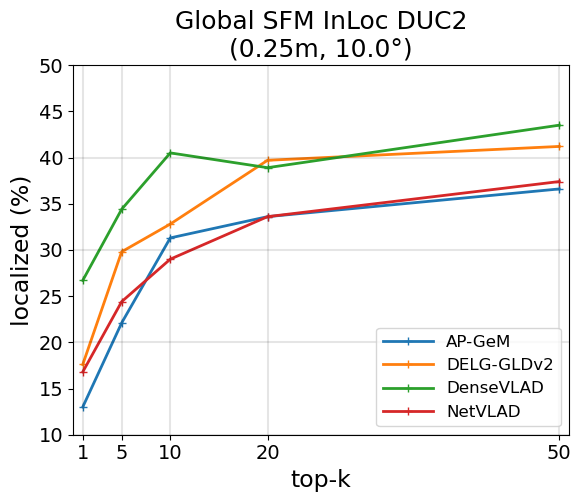}
 \includegraphics[width=0.24\textwidth]{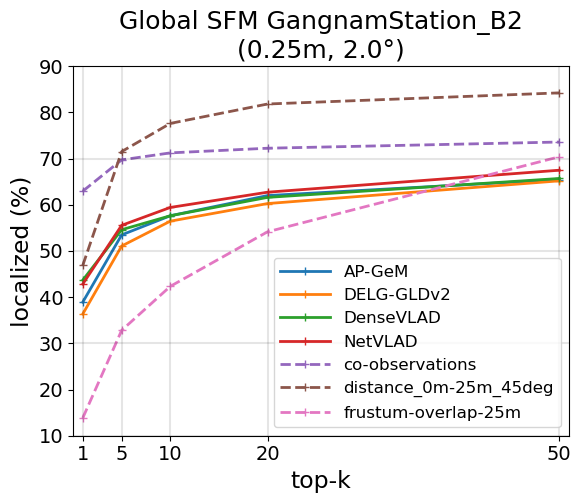} 
\end{center}
\vspace{-0.3cm}
   \caption{\textbf{\Task~2b (pose estimation with a global map)}. Percentage of images localized within the high accuracy threshold as a function of $k$ retrieved images. 
   For datasets with available retrieval GT (see Sec.~\ref{sec:gt}), we show it as upper bound (dashed lines). We observe that all representations perform similarly on Aachen day, RobotCar day, GagnamStation\_B2, and Baidu-mall but AP-GeM and DELG-GLDv2 handle the nighttime queries (Aachen night and RobotCar night) better than NetVLAD and DenseVLAD.}
\label{fig:exp:irbench:task2b}
\end{figure*}

In Fig.~\ref{fig:exp:irbench:task2b}, we compare the four global feature types on accurate pose estimation by registering the query image within a global SFM map (see Sec.~\ref{sec:task2b}).
As for \Task~2a, we report results using the high accuracy threshold (0.25m, 2$^\circ$) and, to provide an upper bound, we show the localization results obtained when registering the top $k$ (GT) images. 

\PAR{Comparison between global representations}
While we do not see a clear winner for this \task, the observation that the learned descriptors AP-GeM and DELG-GLDv2 perform best on night images, hold for this \task~as well.
Here we also show results on the more challenging InLoc dataset\footnote{For InLoc, the viewpoint difference between the reference images is too large to allow robust feature matching and point triangulation. Using the available depth maps to obtain the 3D points for all features for local SFM is identical to the way we perform global SFM on InLoc. That is why for InLoc, we do not show results for local SFM.} using the same pipeline.
Surprisingly, on DUC2 (one of the two buildings where the dataset was acquired) DenseVLAD outperforms the learned features, especially for small $k$s. 
Also, DELG-GLDv2 performs better than AP-GeM and NetVLAD on InLoc.
The main reason is probably that in the case of InLoc there are many repetitive structures and content-wise unrelated training images can be very similar (similar corridors and class-rooms).
Hence it is important to consider local details instead of global structures, as is done with DenseVLAD and DELG-GLDv2.
To capture more details, DELG-GLDv2 was jointly trained with its local keypoints (not used because outperformed by R2D2~\cite{RevaudNIPS19R2D2ReliableRepeatableDetectorsDescriptors} and D2-Net~\cite{DusmanuCVPR19D2NetDeepLocalFeatures}). 

\PAR{Comparison between GT upper bounds}
There is no clear winner between co-observation and distance-based ranking. 
The fact that they both perform better than the frustrum-overlap ranking can be explained by the nature of the \task.
On the one hand, many co-observations (note that they were computed using an SFM map in the first place) very likely yield better global SFM-based localization, on the other hand, local feature matching is more likely to succeed with spatially close images (as are obtained by the distance-based GT).
Furthermore, contrary to distance-based, our frustum-based GT ranking is not limited by a maximum angular difference, \ie, there can be images with a large frustum overlap taken from very different directions.
This means that a large frustum-overlap with a large viewpoint change would still rank high, even if local feature matching might not work anymore.
The gap between the upper bounds and the performance obtained with global representations is smaller than for \Task~2a but it still opens possibilities for research to improve image retrieval for this \task, especially for day-night changes. 

\ccc{\PAR{Discussion} In order to maximize the localization accuracy for this \task, in theory, only very few relevant images are needed. 
Our experiments show that these relevant images are not ranked first using current image representations (see the experiments with ground truth rankings in Fig.~\ref{fig:exp:irbench:task2b}).
Instead, while this also depends on the dataset and the density of mapping images, we need to retrieve about $k = 20$ images in order to achieve top performance.
Further increasing $k$ helps a bit but comes with the cost of additional processing time and, as can be seen in Fig.~\ref{fig:exp:irbench:task2b}, is probably not worth it in practice. 
Ranking with better suited image representations would decrease the required amount of $k$ and, thus, lead to faster processing and more accurate localization.}

\subsection{Evaluating correlation between retrieval and localization metrics}
\label{sec:CorrExp}

In this section, and as a core question of this paper, we analyze the correlation between visual localization and image retrieval metrics.
This is motivated by the fact that image representations often used for visual localization were originally designed for more general retrieval tasks.
Note that we only analyze this correlation for datasets with available retrieval GT (Baidu-mall, GangnamStation\_B2, and RobotCar)\footnote{For the other datasets, GT camera poses are not available for the test images, which makes it hard to generate retrieval GT.}.
Furthermore, for our analysis, we do not separate day and night images of RobotCar but analyze them jointly.
As a starting point, Tab.~\ref{tab:RoxfordRparis} shows the performance of the four image representations on the established $\mathcal{R}$Oxford ($\mathcal{R}$O) and $\mathcal{R}$Paris ($\mathcal{R}$P) landmark retrieval benchmarks using the Medium (m) and Hard (h) protocols~\cite{RadenovicCVPR18RevisitingOxfordParisImRetBenchmarking}. 
As expected, DELG-GLDv2 and AP-GeM descriptors, designed and trained for these \task, significantly outperform DenseVLAD and NetVLAD. 
However, our previous experiments have shown DenseVLAD and NetVLAD can perform similarly or better than DELG-GLDv and AP-GeM on the localization \tasks. 
This contradiction is a clear indicator of the different requirements of the localization and landmark retrieval problems. 

\subsubsection{\Task~3a: Landmark retrieval}
\label{sec:expTask1}

\begin{figure}[t!]
\begin{center}
 \includegraphics[width=0.23\textwidth]{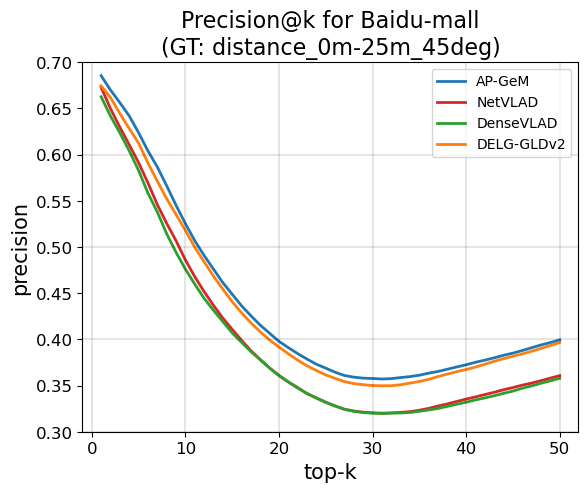}%
 \includegraphics[width=0.23\textwidth]{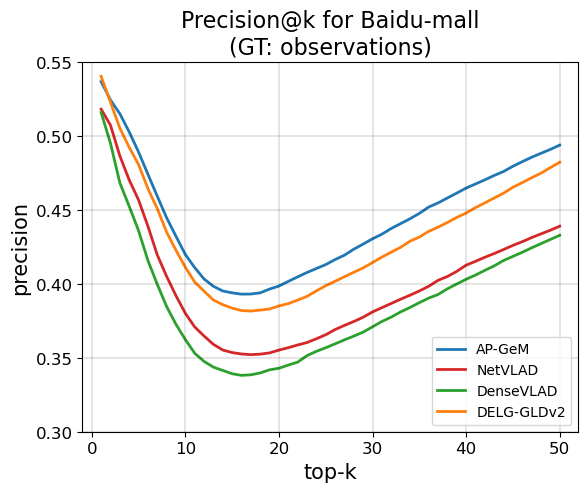}\\
 \includegraphics[width=0.23\textwidth]{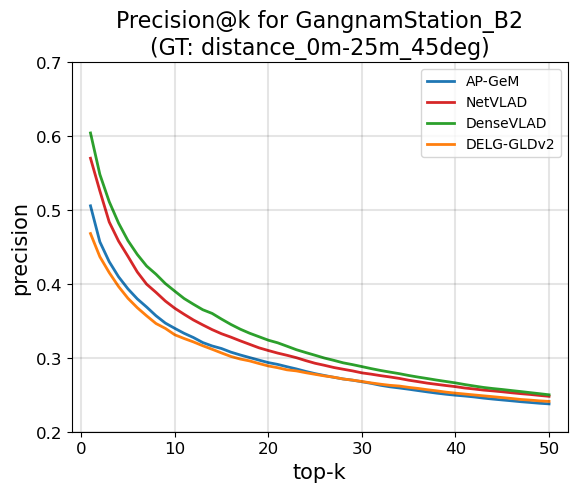}%
 \includegraphics[width=0.23\textwidth]{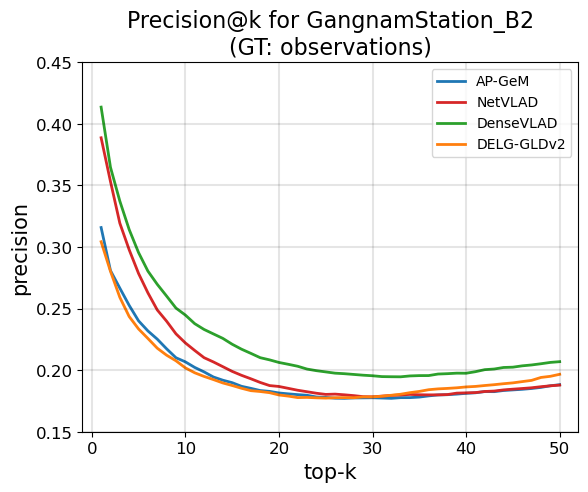} \\
 \includegraphics[width=0.23\textwidth]{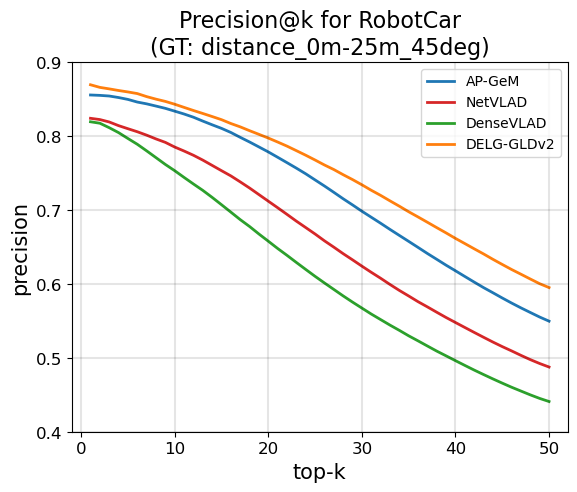}%
 \includegraphics[width=0.23\textwidth]{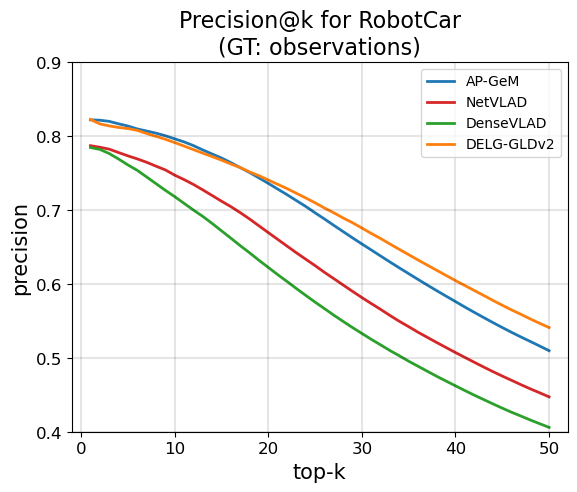}
\end{center}
   \caption{\textbf{Landmark retrieval (Precision@k)}. Landmark retrieval performance measured with Precision@$k$ (P@$k$) where we use the distance and (left column) and co-observation-based (right column) GT. While the curves differ depending on the GT used, the ranking of the four global representations is unchanged for each dataset. There is no feature that performs best on all the datasets but AP-GeM and NetVLAD are often among the top.}
\label{fig:landmarkretrieval}
\end{figure}

In order to evaluate landmark retrieval, the literature~\cite{PhilbinCVPR07ObjectRetrievalFastSpatialMatching,PhilbinCVPR08LostInQuantization,ToliasPR14VisualQueryExpansionFeatureAggregation,RadenovicCVPR18RevisitingOxfordParisImRetBenchmarking} suggest to use precision at $k$ (P@$k$), which measures the percentage of relevant images among the retrieved $k$ images.
In Fig.~\ref{fig:landmarkretrieval}, we show P@$k$ for all datasets with available retrieval GT.
We only compare distance (left) and co-observations-based (right column) GT rankings since they perform better than using frustum-overlap (see Sec.~\ref{sec:locExp}). 

Ideally, P@$k$ would be high (many relevant images among the top $k$) and constant. 
However, as can be seen in the figure, instead it decreases rapidly.
Interestingly, when increasing $k$ on Baidu-mall, P@$k$ rises again.
This shows that for this dataset, the global descriptors start to find more relevant images when $k$ is larger.
This can also be seen in Fig.~\ref{fig:exp:irbench:task2b} where the localization performance increases with $k$.

\begin{table}[]
 \begin{center}
  \caption{Performance evaluation (mAP) on $\mathcal{R}$Oxford ($\mathcal{R}$O) and $\mathcal{R}$Paris ($\mathcal{R}$P) for the Medium (m) and Hard (h) protocols.}
 \label{tab:RoxfordRparis}
\resizebox{0.9\linewidth}{!}{
\input{IJCV_plots/table_roxrpar}
}
\end{center}
\end{table}

\PAR{Comparison between global representations}
First we observe that, while the curves differ depending on the GT used, the ranking of the four global representations is unchanged for each dataset.
There is no feature that performs best on all the datasets but AP-GeM and DELG-GLDv2 are often among the top, except GangnamStation\_B2 where both perform worst.

\begin{figure*}[t]
\begin{center}
 \includegraphics[width=\widthfivefigs\textwidth]{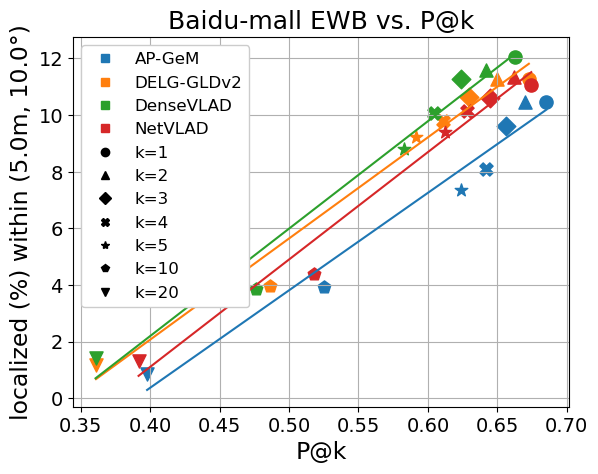}
 \includegraphics[width=\widthfivefigs\textwidth]{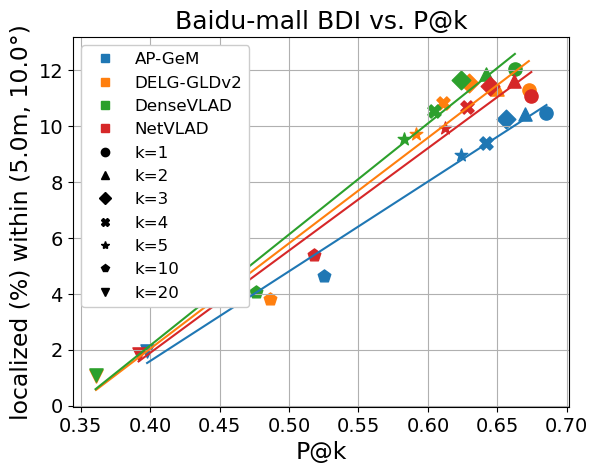}
 \includegraphics[width=\widthfivefigs\textwidth]{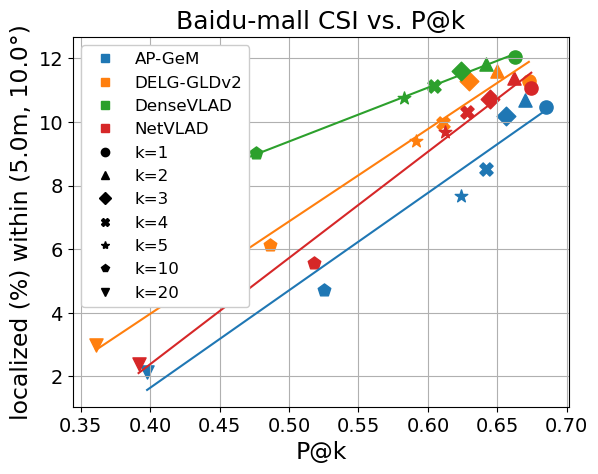}
 \includegraphics[width=\widthfivefigs\textwidth]{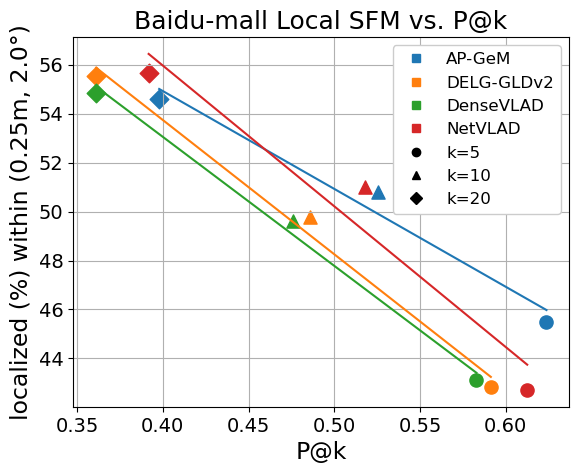}
 \includegraphics[width=\widthfivefigs\textwidth]{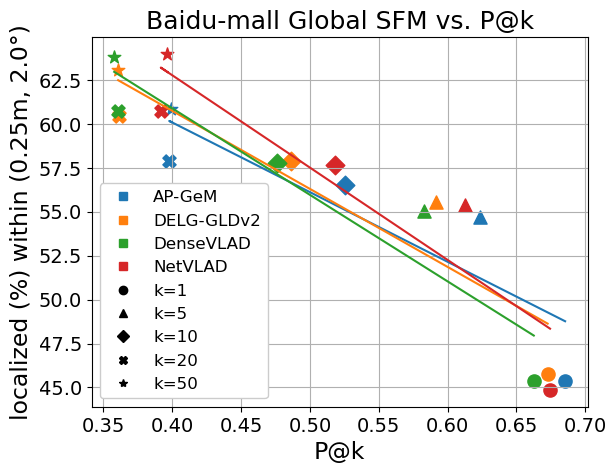}
 \\
 \includegraphics[width=\widthfivefigs\textwidth]{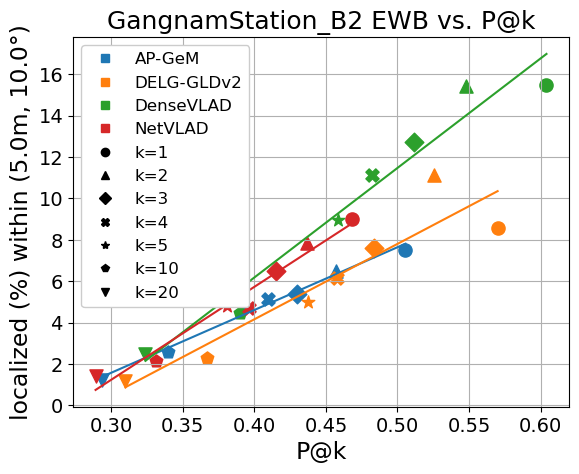}
 \includegraphics[width=\widthfivefigs\textwidth]{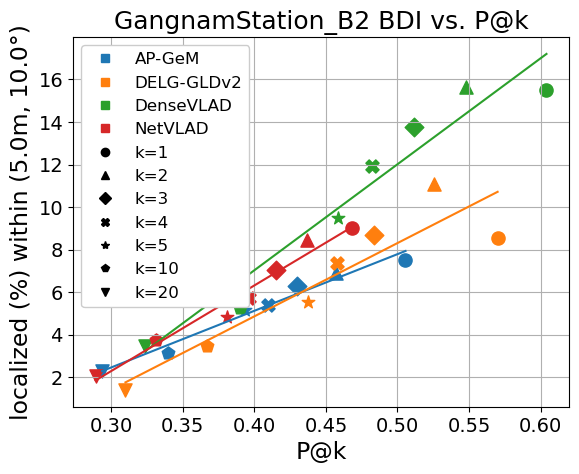}
 \includegraphics[width=\widthfivefigs\textwidth]{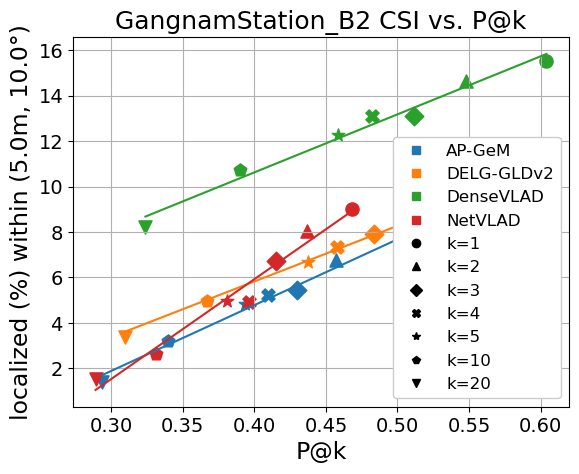}
 \includegraphics[width=\widthfivefigs\textwidth]{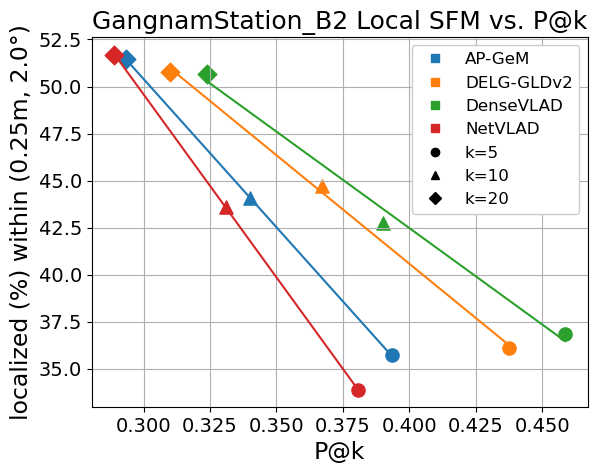}
 \includegraphics[width=\widthfivefigs\textwidth]{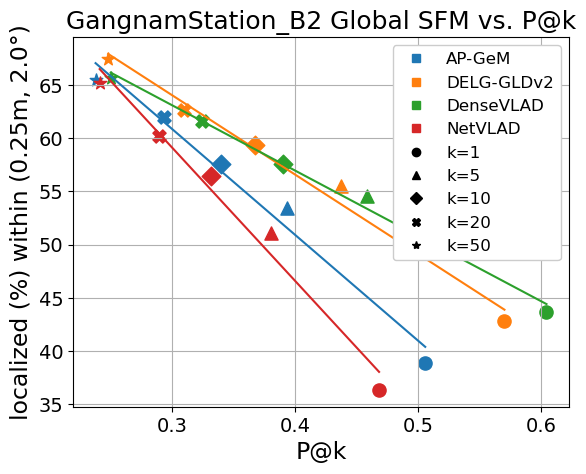}
 \\
 \includegraphics[width=\widthfivefigs\textwidth]{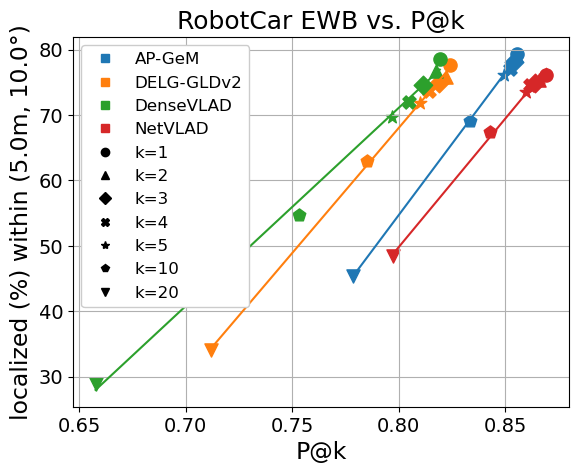}
 \includegraphics[width=\widthfivefigs\textwidth]{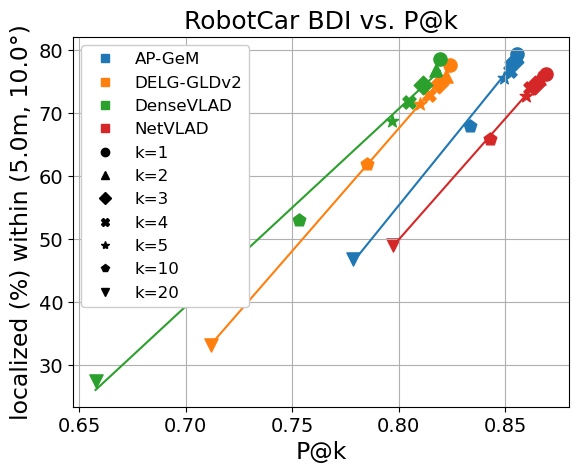}
 \includegraphics[width=\widthfivefigs\textwidth]{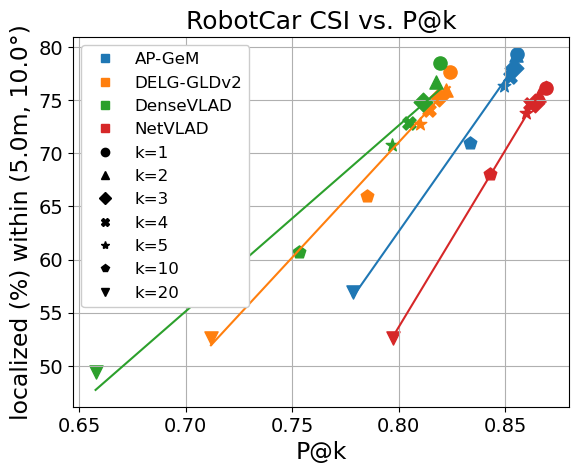}
 \includegraphics[width=\widthfivefigs\textwidth]{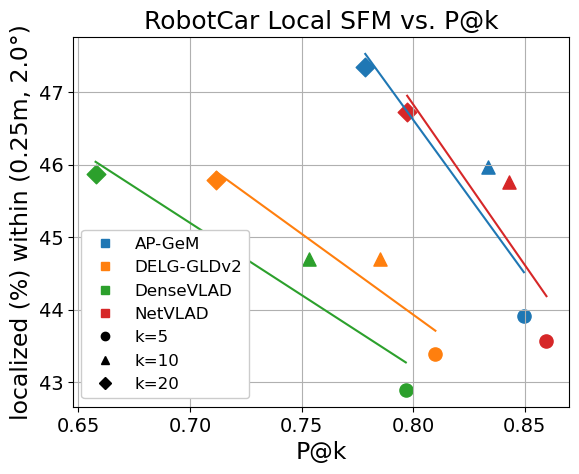}
 \includegraphics[width=\widthfivefigs\textwidth]{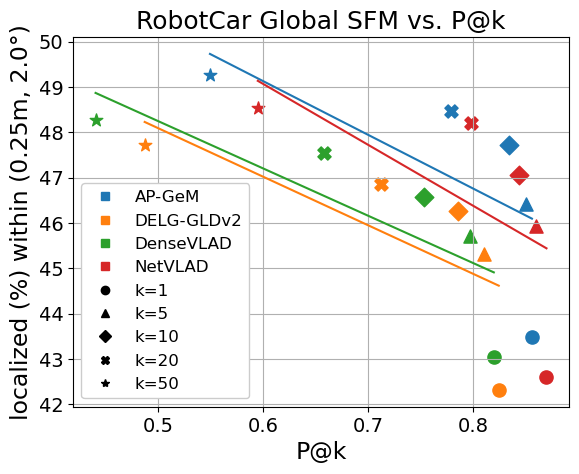}
\end{center}
\vspace{-0.3cm}
   \caption{\textbf{Landmark retrieval correlation}. Correlation between landmark retrieval (P@$k$ using distance-based GT) and the three visual localization \tasks \, (successfully localized images in \%) shown as scatter plot, one per \task \, (column), and dataset (row), where each point represents a global feature and a top $k$ value. To show whether or not the two metrics are linearly correlated, we fit a line on all top $k$ experiments for each feature type. As can be seen, there is a linear correlation for pose approximation and an inverse linear correlation for accurate pose estimation (last two columns). 
   } 
\label{fig:landmarkretrieval_scatter}
\end{figure*}

\begin{figure*}[t]
\begin{center}
 \includegraphics[width=\widthfivefigs\textwidth]{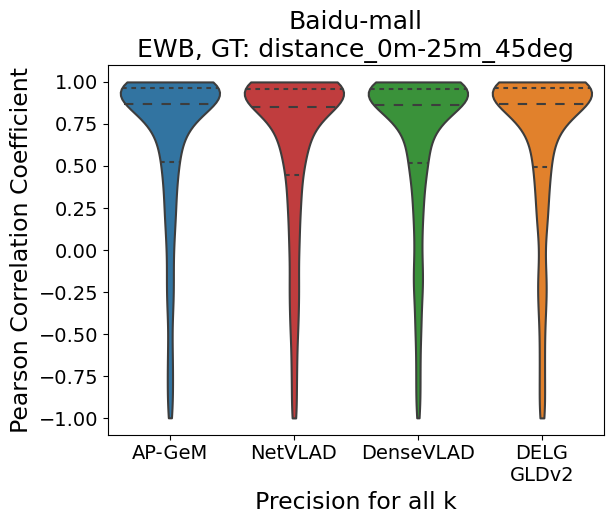}
 \includegraphics[width=\widthfivefigs\textwidth]{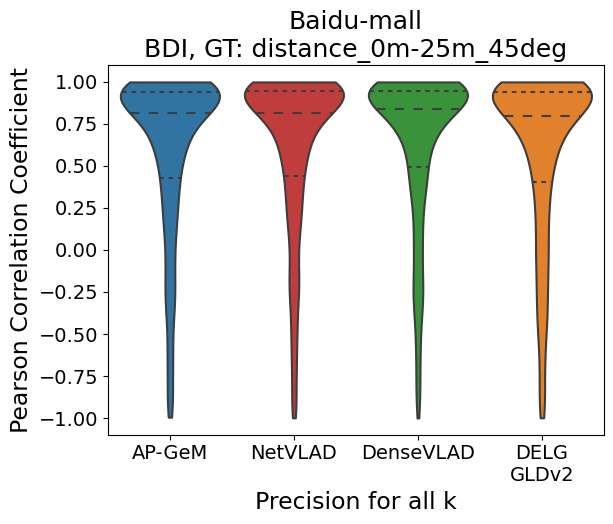}
 \includegraphics[width=\widthfivefigs\textwidth]{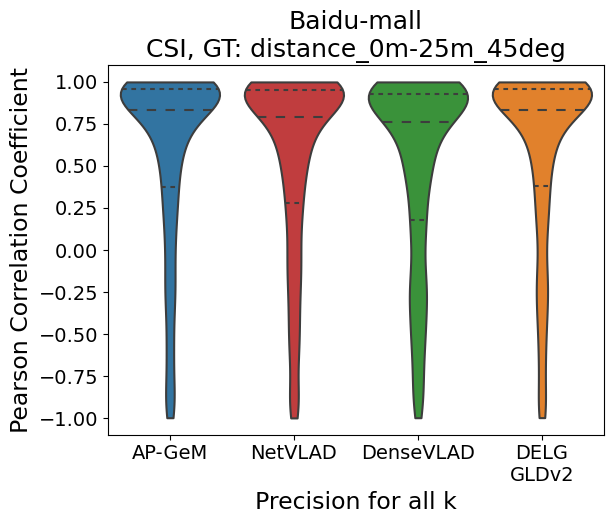}
 \includegraphics[width=\widthfivefigs\textwidth]{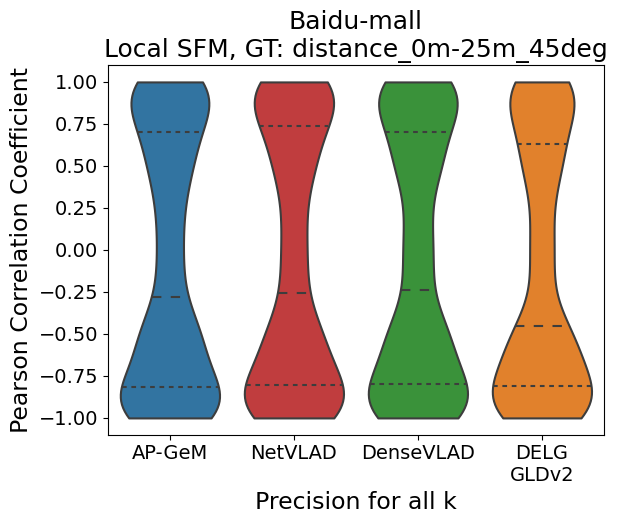}
 \includegraphics[width=\widthfivefigs\textwidth]{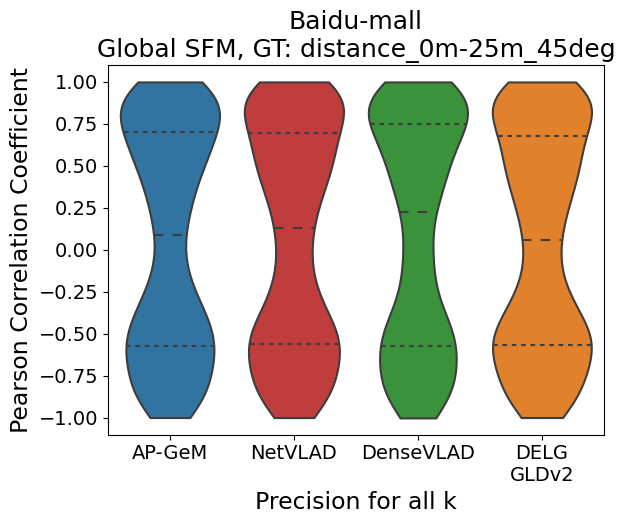} \\
 \includegraphics[width=\widthfivefigs\textwidth]{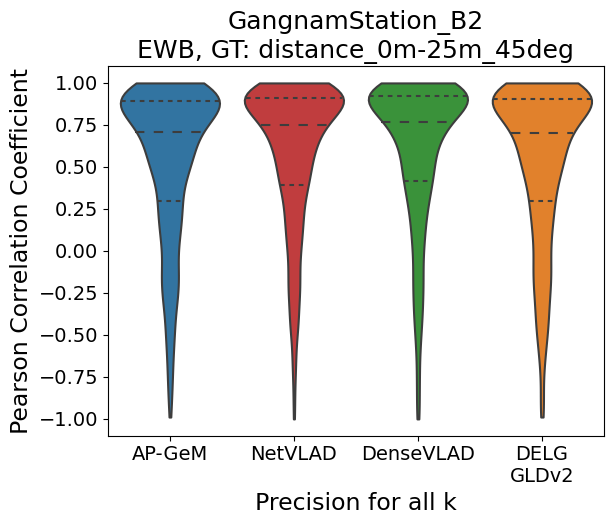}
 \includegraphics[width=\widthfivefigs\textwidth]{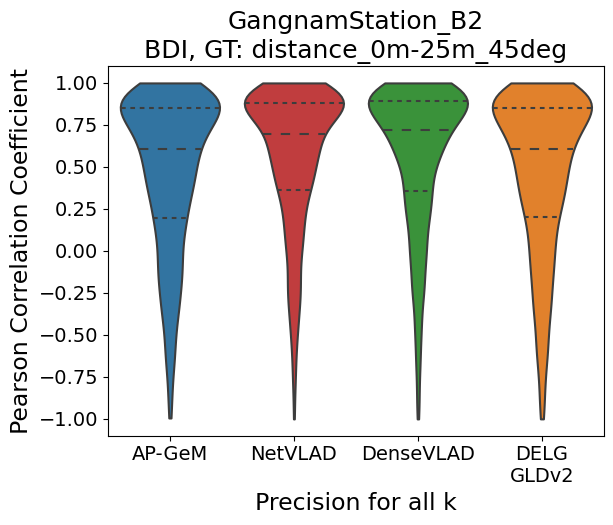}
 \includegraphics[width=\widthfivefigs\textwidth]{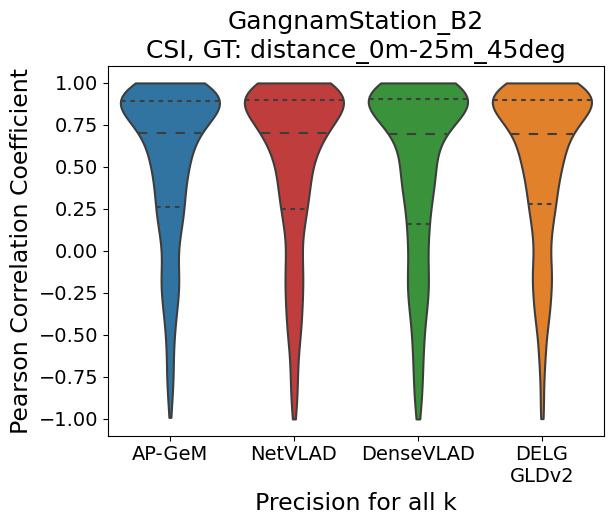}
 \includegraphics[width=\widthfivefigs\textwidth]{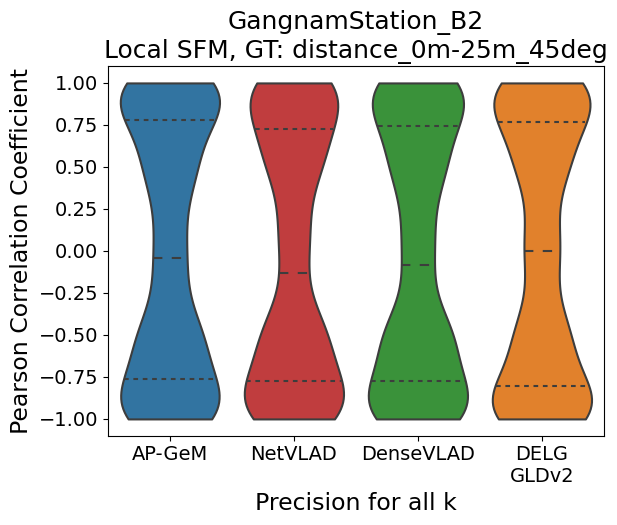}
 \includegraphics[width=\widthfivefigs\textwidth]{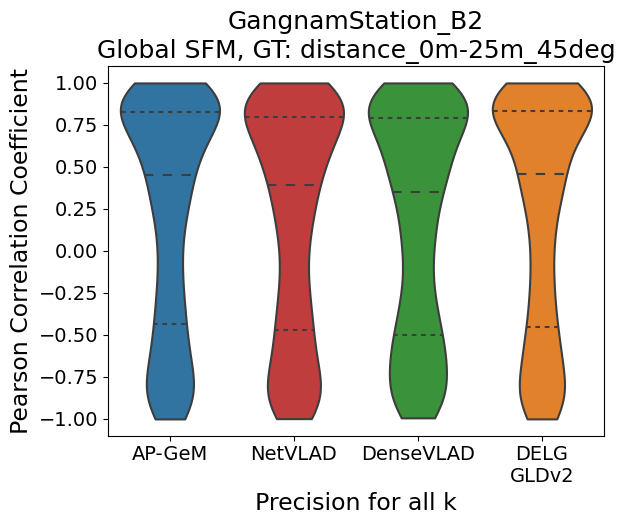} \\
 \includegraphics[width=\widthfivefigs\textwidth]{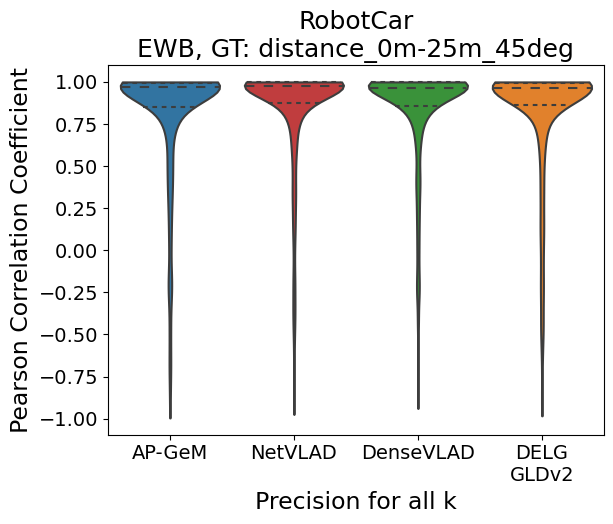}
 \includegraphics[width=\widthfivefigs\textwidth]{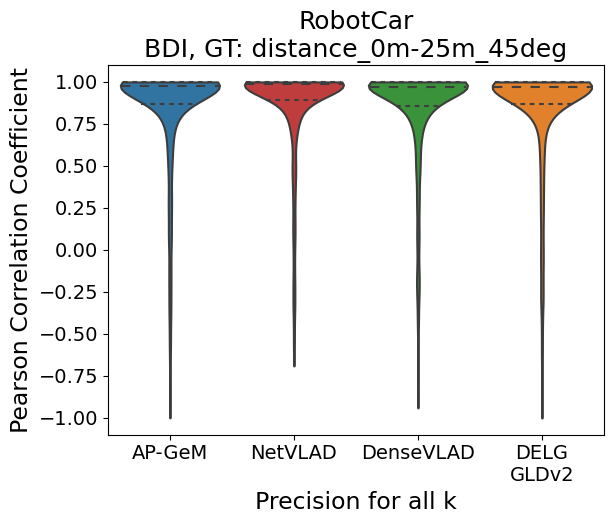}
 \includegraphics[width=\widthfivefigs\textwidth]{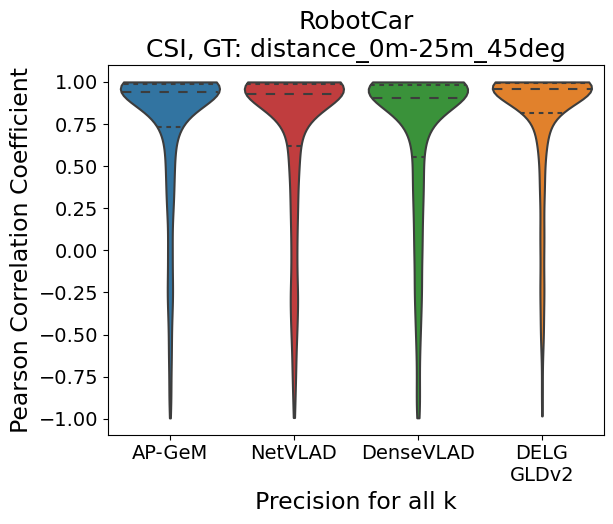}
 \includegraphics[width=\widthfivefigs\textwidth]{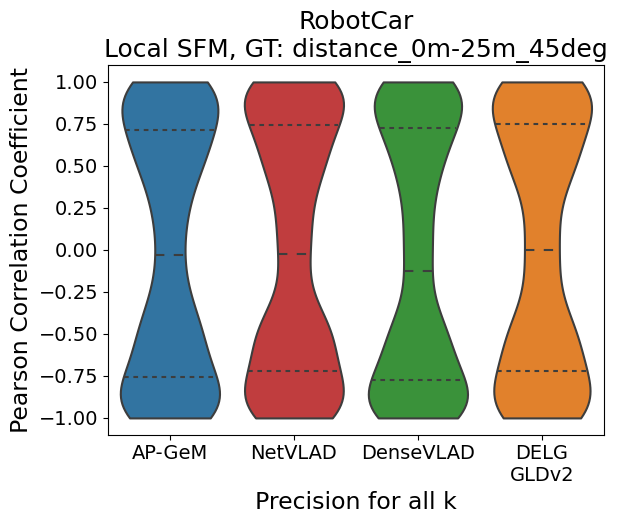}
 \includegraphics[width=\widthfivefigs\textwidth]{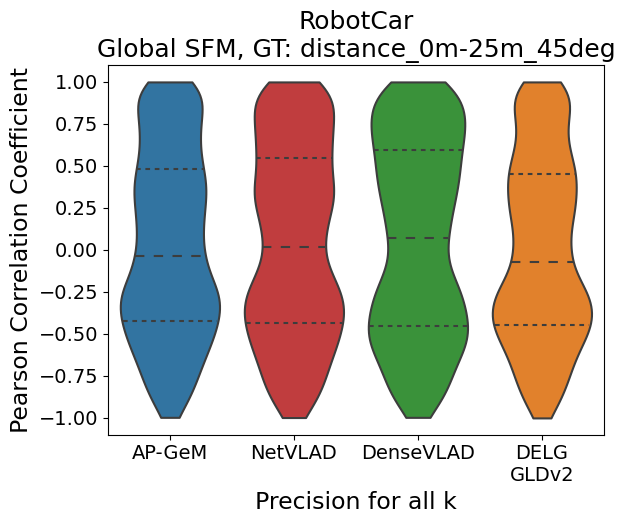}
\end{center}
\vspace{-0.3cm}
   \caption{\textbf{Landmark retrieval per image linear correlation}. Pearson coefficients computed for each query image individually (directly using the pose error as localization metric) and visualized as violin plots. The columns show the localization \tasks \, and the rows show different datasets. For pose approximation, the Pearson coefficients are densely sampled in the upper part of the violins, meaning good linear correlation, whereas for pose estimation, they are sampled on both sides and in the middle (RobotCar), meaning high, inverse, and low correlation. We can observe similar behaviour for all feature types.}
\label{fig:landmarkretrieval_violin}
\end{figure*}

\PAR{Evaluating linear correlation between retrieval and localization}
We evaluate linear correlation between landmark retrieval (P@$k$ using distance-based GT) and the three visual localization \tasks \, (successfully localized images in \%) visually using a scatter plot and numerically using the Pearson~(\ref{eq:pearson}) correlation coefficient.
Figure~\ref{fig:landmarkretrieval_scatter} shows one scatter plot per \task \, (column) and dataset (row), where each point represents a global feature and a top $k$ value.
To show whether or not the two metrics are linearly correlated, for each feature type we fit a line on all top $k$ experiments.
As can be seen, there is a linear correlation for pose approximation (\Task~1), and an inverse linear correlation for pose estimation (\Task~2).
Pose approximation is linearly correlated with precision because it benefits from retrieving more relevant images. 
Furthermore, same as for precision, its performance drops with increasing $k$.
Naturally, both accurate pose estimation \tasks \, need to retrieve relevant images.
In particular, local SFM requires multiple relevant images to construct the local map. 
However, both \tasks \, can handle irrelevant images among the top $k$, \ie, high precision is not necessarily required. 
While P@$k$ decreases with increasing $k$ (Fig.~\ref{fig:landmarkretrieval}), a larger number of relevant images are typically found for larger values of $k$. 
This explains why local SFM starts to work for $k>5$, leading to an inverse linear correlation. 
Compared to local SFM, global SFM requires a smaller number of relevant images to be among the top $k$ in order to succeed (retrieving one relevant image is already sufficient in theory). 
Thus, smaller numbers of $k$ are sufficient for global SFM, which explains the weaker inverse correlation for \Task~2b.

Numerically, these correlations can be seen with Pearson correlation coefficients close to 1 for pose approximation and low coefficients (close to -1) for accurate pose estimation. 
The Pearson coefficients computed for the entire dataset (using successfully localized images as localization metric) and for each global feature type can be found  
in Tab.~\ref{tab:landmarkretrieval_pearson}. 

To further analyze the correlation, we computed the Pearson coefficient for each query image individually (directly using the pose error as the localization metric) and visualize its distribution as violin plots in Fig.~\ref{fig:landmarkretrieval_violin}.
As for Fig.~\ref{fig:landmarkretrieval_scatter}, the columns show the localization \tasks~and the rows show different datasets. 
Each plot contains a violin per global feature type showing consistent behaviour between features. 
The most important difference to the per-dataset Pearson coefficient is that here the localization error of each image is used instead of the percentage within an accuracy bin. 
This allows a more detailed analysis:
For pose approximation, the Pearson coefficients are densely sampled in the upper part of the violins, meaning very good linear correlation, whereas for pose estimation, they are sampled on both sides and in the middle (RobotCar), meaning high, inverse, and low correlation.
This shows that actually for \Task~2 (except RobotCar), a large part of the query images is linearly correlated (when P@$k$ is high) and another large part inverse linearly (when P@$k$ is low). 
Intuitively, the former correspond to "easier" images, where retrieval is able to find many relevant images at high precision, and the latter to test images for which it is harder to find relevant images (leading to a lower precision).

\begin{table}[t!]
\center
\caption{\textbf{Landmark retrieval Pearson correlation} coefficients between P@$k$ (using distance-based GT) and localization accuracy (successfully localized images in \%) shown per datasets. Overall, we can see a very good correlation (Pearson coefficient close to 1) between P@$k$ and the interpolation-based pose approximation methods and inverse correlation (Pearson coefficient close to -1) between P@$k$ and accurate pose estimation with local or global SFM map.}
\label{tab:landmarkretrieval_pearson}
\resizebox{\linewidth}{!}{
\input{IJCV_plots/table_precision_pearson}
}
\end{table}

\PAR{Evaluating rank correlation between retrieval and localization}
In addition to linear correlation, we evaluate rank correlation between landmark retrieval and visual localization.
While linear correlation means that if one metric increases, the other one follows, rank correlation investigates if the two metrics produce the same ranking between different global features.
It evaluates if for two given \tasks \, (\eg landmark retrieval and global SFM) the performance rankings of the global features types correlate.
In detail, it answers the question: If AP-GeM $>$ NetVLAD $>$ DELG-GLDv2 $>$ DenseVLAD for precision, does this mean it is also true for global SFM?

This is an interesting measure to evaluate correlation, even if the two metrics are not strictly linear.
As ranking is hard to assess in scatter plots, we compute the Spearman~(\ref{eq:spearman}) correlation coefficient for each $k$ instead. 
We report it per dataset in Tab.~\ref{tab:landmarkretrieval_spearman} which summarizes the rank correlation between the global features in one number. 

As can also be seen from the P@$k$ plot in Fig.~\ref{fig:landmarkretrieval} and the localization results in Fig.~\ref{fig:exp:irbench:task1}, \ref{fig:exp:irbench:task2a}, and ~\ref{fig:exp:irbench:task2b}, the feature ranking of P@k and BDI (5.0m, 10.0\degree) with $k=1$ for Baidu-mall is inverse and thus  $SRC=-1.0$, (see Tab.~\ref{tab:landmarkretrieval_spearman}). 
On the other side of the spectrum, P@$k$ and BDI (5.0m, 10.0\degree) with $k=5$ for GangnamStation\_B2 have exactly the same ranking between the four feature types which results in $SRC=1.0$.

Summarizing the previous results, we see a clear linear correlation between landmark retrieval and visual localization for \Task~1. 
However, there is no rank correlation between P@k and localization accuracy (all \tasks), \ie, the performance of the features between both settings is not consistent.

\begin{table}[t!]
\center
\caption{\textbf{Landmark retrieval Spearman correlation} coefficients between P@$k$ (using distance-based GT) and localization accuracy (successfully localized images in \%) shown per datasets. We observe that overall there is no rank correlation between P@$k$ and localization accuracy. 
}
\label{tab:landmarkretrieval_spearman}
\resizebox{\linewidth}{!}{
\input{IJCV_plots/table_precision_spearman}
}
\end{table}

\subsubsection{\Task~3b: Place recognition}
\label{sec:expTask1}

\begin{figure}[t!]
\begin{center}
 \includegraphics[width=0.23\textwidth]{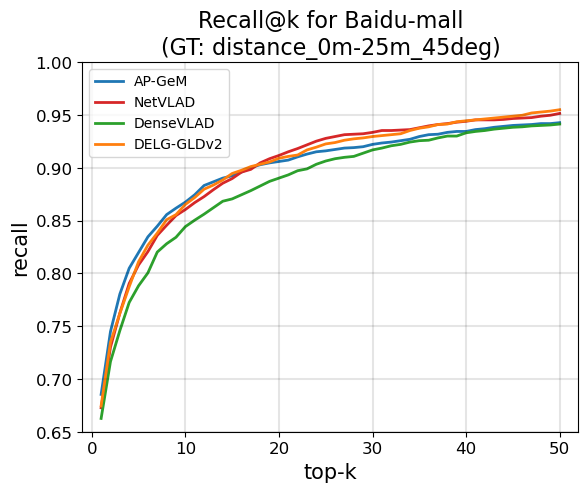}
 \includegraphics[width=0.23\textwidth]{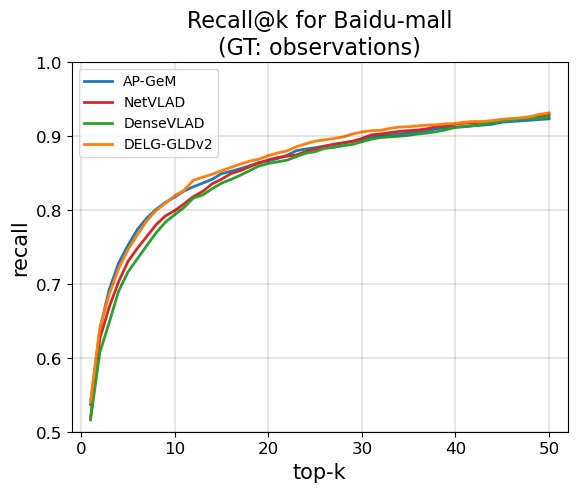} \\
 \includegraphics[width=0.23\textwidth]{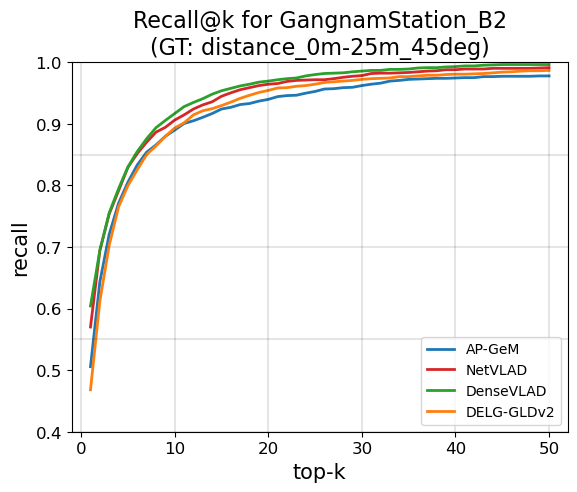}
 \includegraphics[width=0.23\textwidth]{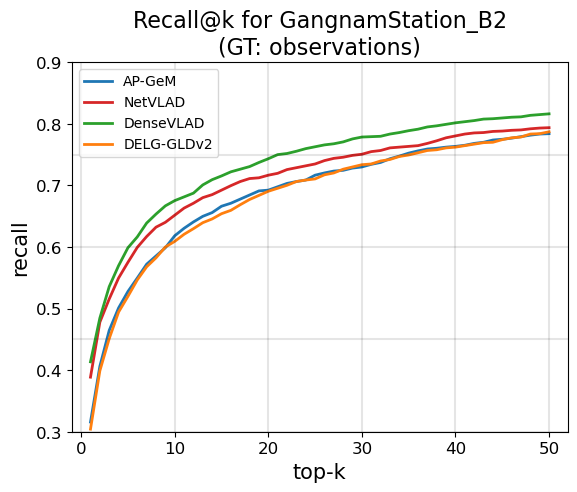} \\
 \includegraphics[width=0.23\textwidth]{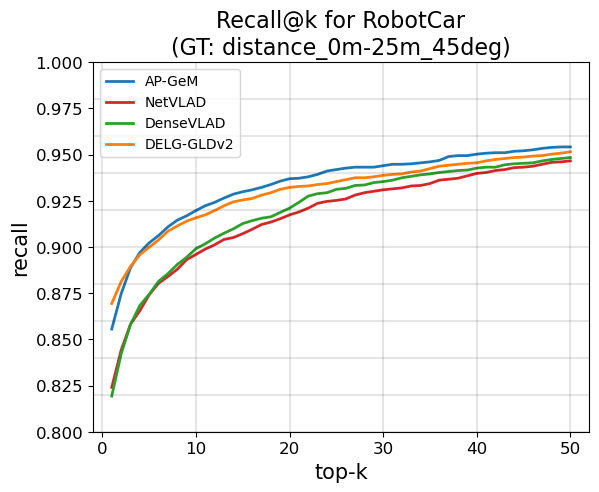}
 \includegraphics[width=0.23\textwidth]{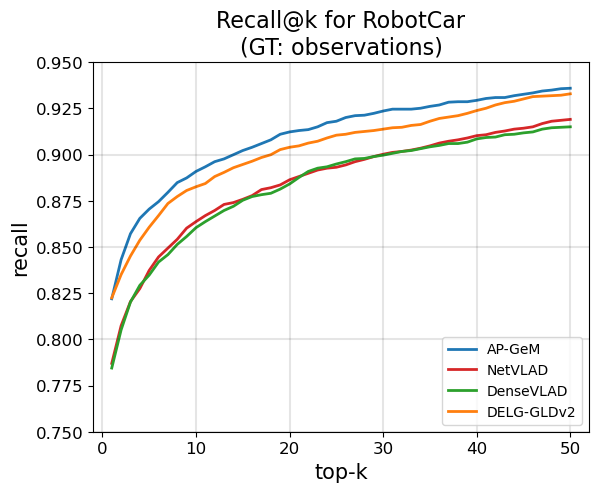}
\end{center}
\vspace{-0.3cm}
   \caption{\textbf{Place recognition (Recall@$k$)}. Place recognition performance measured with Recall@$k$ (R@$k$) using the pose distance (left column) and co-observation-based (right column) GT definitions.
   While there is no feature that performs best on all datasets, AP-GeM and DELG-GLDv2 are often among the top, except on GangnamStation\_B2 where they perform worst.}
\label{fig:placerecognition_metric}
\end{figure}

To evaluate place recognition, the literature often suggest to use \emph{Recall@$k$} (R@$k$)~\cite{ToriiPAMI18247PlaceRecognitionViewSynthesis,ArandjelovicACCV14DislocationDistinctivenessForLocation,ArandjelovicCVPR16NetVLADPlaceRecognition}, which measures the percentage of query images with at least one relevant database image amongst the top $k$ retrieved images.
A database image is typically considered relevant if it was taken within a neighborhood of the query image ~\cite{ArandjelovicACCV14DislocationDistinctivenessForLocation,ToriiPAMI15VisualPlaceRecognRepetitiveStructures,ToriiPAMI18247PlaceRecognitionViewSynthesis}. 
Alternatively, co-observations were also used to determine whether or not a database image represents the same place~\cite{ChenCVPR11CityScaleLandmarkIdentification}.
Therefore, in Fig.~\ref{fig:placerecognition_metric}, we show R@$k$ for all datasets with available retrieval GT based on relative camera pose between query and database images (left) and co-observations-based (right column) GT rankings. 

\PAR{Comparison between global representations}
The figure shows that R@$k$ rises rapidly with increasing $k$.
For almost all feature types and datasets considered, at least one image is relevant among top 50 retrieved images.
While there is no feature that performs best on all datasets, AP-GeM and DELG-GLDv2 are often among the top, except on GangnamStation\_B2 where they perform worst. 
The main reason for this behavior is that the images from GangnamStation\_B2 contain many dynamic objects, blur, and low frequencies in general.
As will be discussed in Sec.~\ref{sec:BlurrExp}, both AP-GeM and DELG-GLDv2 show a more significant drop caused by these challenges (especially blur), than the other features types.

\begin{figure*}[t]
\begin{center}
 \includegraphics[width=\widthfivefigs\textwidth]{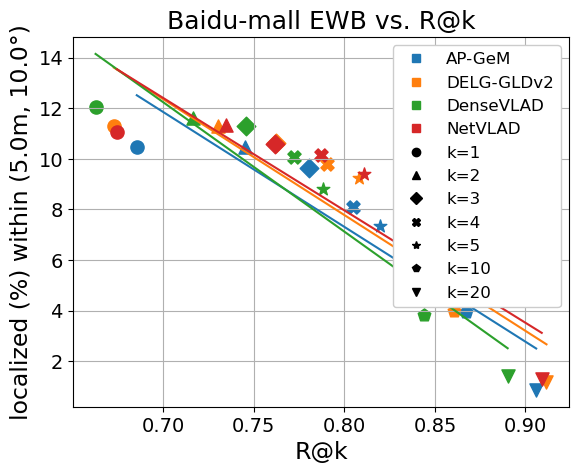}
 \includegraphics[width=\widthfivefigs\textwidth]{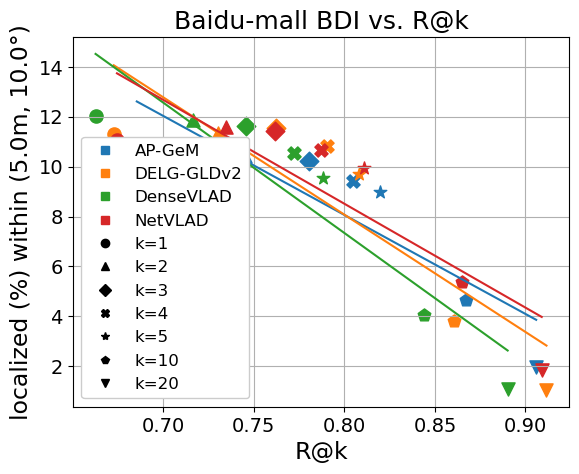}
 \includegraphics[width=\widthfivefigs\textwidth]{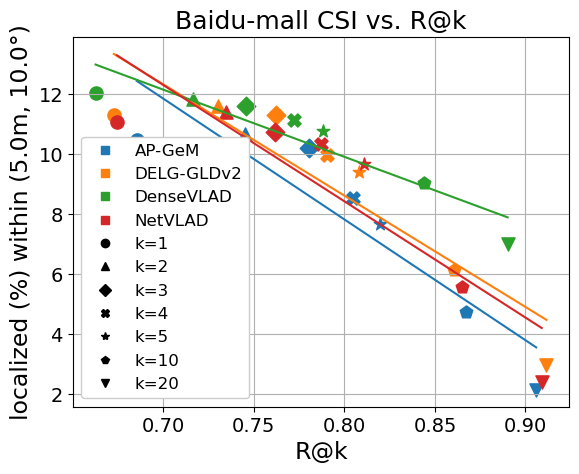}
 \includegraphics[width=\widthfivefigs\textwidth]{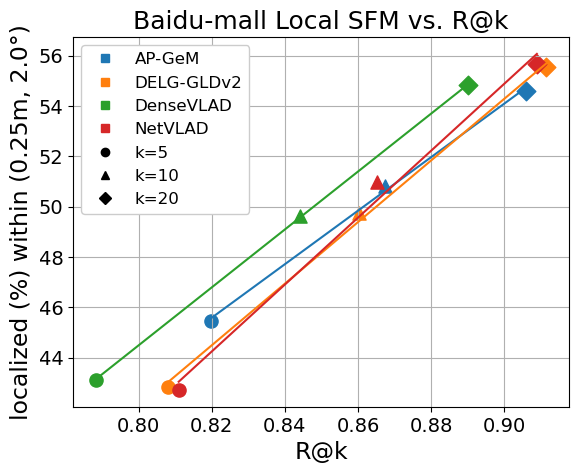}
 \includegraphics[width=\widthfivefigs\textwidth]{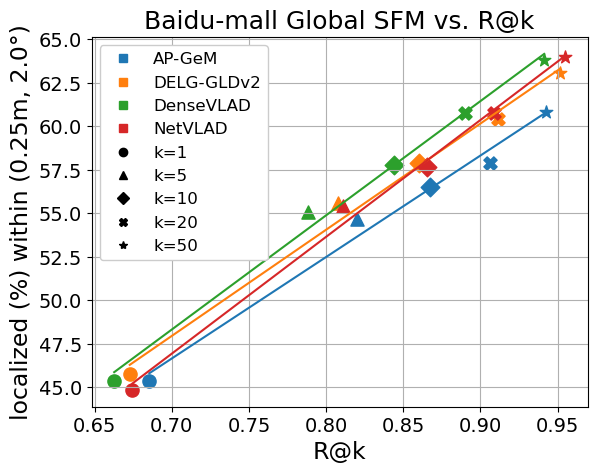}
 \\
 \includegraphics[width=\widthfivefigs\textwidth]{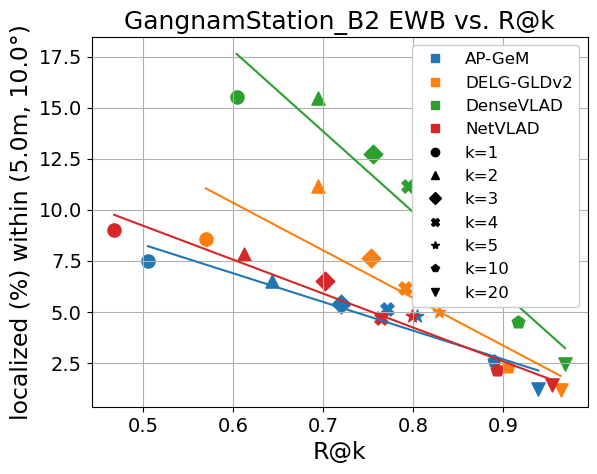}
 \includegraphics[width=\widthfivefigs\textwidth]{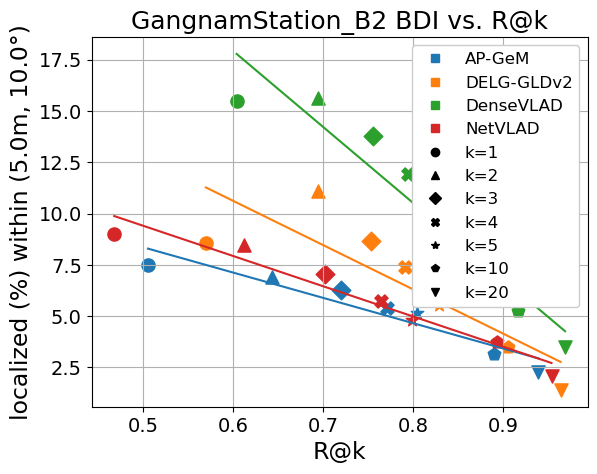}
 \includegraphics[width=\widthfivefigs\textwidth]{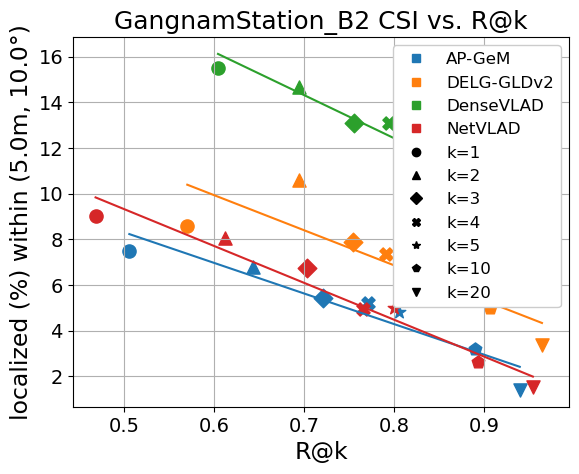}
 \includegraphics[width=\widthfivefigs\textwidth]{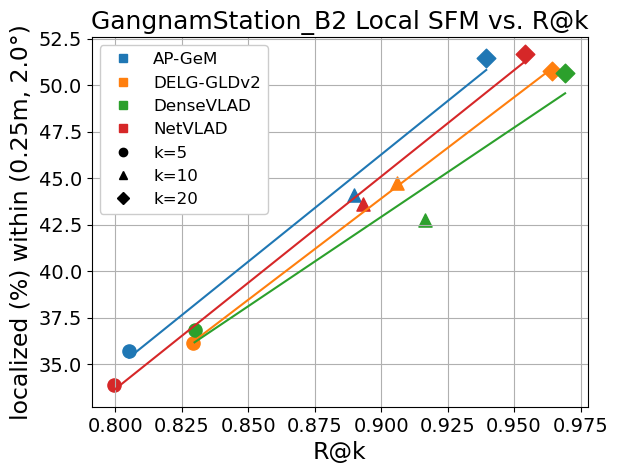}
 \includegraphics[width=\widthfivefigs\textwidth]{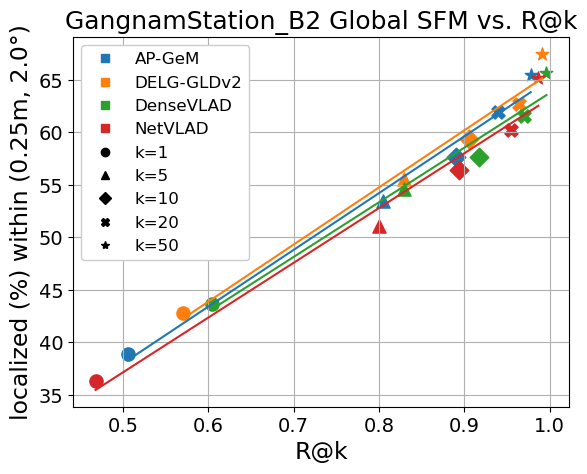}
 \\
 \includegraphics[width=\widthfivefigs\textwidth]{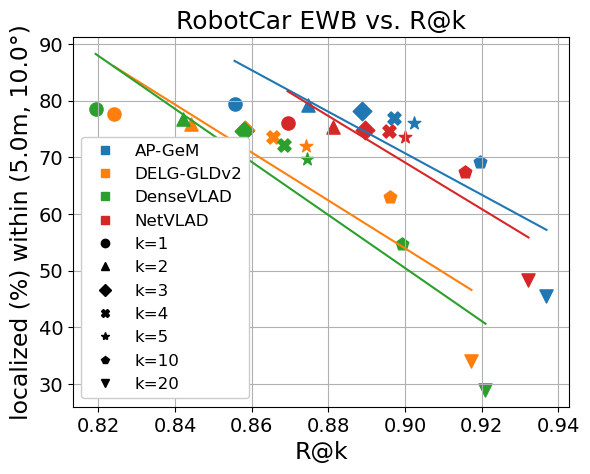}
 \includegraphics[width=\widthfivefigs\textwidth]{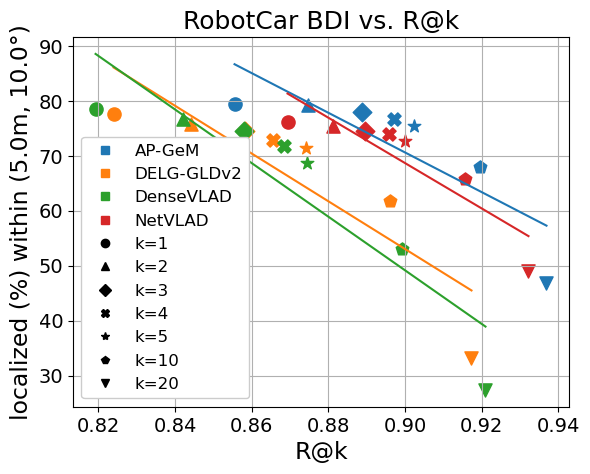}
 \includegraphics[width=\widthfivefigs\textwidth]{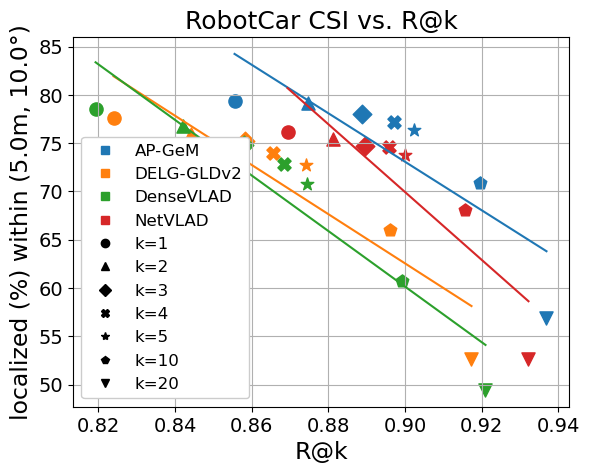}
 \includegraphics[width=\widthfivefigs\textwidth]{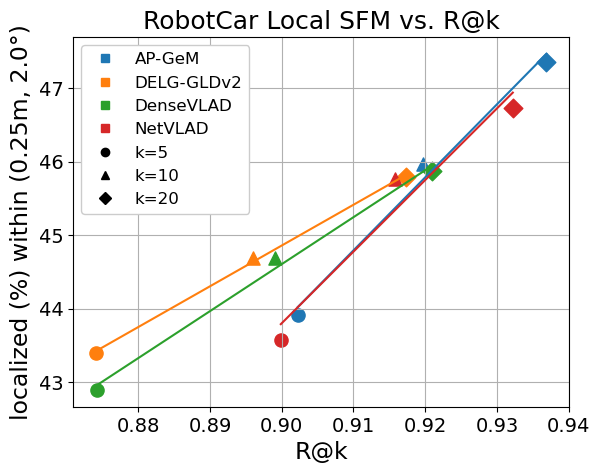}
 \includegraphics[width=\widthfivefigs\textwidth]{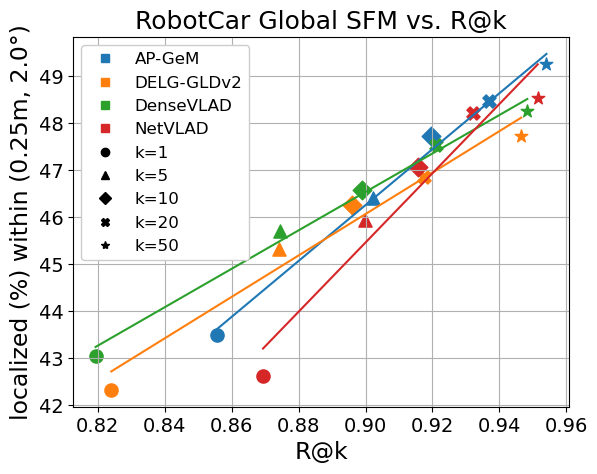}
\end{center}
\vspace{-0.3cm}
   \caption{\textbf{Place recognition correlation}. Correlation between place recognition (R@$k$ using distance-based GT) and the three visual localization \tasks \, (successfully localized images in \%) shown as scatter plot, one per \task \, (column), and dataset (row), where each point represents a global feature and a top $k$ value. To show whether or not the two metrics are correlated, we fit a line on all top $k$ experiments for each feature type. As can be seen, there is a linear correlation for accurate pose estimation and an inverse correlation for pose approximation. 
   }
\label{fig:placerecognition_scatter}
\end{figure*}

\PAR{Evaluating linear correlation between place recognition (R@$k$) and localization accuracy}
As for landmark retrieval, we evaluate linear correlation between place recognition (R@$k$ using distance-based GT) and visual localization (considering the successfully localized images in \%) visually using a scatter plot and numerically using the Pearson~(\ref{eq:pearson}) correlation coefficient.
Figure~\ref{fig:placerecognition_scatter} shows one plot per method (column) and dataset (row), where each point represents the results obtained with a global feature and a top $k$ value. 
To show whether or not the two metrics are correlated, we fit a line on all top $k$ experiments for each feature type.
As can be seen in the figure, and contrary to landmark retrieval, there is a linear correlation for accurate pose estimation (\Task~2), but only an inverse correlation for pose approximation. 
The linear correlation with \Task~2 can be explained by the observation that the larger $k$ the higher both R@$k$ (Fig.~\ref{fig:placerecognition_metric}) and localization accuracy (Fig.~\ref{fig:exp:irbench:task2a}~and~\ref{fig:exp:irbench:task2b}).
Since we need at least one relevant image for global SFM, a high recall increases the chance for successful pose estimation.
Similar for local SFM, where larger $k$s (\ie higher recall) lead to better performance because multiple relevant images are needed to construct the local SFM map.
Pose approximation benefits from high precision and smaller $k$s (where R@$k$ is low), which leads to the inverse correlation. 

\begin{figure*}[t]
\begin{center}
 \includegraphics[width=\widthfivefigs\textwidth]{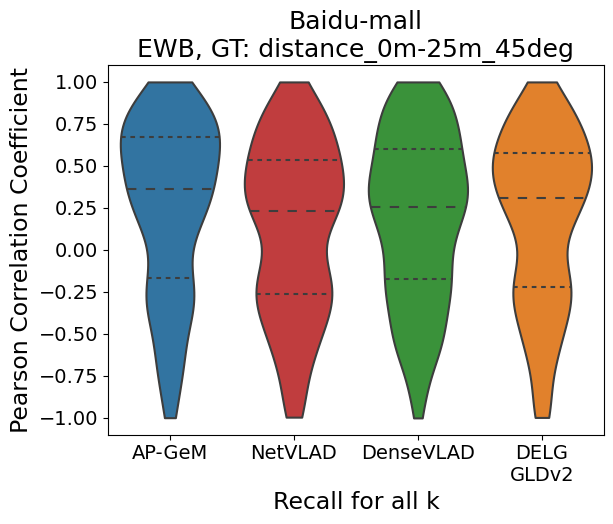}
 \includegraphics[width=\widthfivefigs\textwidth]{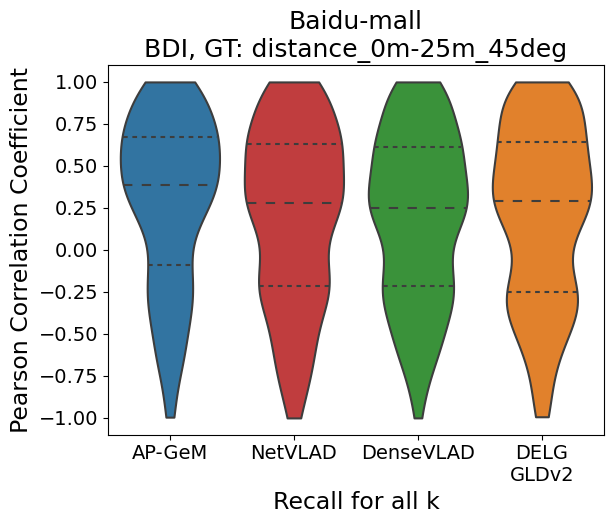}
 \includegraphics[width=\widthfivefigs\textwidth]{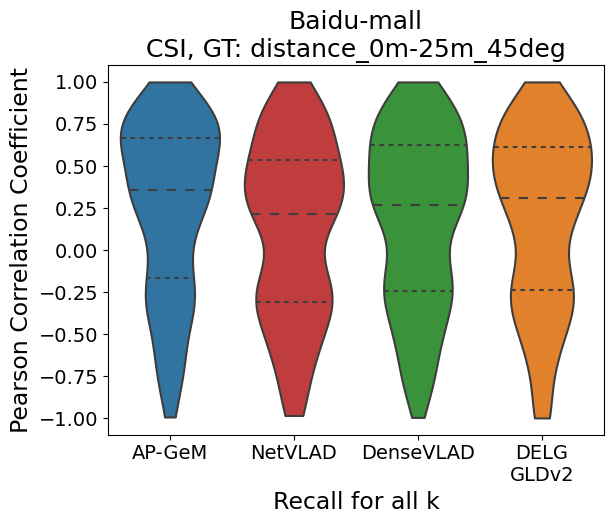}
 \includegraphics[width=\widthfivefigs\textwidth]{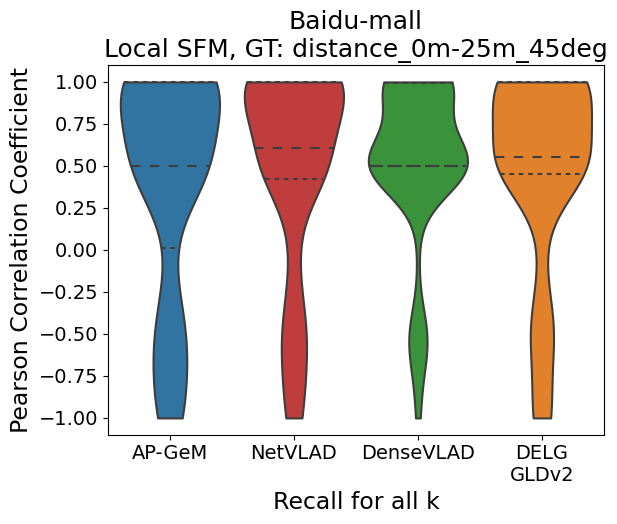}
 \includegraphics[width=\widthfivefigs\textwidth]{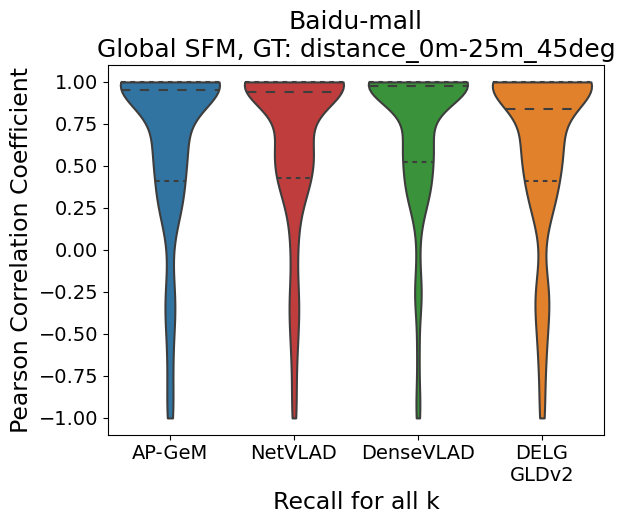} \\
 \includegraphics[width=\widthfivefigs\textwidth]{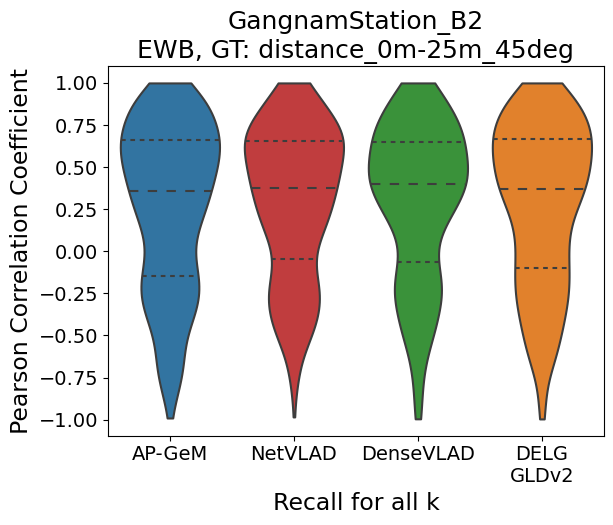}
 \includegraphics[width=\widthfivefigs\textwidth]{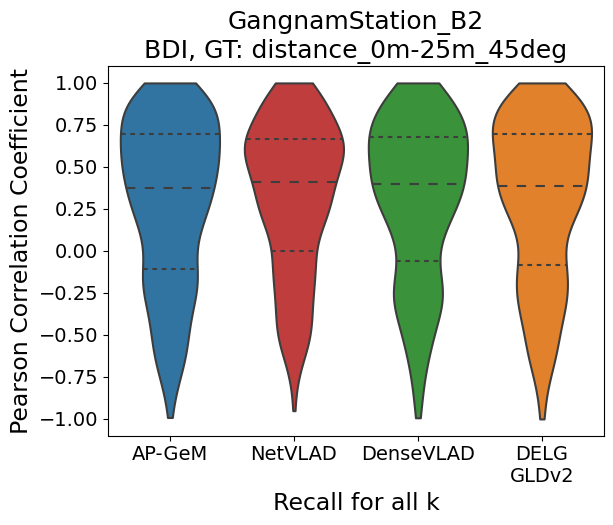}
 \includegraphics[width=\widthfivefigs\textwidth]{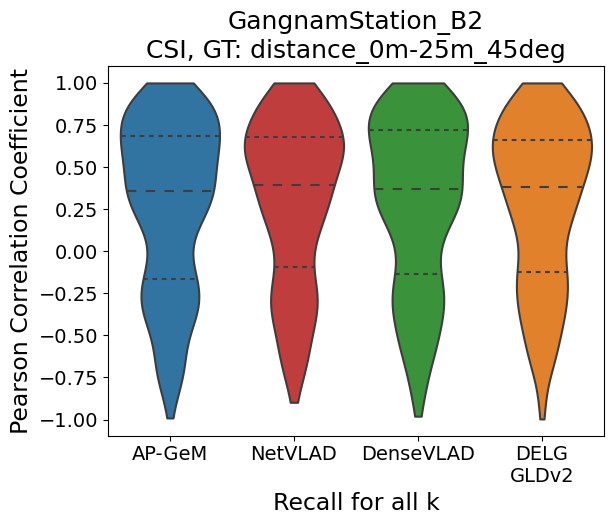}
 \includegraphics[width=\widthfivefigs\textwidth]{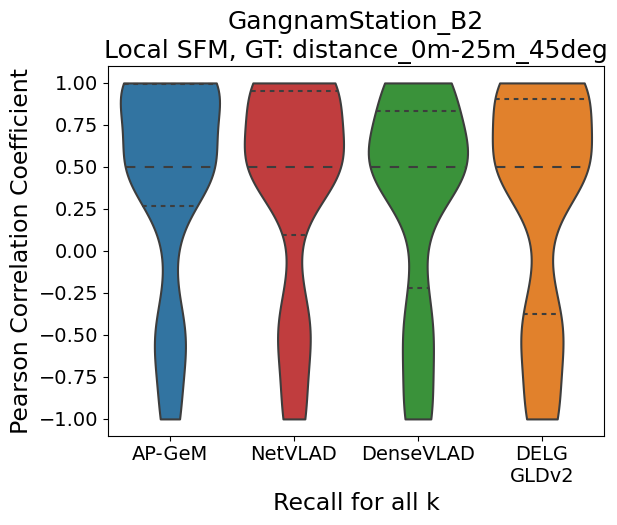}
 \includegraphics[width=\widthfivefigs\textwidth]{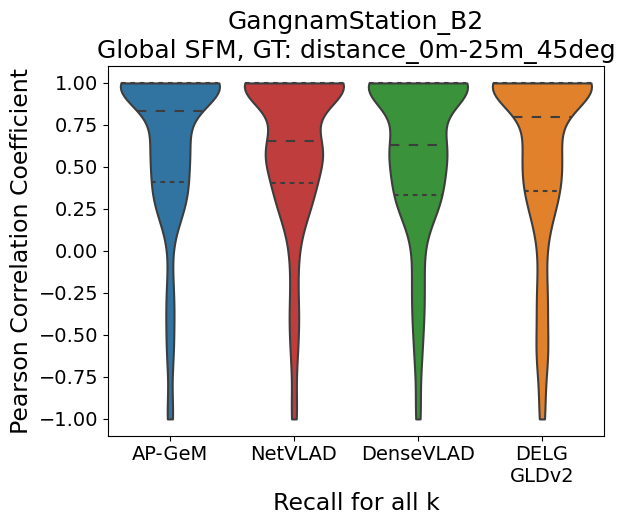} \\
 \includegraphics[width=\widthfivefigs\textwidth]{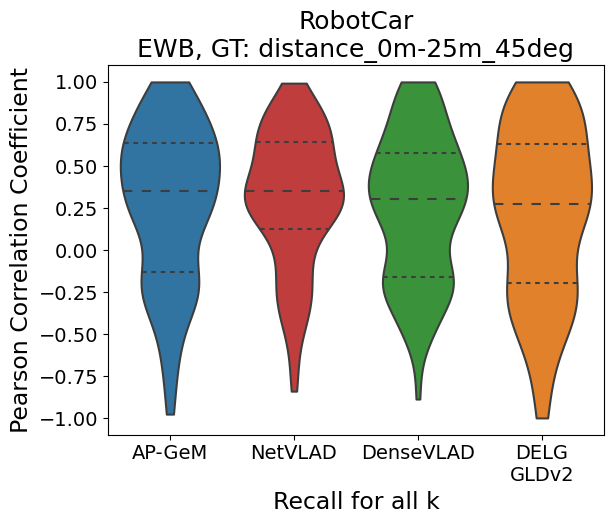}
 \includegraphics[width=\widthfivefigs\textwidth]{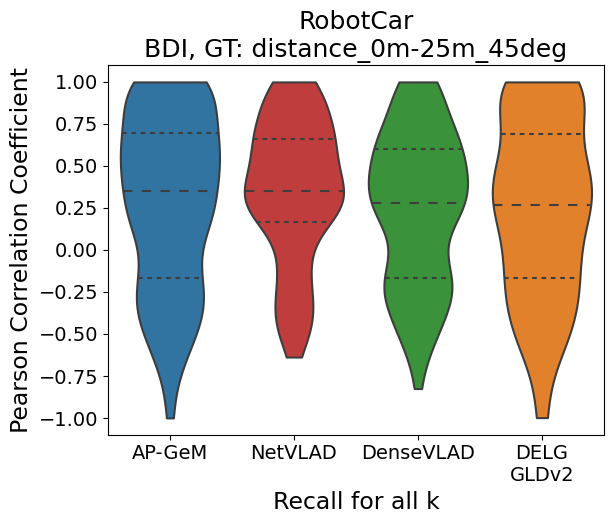}
 \includegraphics[width=\widthfivefigs\textwidth]{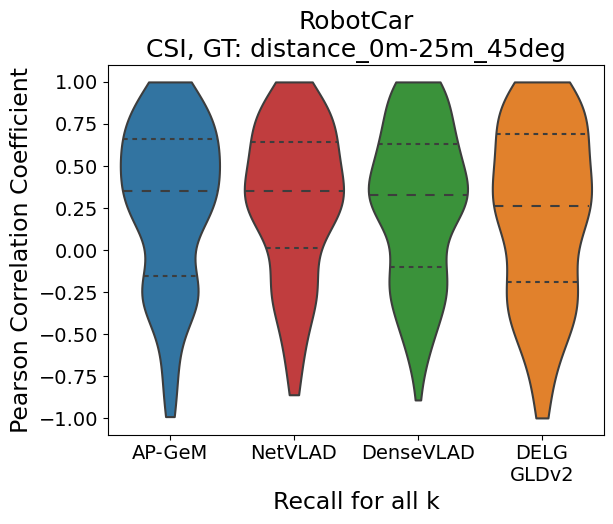}
 \includegraphics[width=\widthfivefigs\textwidth]{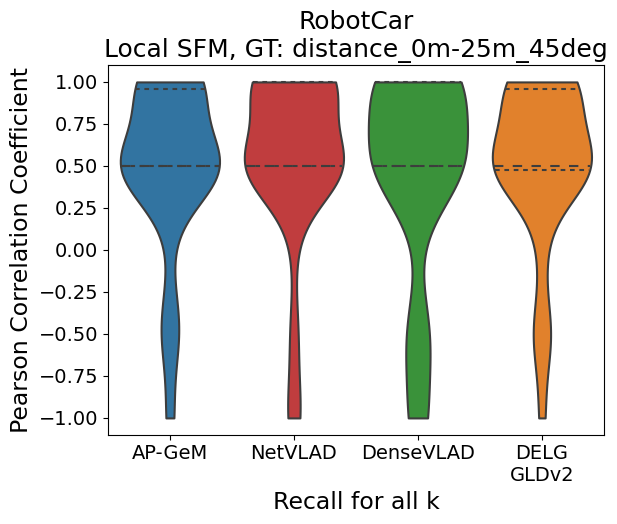}
 \includegraphics[width=\widthfivefigs\textwidth]{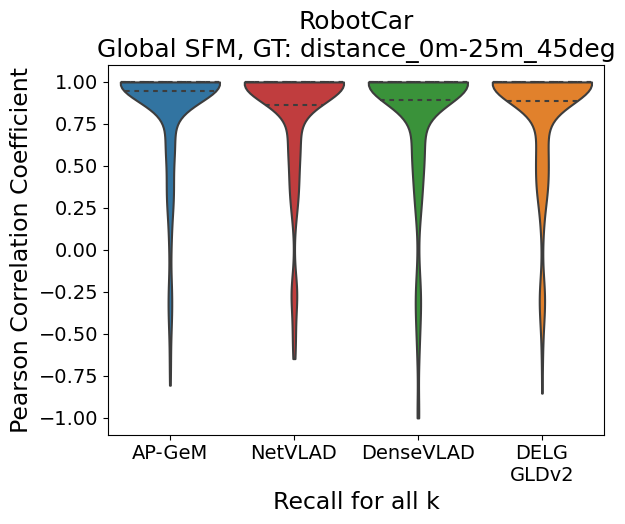}
\end{center}
\vspace{-0.3cm}
   \caption{\textbf{Place recognition per image linear correlation}. Pearson coefficients distribution over each query image individually (directly using the pose error as localization metric) and visualized as violin plots. The columns show the localization \tasks \, and the rows show different datasets. For pose estimation, the Pearson coefficients are densely sampled in the upper part of the violins, meaning good linear correlation, whereas for pose approximation, they are sampled more evenly which weakens the inverse linear correlation observed in the scatter plot.}
\label{fig:placerecognition_violin}
\end{figure*}

Again, the correlation is also shown numerically using the Pearson correlation coefficient, computed for the entire dataset (using successfully localized images as the localization metric) and for each global feature type, which is close to 1 for accurate pose estimation methods (\Task~2) and close to -1 for pose approximation (\Task~1), see Tab.~\ref{tab:placerecognition_pearson}. 
Figure~\ref{fig:placerecognition_violin} shows the distribution of the Pearson coefficients computed for each image individually.

Confirming the findings from above, for accurate pose estimation (\Task~2) the Pearson coefficients are densely sampled in the upper part of the violins (linear correlation), whereas for pose approximation, they are sampled more evenly between -1 and 1.
This weakens the inverse linear correlation observed in the scatter plot and the per dataset PCC.

\begin{table}[t!]
\center
\caption{\textbf{Place recognition Spearman correlation} coefficient between R@$k$ (using distance-based GT) and localization accuracy (successfully localized images in \%) shown per datasets. Interestingly, contrary to rank correlation of landmark retrieval, here the rank coefficients show similar rank correlation for all localization \tasks.}
\label{tab:placerecognition_spearman}
\resizebox{\linewidth}{!}{
\input{IJCV_plots/table_recall_spearman}
}
\end{table}

\begin{table}[t!]
\center
\caption{\textbf{Place recognition Pearson correlation} coefficients between R@$k$ (using distance-based GT) and localization accuracy (successfully localized images in \%) shown per datasets. Overall, we can see a good correlation (Pearson coefficient close to 1) between R@$k$ and the accurate pose estimation with local or global SFM map and an inverse correlation (Pearson coefficient close to -1)
with interpolation-based localization methods.}
\label{tab:placerecognition_pearson}
\resizebox{\linewidth}{!}{
\input{IJCV_plots/table_recall_pearson}
}
\end{table}

\PAR{Evaluating rank correlation between place recognition (R@$k$) and localization}
We report rank correlation as Spearman coefficient computed per dataset in Tab.~\ref{tab:placerecognition_spearman}. 
As for landmark retrieval, we do not observe a clear (for all datasets) rank correlation between R@k and localization accuracy (all \tasks), \ie, the performance of the features between both settings is not consistent.

\subsection{Analysis of impact of blur and dynamic scenes}
\label{sec:BlurrExp}

\begin{figure*}[t]
\begin{center}
 \includegraphics[width=0.23\textwidth]{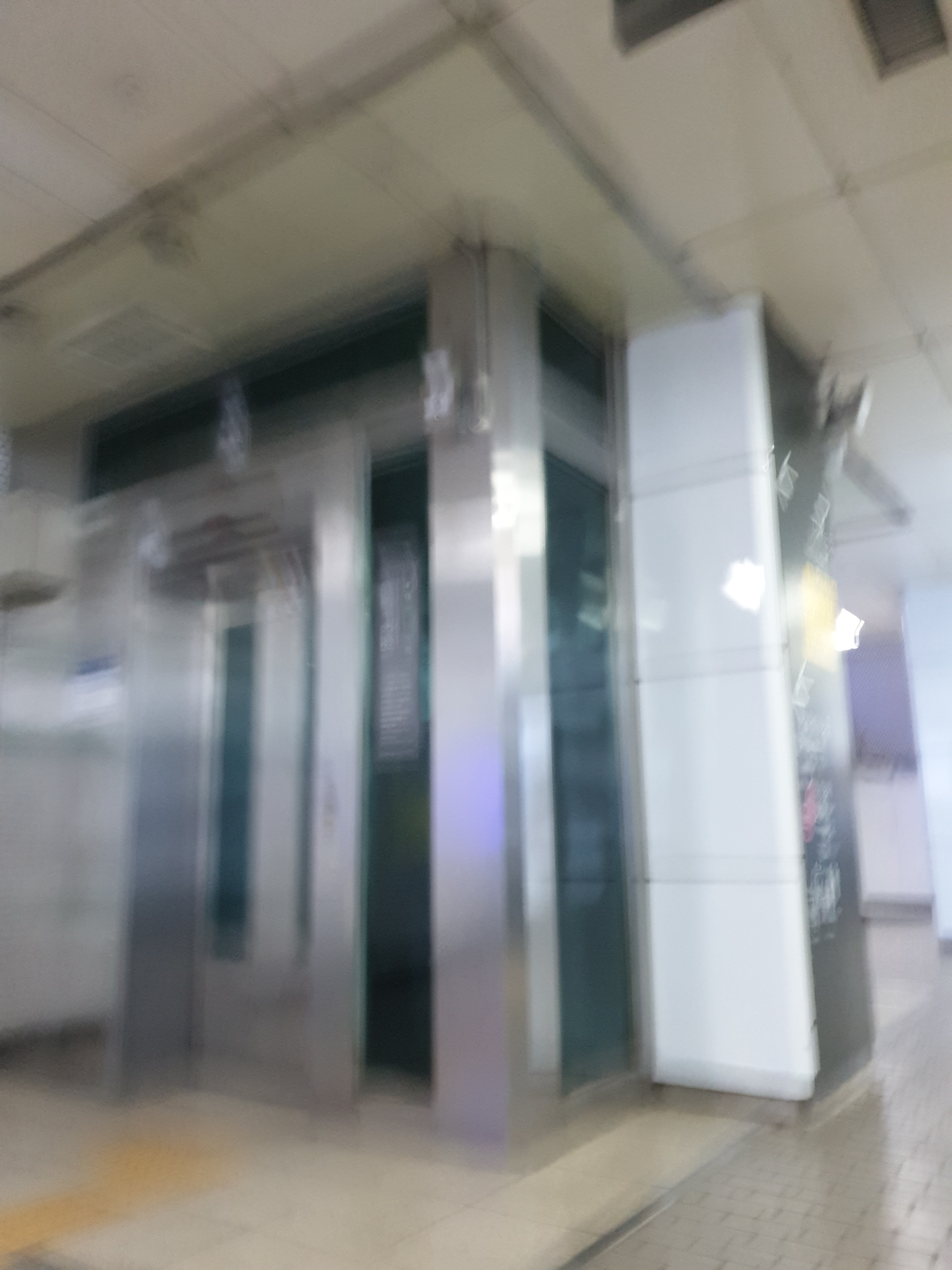}
 \includegraphics[width=0.23\textwidth]{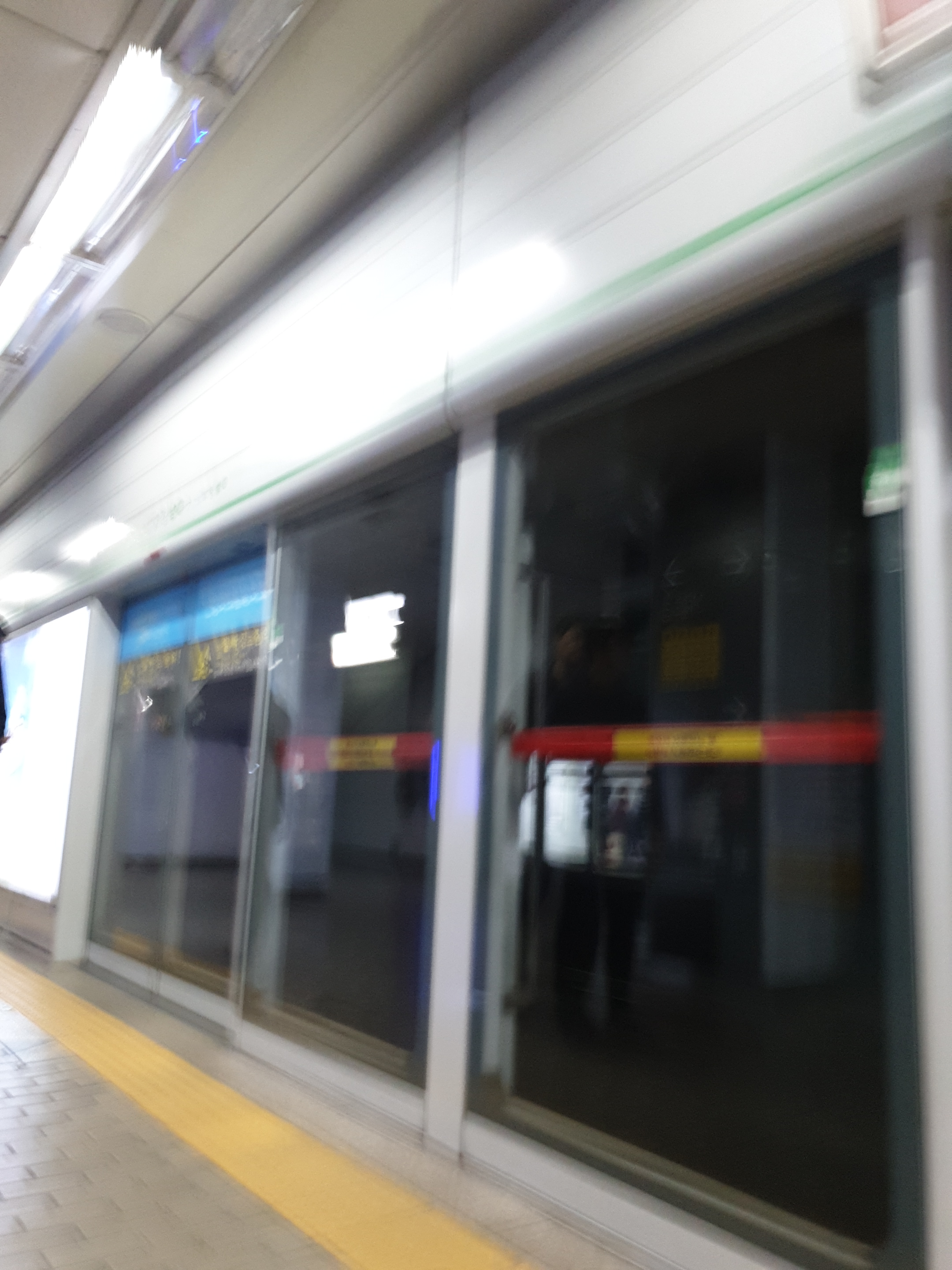}
 \includegraphics[width=0.23\textwidth]{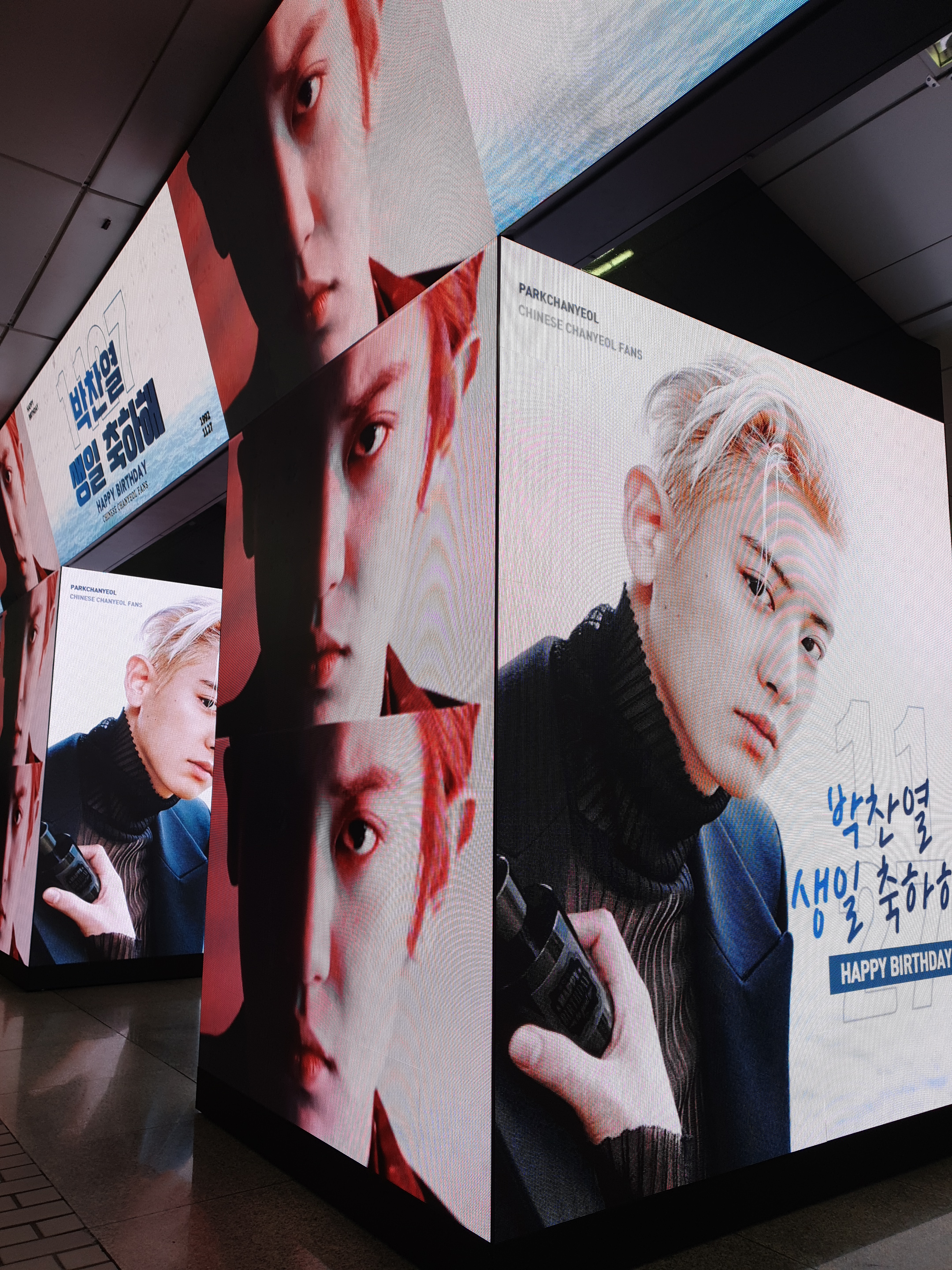}
 \includegraphics[width=0.23\textwidth]{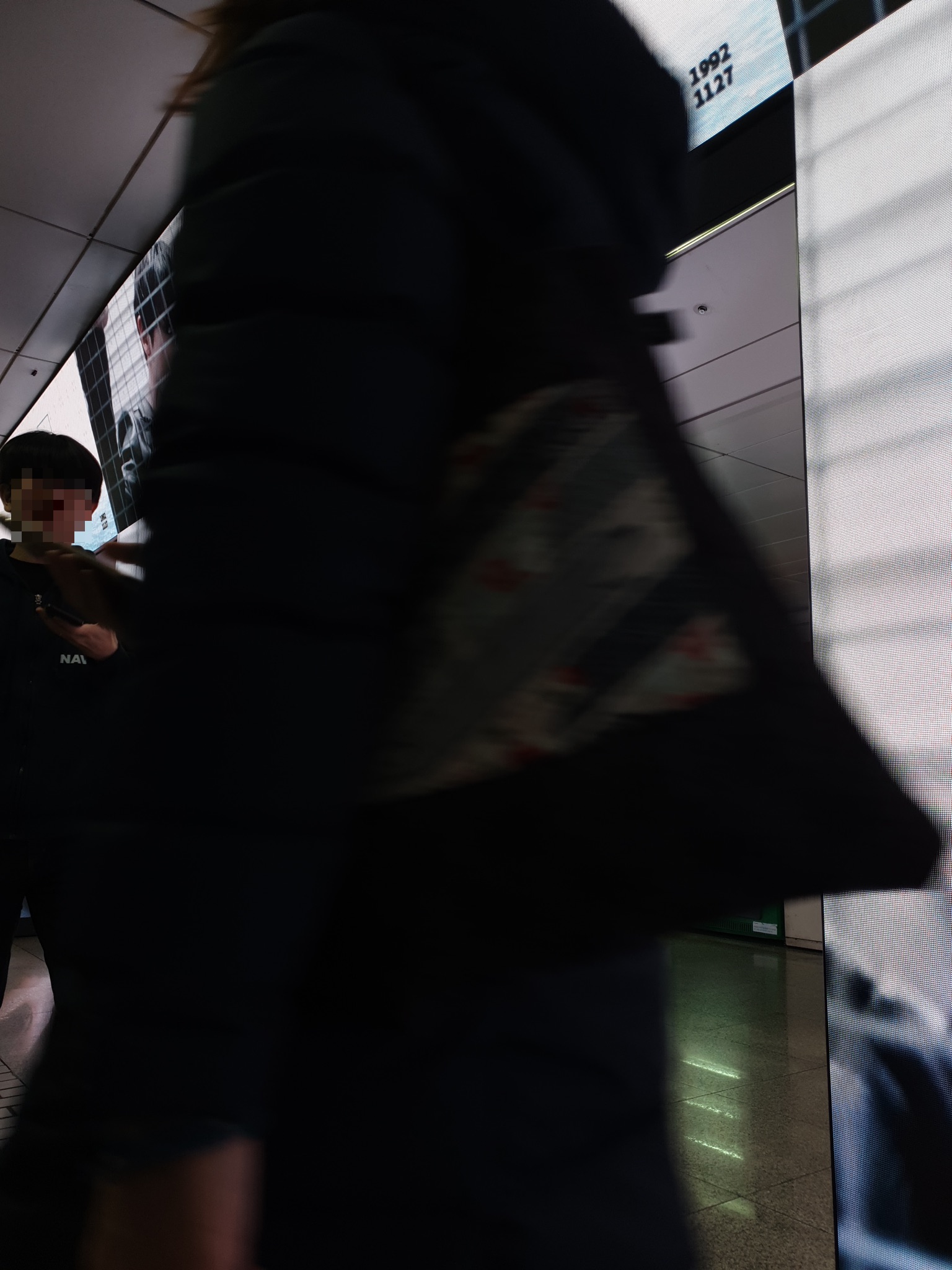} \\
 \includegraphics[width=0.23\textwidth]{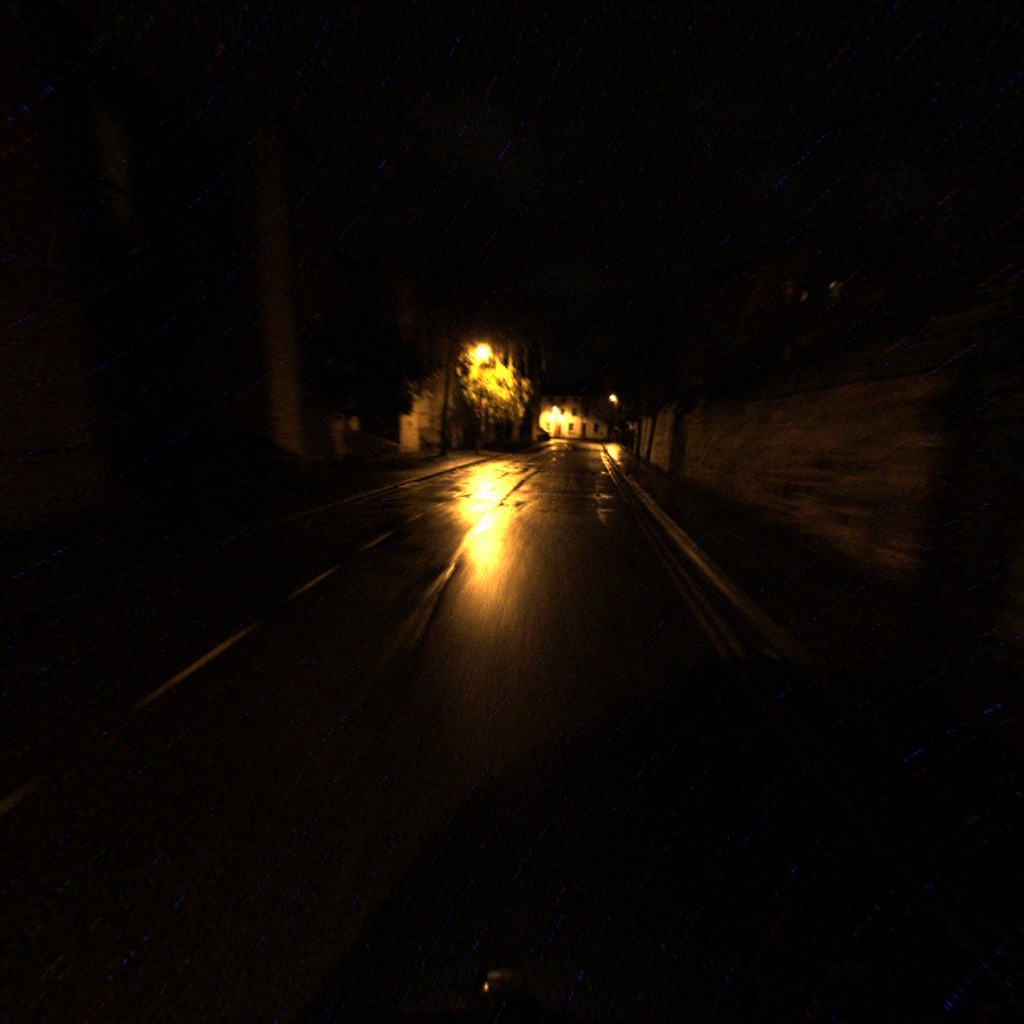}
 \includegraphics[width=0.23\textwidth]{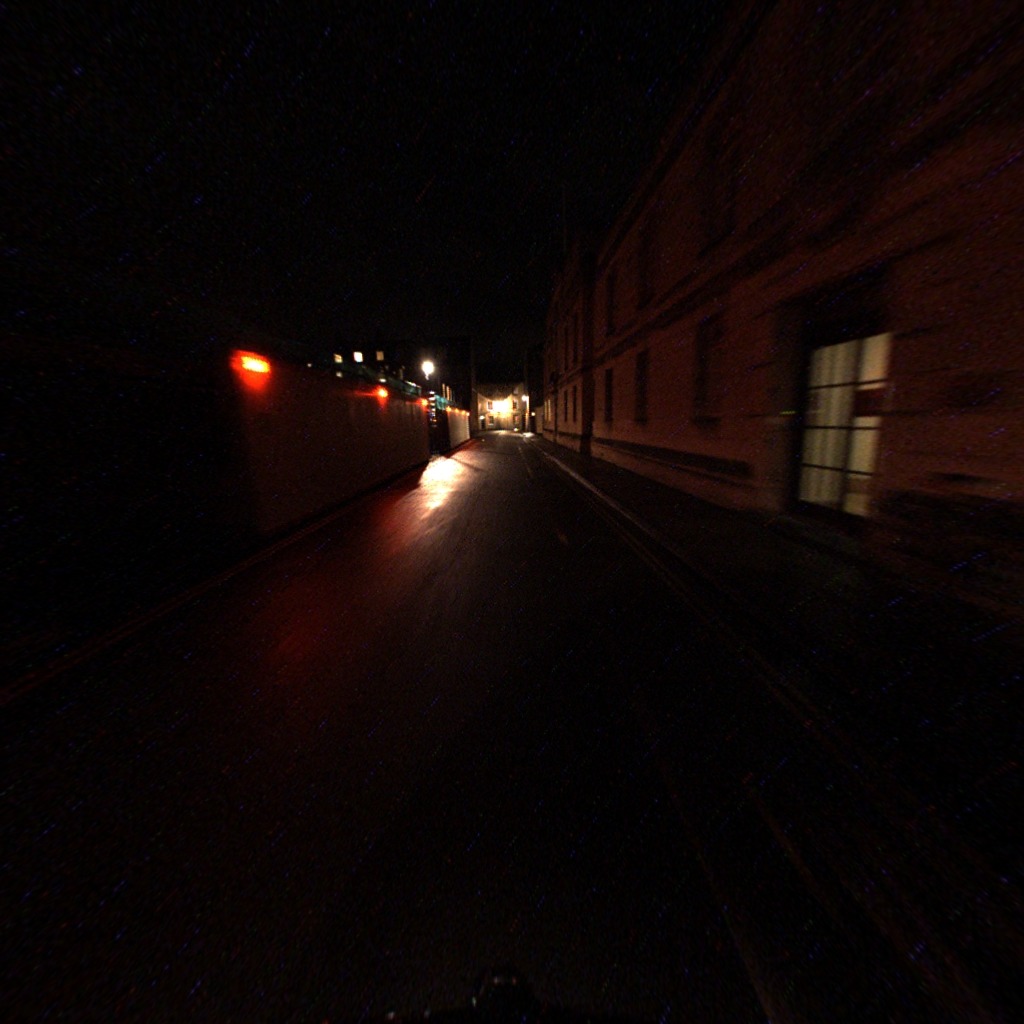}
 \includegraphics[width=0.23\textwidth]{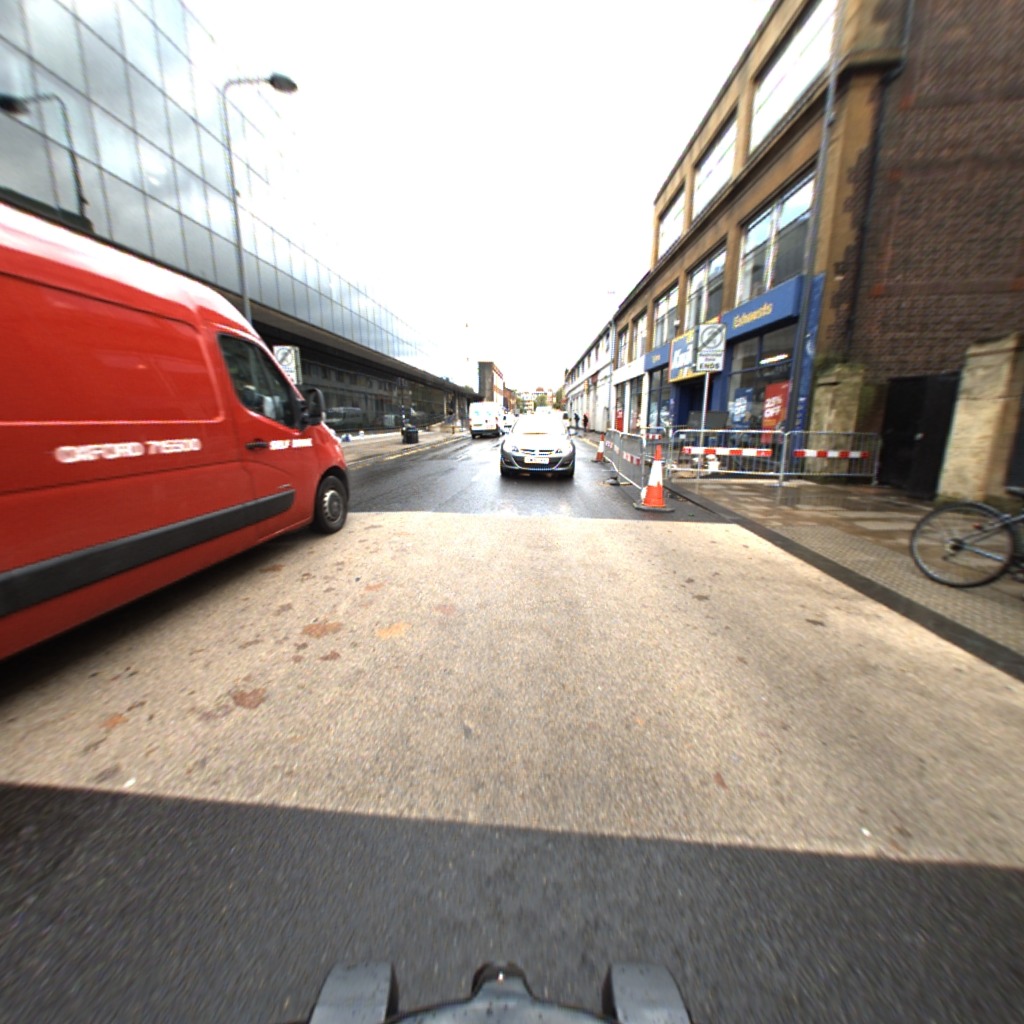}
 \includegraphics[width=0.23\textwidth]{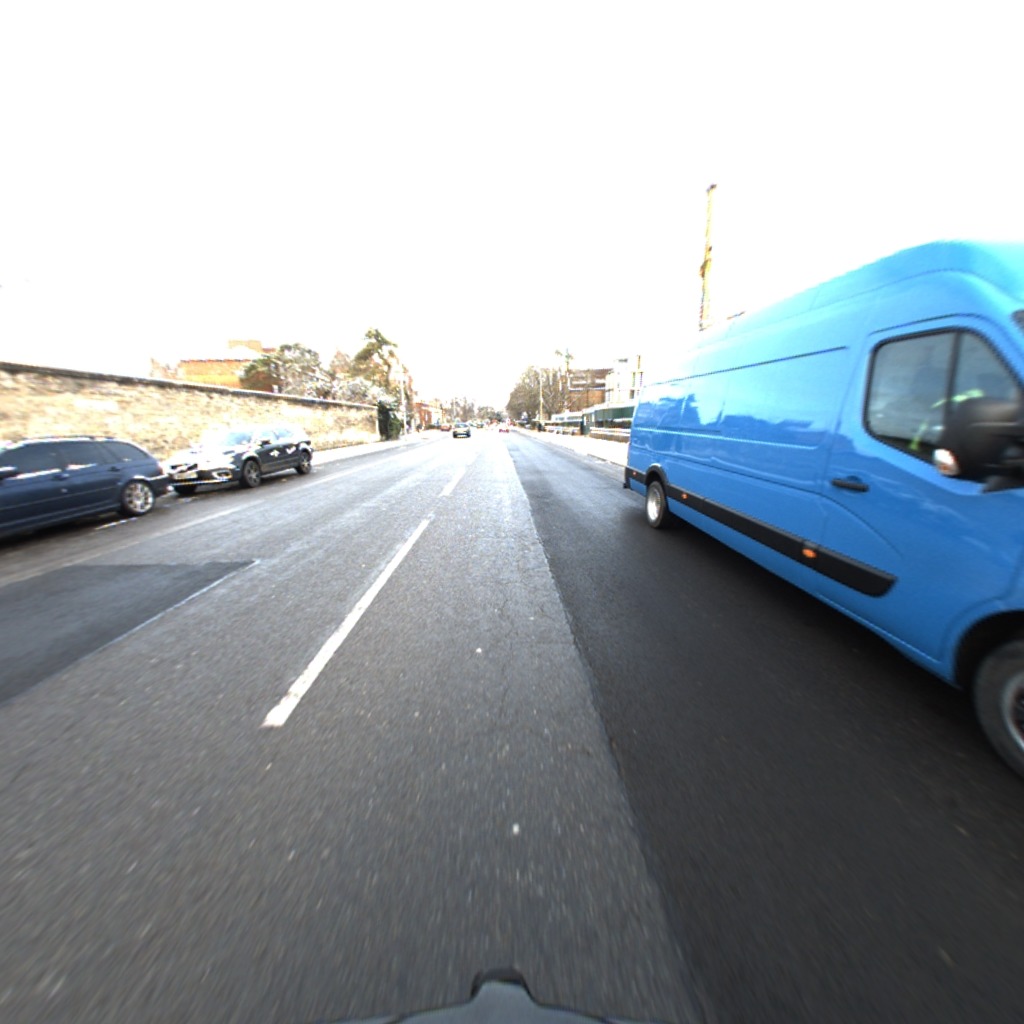}
\end{center}
\vspace{-0.3cm}
   \caption{\textbf{Blur and dynamic scenes}. Example images of blur (left) and dynamic scenes (right) in query sets for GangnamStation\_B2 (top) and RobotCar (bottom).}
\label{fig:blur_examples}
\end{figure*}

In this section, we study the impact of particular challenges on localization and retrieval performance. Namely, we consider the following two challenges: (a) blur, caused by, \eg, rapid motion of the capturing device, and (b) the fact that scenes contain dynamic or moving objects such as people or cars.
More precisely, we compare the performance when considering all test images and when considering only a subset of difficult images for a given challenge.
Figure~\ref{fig:blur_examples} shows examples for both challenges of GangnamStation\_B2 and RobotCar.

To characterize image blur, we compute the mean absolute difference between the original image and its reconstructed version when removing the high frequencies in the Fourier domain, following~\cite{LiuCVPR2008ImagePartialBlur}\footnote{{{https://www.pyimagesearch.com/2020/06/15/opencv-fast-fourier-transform-fft-for-blur-detection-in-images-and-video-streams/}}}.
This in fact measures the absence of high frequency signals in the images.
We use a mean absolute difference threshold of 20 to select the subset of blurry images.

For defining the dynamic scenes, we run a Mask R-CNN instance segmentation model~\cite{HeICCV2017MaskRCNN} trained on the MS COCO dataset~\cite{LinECCV2014microsoftcoco} and consider the subset of images where at least 20\% of the image pixels belong to a category that is dynamic such as person or car but also objects and animals.

\begin{figure*}[t]
\begin{center}
 \includegraphics[width=0.23\textwidth]{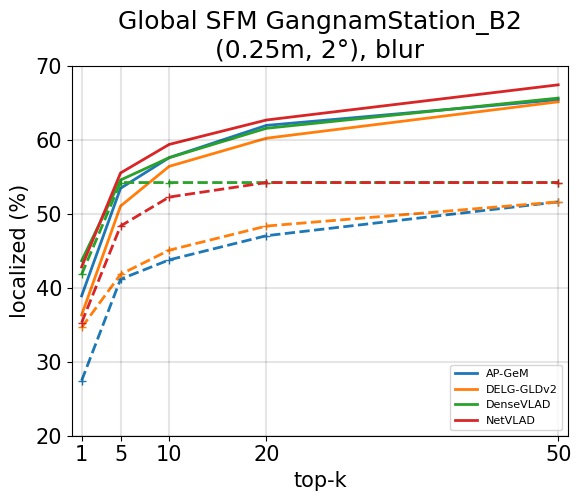}
 \includegraphics[width=0.23\textwidth]{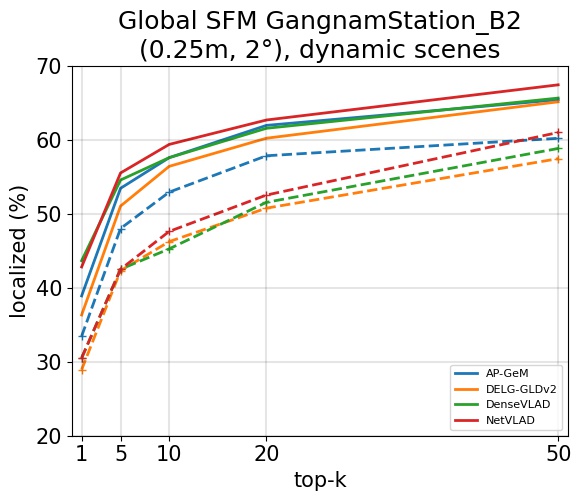}
 \includegraphics[width=0.23\textwidth]{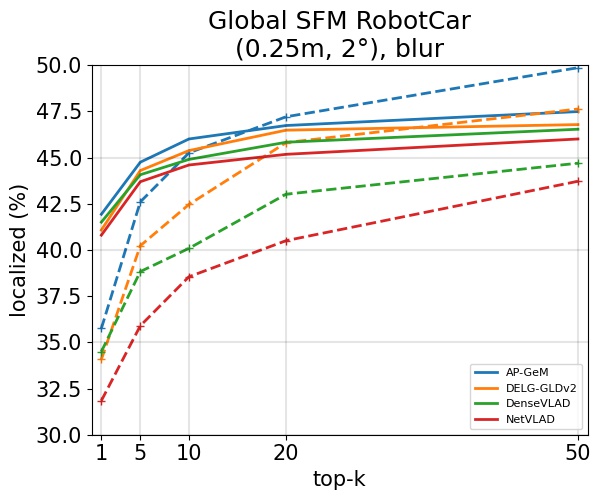}
 \includegraphics[width=0.23\textwidth]{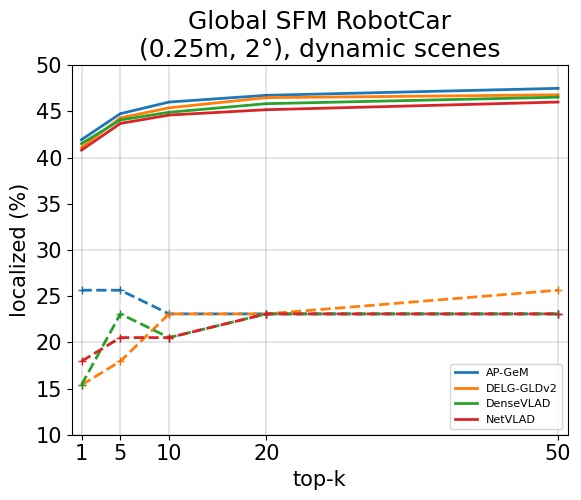} \\
 \includegraphics[width=0.23\textwidth]{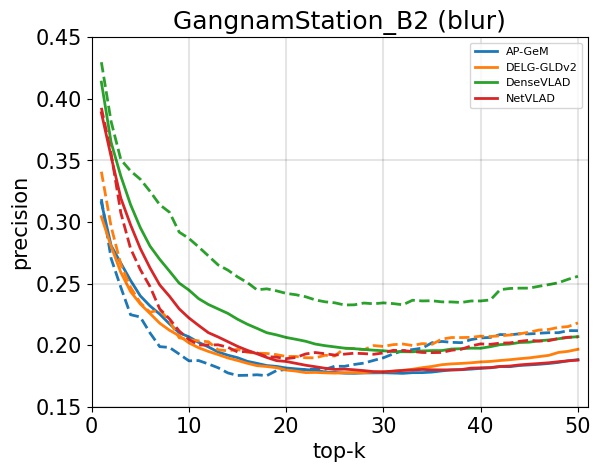}
 \includegraphics[width=0.23\textwidth]{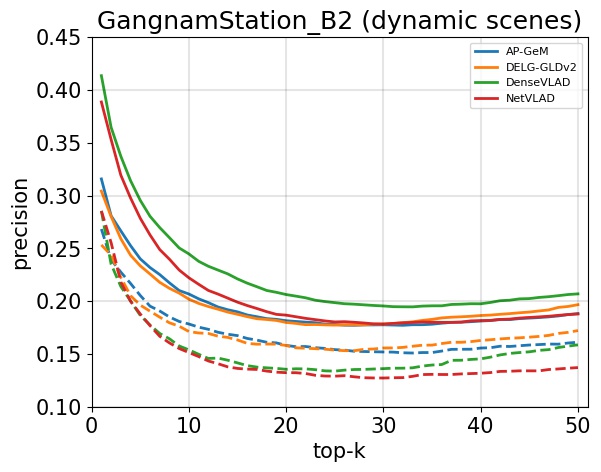}
 \includegraphics[width=0.23\textwidth]{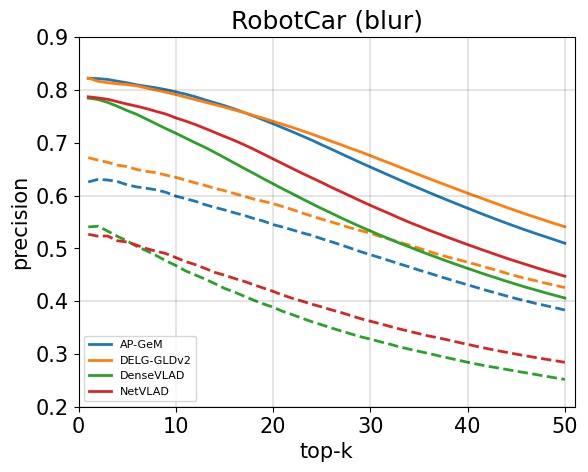}
 \includegraphics[width=0.23\textwidth]{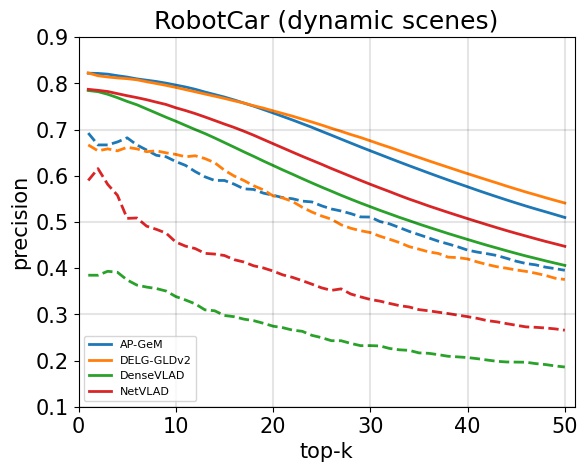} \\
 \includegraphics[width=0.23\textwidth]{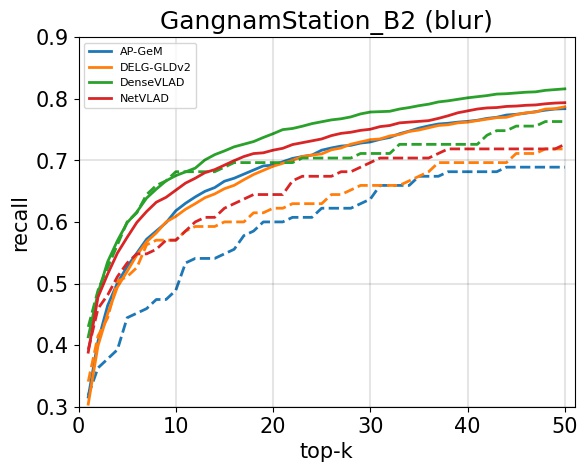}
 \includegraphics[width=0.23\textwidth]{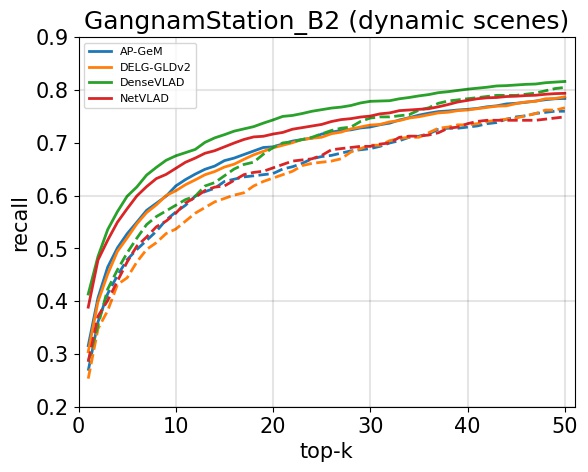}
 \includegraphics[width=0.23\textwidth]{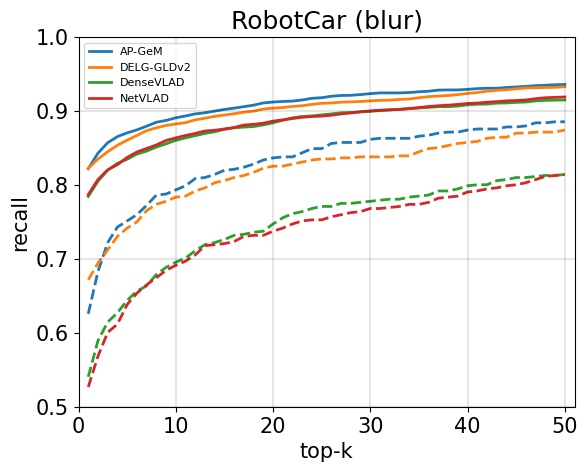}
 \includegraphics[width=0.23\textwidth]{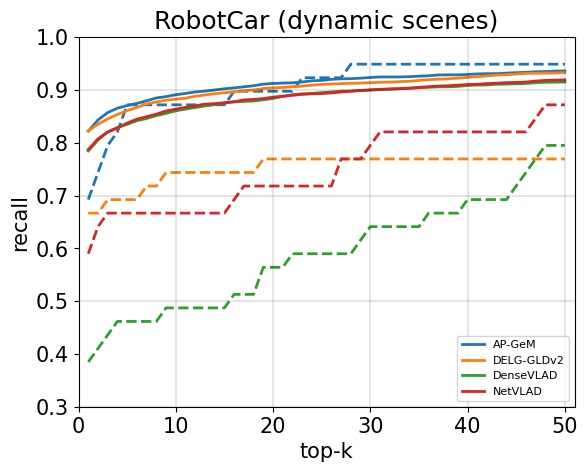}
\end{center}
\vspace{-0.3cm}
   \caption{\textbf{Blur and dynamic scenes}. Impact of blur and dynamic scenes on localization performance (with global SFM) in top row, precision (second row), and recall (third row) for datasets GangnamStation\_B2 (first two columns) and RobotCar (last two columns). The solid lines show localization performance for all test images, while the dashed lines correspond to the subset of difficult images (considering blur and dynamic scenes).}
\label{fig:blur}
\end{figure*}

In the first row of Fig.~\ref{fig:blur}, we show the localization performance with global SFM (\Task~2b) on the datasets GangnamStation\_B2 and RobotCar for all test images (solid lines) and for the subset of challenging images (dashed lines) defined by blur and dynamic scenes.
These two challenges lead to a clear drop in localization performance. 

As can be seen, blur causes the successfully localized images to drop by 15\% on GangnamStation\_B2 (first column) and roughly 20\% for dynamic scenes in RobotCar (last column). 
This drop occurs for all global descriptors, except AP-GeM and DELG-GLDv2 for the blurry images of RobotCar \ccc{when retrieving $k > 10$. Note that in this case the performance even increases. 
It seems like that, because the subset of blurry images is significantly smaller than the full set of images, using these descriptors for this dataset, more relevant images (and most likely better ones among them) can be found in comparison to using the full set of images}.

We then study if this performance drop also appears in the retrieval component of the localization method.
The second (resp.~third) row of Fig.~\ref{fig:blur} show the precision (resp.~recall) for various top $k$ and global descriptors for these two challenges on GangnamStation\_B2 and RobotCar.
We observe that the blur has an impact in particular on the recall (third row, first and third column) on both datasets, and on the precision mainly on RobotCar.

\ccc{Interestingly, as can be seen in Fig.~\ref{fig:blur} (second row, first column), precision increases for DenseVLAD when only considering blurry images.
First, as mentioned above, the set of blurry images is significantly smaller than the complete set of images, which increases the chance that relevant images are among the retrieved ones.
Second, and more importantly, the way of computing image blur (absence of high frequency signals) is not perfect. 
This means an image that consist of a large textureless area but also contains a significant amount of details (such as a well-textured but small poster on a white wall), would be wrongly considered as blurry.
Since DenseVLAD is able to focus on smaller details rather than larger image context (see also the InLoc results discussion in Sec.~\ref{sec:expTask2b}), it is able to retrieve such images better than the other descriptors (actually even if they are really blurry but still contain some distinctive details).
This, and the fact that the blurry image set is smaller, leads to higher precision.}

Concerning the dynamic scenes (second and fourth column), it also impacts both precision and recall.
Precision at top 50 for DELG-GLDv2, for instance, is reduced from about 53\% to 39\% on RobotCar and recall from 92\% to 75\%.
We thus conclude that the drop of performance of localization is also linked to the drop of performance of the retrieval component, highlighting potential challenges to tackle in future work.

\section{Conclusion}
\label{sec:conclusion}

Image retrieval plays an important role in modern visual localization systems.
Retrieval techniques are often used to efficiently approximate the pose of the query image or as an intermediate step towards obtaining a more accurate pose estimate.
Most localization systems simply use state-of-the-art image representations trained for landmark retrieval or place recognition. 
In this paper, we analyzed the correlation between the tasks of visual localization and retrieval/recognition through detailed experiments using four global feature types.
We designed a benchmark framework that uses several definitions of retrieval ground truth to evaluate correlation between these tasks.
Furthermore, we generated subsets of images that contain interesting challenges for visual localization, low frequency regions (\eg blur) and dynamic regions (\eg persons and vehicles), and evaluated multiple global feature types on them.

Our results show that state-of-the-art image-level descriptors for place recognition are a good choice when localizing an image against a pre-built map as performance on both tasks is correlated. 
We also show that the visual localization \task~of pose approximation is directly correlated with landmark retrieval, however only NetVLAD improves the top 1 position by interpolation of a few images, \ie a low $k$ (performance drops when considering more images).
Still, representations that reflect pose similarity in their descriptors are preferable as they tend to retrieve closer images.
This is why DenseVLAD performs well for pose approximation but worse for local SFM.
When estimating pose accuracy without a pre-built map (local SFM), high landmark retrieval performance is preferable, but we do not see a clear correlation with landmark retrieval or place recognition metrics.

We can see that on the night images, AP-GeM and DELG-GLDv2 often outperform the others. 
One of the reasons might be that they were trained not only with geometric data augmentation but also with color jittering.
Since both were trained with outdoor datasets, there is room for improvement for indoor scenes.

By introducing visual localization specific image retrieval ground truth definitions, we have shown upper bounds for the three \tasks.
Altogether, using co-observations of 3D points and relative camera position between query and database images are a good choice for this since they best reflect the requirements for visual localization. 
The gap between the localization performance achieved with the ground truth rankings and actual features show that there is considerable room for improvement. 
In particular, the results indicate that better features might be trained by learning to replicate the ground truth rankings. 

Particular real-world challenges of visual localization, such as motion blur or large presence of dynamic objects, reduce the performance which shows that more representations robust to such challenges are needed.

In summary, our results suggest that developing suitable representations tailored to visual localization \tasks \, are interesting directions for future work, especially when considering the fact that visual localization performance can be significantly improved by proper image retrieval, as shown with the upper bounds.
To support such research, our code and evaluation protocols are publicly available at \url{https://github.com/naver/kapture-localization}.

\PAR{Acknowledgements} This work received funding through the EU Horizon 2020 research and innovation programme under grant agreement No.~857306 (RICAIP) and the European Regional Development Fund under IMPACT No.~CZ.02.1.01/0.0/0.0/15\_003/0000468.

\newpage

\bibliographystyle{spmpsci}      

\bibliography{IJCVbib}

\end{document}

%% file: IJCV_plots/table_GT.tex
\begin{tabular}{|c|l|c|c|}
\hline
\rowcolor[HTML]{EFEFEF} 
Dataset                                                      & Ground truth method & avg.~k & \multicolumn{1}{c|}{\cellcolor[HTML]{EFEFEF}missing (\%)} \\ \hline
\cellcolor[HTML]{EFEFEF} & co-observations         & 16 & 0.17 \\ \cline{2-4} 
\cellcolor[HTML]{EFEFEF} & distance\_0m-25m\_45deg & 30 & 0.03 \\ \cline{2-4} 
\cellcolor[HTML]{EFEFEF} & frustum-overlap         & 50 & 0.00 \\ \cline{2-4} 
\multirow{-4}{*}{\cellcolor[HTML]{EFEFEF}Baidu-mall}         & frustum-overlap-25m & 48    & 0.00                                                      \\ \hline
\cellcolor[HTML]{EFEFEF} & co-observations         & 48 & 0.01 \\ \cline{2-4} 
\cellcolor[HTML]{EFEFEF} & distance\_0m-25m\_45deg & 48 & 0.36 \\ \cline{2-4} 
\cellcolor[HTML]{EFEFEF} & frustum-overlap         & 50 & 0.00 \\ \cline{2-4} 
\multirow{-4}{*}{\cellcolor[HTML]{EFEFEF}RobotCar}           & frustum-overlap-25m & 50    & 0.00                                                      \\ \hline
\cellcolor[HTML]{EFEFEF} & co-observations         & 25 & 0.78 \\ \cline{2-4} 
\cellcolor[HTML]{EFEFEF} & distance\_0m-25m\_45deg & 50 & 0.00 \\ \cline{2-4} 
\cellcolor[HTML]{EFEFEF} & frustum-overlap         & 50 & 0.00 \\ \cline{2-4} 
\multirow{-4}{*}{\cellcolor[HTML]{EFEFEF}GangnamStation\_B2} & frustum-overlap-25m & 50    & 0.00                                                      \\ \hline
\end{tabular}

%% file: IJCV_plots/table_roxrpar.tex
\begin{tabular}{|l|c|c|c|c|}
\hline
 & $\mathcal{R}$O(m) &  $\mathcal{R}$O (h) & $\mathcal{R}$P(m) & $\mathcal{R}$P(h)  \\
\hline
DenseVLAD~\cite{ToriiPAMI18247PlaceRecognitionViewSynthesis} & 36.8 & 13.0 & 42.5 & 13.7  \\
NetVLAD~\cite{ArandjelovicCVPR16NetVLADPlaceRecognition} & 37.1 & 13.8  & 59.8 & 35.0  \\
AP-GeM~\cite{RevaudICCV19LearningwithAPTrainingImgRetrievalListwiseLoss} & 67.4 & 42.8 & 80.4 & 61.0 \\
DELG-GLDv2~\cite{CaoECCV20UnifyingDeepLocalGlobalFeatures} & 76.3 & 55.6 & 86.6 & 72.4 \\
\hline
\end{tabular}

%% file: IJCV_plots/table_precision_pearson.tex
\begin{tabular}{|l|c|c|c|c|}
\hline
\rowcolor[HTML]{EFEFEF} 
\multicolumn{1}{|c|}{\cellcolor[HTML]{EFEFEF}}                                                                                                            & \multicolumn{4}{c|}{\cellcolor[HTML]{EFEFEF}Pearson correlation coefficient} \\ \cline{2-5} 
\rowcolor[HTML]{EFEFEF} 
\multicolumn{1}{|c|}{\cellcolor[HTML]{EFEFEF}}                                                                                                            & \multicolumn{4}{c|}{\cellcolor[HTML]{EFEFEF}Baidu-mall}                      \\ \cline{2-5} 
\rowcolor[HTML]{EFEFEF} 
\multicolumn{1}{|c|}{\multirow{-3}{*}{\cellcolor[HTML]{EFEFEF}\begin{tabular}[c]{@{}c@{}}Landmark retrieval\\ precision vs. localized (\%)\end{tabular}}} & AP-GeM          & NetVLAD          & DenseVLAD          & DELG-GLDv2         \\ \hline
BDI (5.0m, 10.0°)                                                                                                                                         & \nc{0.99}            & \nc{0.98}             & \nc{0.99}               & \nc{0.99}               \\ \hline
CSI (5.0m, 10.0°)                                                                                                                                         & \nc{0.98}            & \nc{0.99}             & \nc{1.00}               & 0.99               \\ \hline
EWB (5.0m, 10.0°)                                                                                                                                         & \nc{0.98}            & \nc{0.99}             & \nc{0.99}               & \nc{0.99}               \\ \hline
Local SFM (0.25m, 2.0°)                                                                                                                                   & \nc{-0.99}           & \nc{-0.99}            & \nc{-1.00}              & \nc{-0.97}              \\ \hline
Global SFM (0.25m, 2.0°)                                                                                                                                  & \nc{-0.88}           & \nc{-0.93}            & \nc{-0.94}              & \nc{-0.91}              \\ \hline
\rowcolor[HTML]{EFEFEF} 
                                                                                                                                                          & \multicolumn{4}{c|}{\cellcolor[HTML]{EFEFEF}GangnamStation\_B2}              \\ \hline
BDI (5.0m, 10.0°)                                                                                                                                         & \nc{0.99}            & \nc{0.92}             & \cc{0.97}               & \cc{0.98}               \\ \hline
CSI (5.0m, 10.0°)                                                                                                                                         & \nc{0.99}            & \nc{0.92}             & \nc{0.99}               & \nc{0.99}               \\ \hline
EWB (5.0m, 10.0°)                                                                                                                                         & \nc{0.99}            & \nc{0.93}             & \cc{0.98}               & \cc{0.98}               \\ \hline
Local SFM (0.25m, 2.0°)                                                                                                                                   & \nc{-1.00}           & \nc{-1.00}            & \cc{-1.00}              & \nc{-1.00}              \\ \hline
Global SFM (0.25m, 2.0°)                                                                                                                                  & -0.99           & \cc{-0.99}            & \cc{-1.00}              & \cc{-0.99}              \\ \hline
\rowcolor[HTML]{EFEFEF} 
                                                                                                                                                          & \multicolumn{4}{c|}{\cellcolor[HTML]{EFEFEF}RobotCar}                        \\ \hline
BDI (5.0m, 10.0°)                                                                                                                                         & \cc{1.00}            & \nc{1.00}             & \nc{1.00}               & \nc{1.00}               \\ \hline
CSI (5.0m, 10.0°)                                                                                                                                         & \cc{1.00}            & \nc{0.99}             & \nc{0.98}               & \nc{1.00}               \\ \hline
EWB (5.0m, 10.0°)                                                                                                                                         & \nc{1.00}            & \nc{1.00}             & \nc{1.00}               & \nc{1.00}               \\ \hline
Local SFM (0.25m, 2.0°)                                                                                                                                   & \ccc{-0.91}           & \nc{-0.95}            & \nc{-0.95}              & \cc{-0.89}              \\ \hline
Global SFM (0.25m, 2.0°)                                                                                                                                  & \cc{-0.68}           & \cc{-0.72}            & \cc{-0.80}              & \cc{-0.64}              \\ \hline
\end{tabular}

%% file: IJCV_plots/table_precision_spearman.tex
\begin{tabular}{|l|c|c|c|c|}
\hline
\rowcolor[HTML]{EFEFEF} 
\multicolumn{1}{|c|}{\cellcolor[HTML]{EFEFEF}}                                                                                                            & \multicolumn{4}{c|}{\cellcolor[HTML]{EFEFEF}Spearman correlation coefficient} \\ \cline{2-5} 
\rowcolor[HTML]{EFEFEF} 
\multicolumn{1}{|c|}{\cellcolor[HTML]{EFEFEF}}                                                                                                            & \multicolumn{4}{c|}{\cellcolor[HTML]{EFEFEF}Baidu-mall}                       \\ \cline{2-5} 
\rowcolor[HTML]{EFEFEF} 
\multicolumn{1}{|c|}{\multirow{-3}{*}{\cellcolor[HTML]{EFEFEF}\begin{tabular}[c]{@{}c@{}}Landmark retrieval\\ precision vs. localized (\%)\end{tabular}}} & 1                 & 5                 & 10                & 20                \\ \hline
BDI (5.0m, 10.0°)                                                                                                                                         & \nc{-1.00}             & \nc{-0.20}             & \nc{0.60}              & \nc{0.80}              \\ \hline
CSI (5.0m, 10.0°)                                                                                                                                         & \nc{-1.00}             & \nc{-0.80}             & \nc{-1.00}             & \nc{-1.00}             \\ \hline
EWB (5.0m, 10.0°)                                                                                                                                         & \nc{-1.00}             & \nc{-0.20}             & \nc{0.40}              & \nc{-0.80}             \\ \hline
Local SFM (0.25m, 2.0°)                                                                                                                                   &                   & \nc{0.20}              & \nc{0.80}              & \nc{-0.20}             \\ \hline
Global SFM (0.25m, 2.0°)                                                                                                                                  & \cc{-0.32}             & \nc{-0.40}             & \nc{-0.80}             & \cc{-0.63}             \\ \hline
\rowcolor[HTML]{EFEFEF} 
                                                                                                                                                          & \multicolumn{4}{c|}{\cellcolor[HTML]{EFEFEF}GangnamStation\_B2}               \\ \hline
BDI (5.0m, 10.0°)                                                                                                                                         & \nc{0.40}              & \nc{1.00}              & \nc{0.40}              & \nc{0.40}              \\ \hline
CSI (5.0m, 10.0°)                                                                                                                                         & \nc{0.40}              & \nc{0.80}              & \nc{1.00}              & \nc{0.80}              \\ \hline
EWB (5.0m, 10.0°)                                                                                                                                         & \nc{0.40}              & \cc{0.95}              & \nc{0.80}              & \nc{0.20}              \\ \hline
Local SFM (0.25m, 2.0°)                                                                                                                                   &                   & \nc{1.00}              & \nc{-0.20}             & \nc{-1.00}             \\ \hline
Global SFM (0.25m, 2.0°)                                                                                                                                  & \nc{1.00}              & \nc{0.80}              & \cc{0.63}            & \nc{0.40}              \\ \hline
\rowcolor[HTML]{EFEFEF} 
                                                                                                                                                          & \multicolumn{4}{c|}{\cellcolor[HTML]{EFEFEF}RobotCar}                         \\ \hline
BDI (5.0m, 10.0°)                                                                                                                                         & \nc{-0.40}             & \nc{0.80}              & \nc{0.80}              & \nc{1.00}              \\ \hline
CSI (5.0m, 10.0°)                                                                                                                                         & \nc{-0.40}             & \nc{0.80}              & \nc{0.80}              & \nc{0.63}              \\ \hline
EWB (5.0m, 10.0°)                                                                                                                                         & \nc{-0.40}             & \nc{0.80}              & \nc{0.80}              & \nc{1.00}              \\ \hline
Local SFM (0.25m, 2.0°)                                                                                                                                   &                   & \nc{0.80}              & \cc{0.74}              & \nc{0.60}              \\ \hline
Global SFM (0.25m, 2.0°)                                                                                                                                  & \nc{0.00}              & \nc{0.60}              & \nc{0.60}             & \nc{0.60}              \\ \hline
\end{tabular}

%% file: IJCV_plots/table_recall_spearman.tex
\begin{tabular}{|l|c|c|c|c|}
\hline
\rowcolor[HTML]{EFEFEF} 
\multicolumn{1}{|c|}{\cellcolor[HTML]{EFEFEF}}                                                                                                        & \multicolumn{4}{c|}{\cellcolor[HTML]{EFEFEF}Spearman correlation coefficient} \\ \cline{2-5} 
\rowcolor[HTML]{EFEFEF} 
\multicolumn{1}{|c|}{\cellcolor[HTML]{EFEFEF}}                                                                                                        & \multicolumn{4}{c|}{\cellcolor[HTML]{EFEFEF}Baidu-mall}                       \\ \cline{2-5} 
\rowcolor[HTML]{EFEFEF} 
\multicolumn{1}{|c|}{\multirow{-3}{*}{\cellcolor[HTML]{EFEFEF}\begin{tabular}[c]{@{}c@{}}Place recognition\\ recall vs. localized (\%)\end{tabular}}} & 1                 & 5                 & 10                & 20                \\ \hline
BDI (5.0m, 10.0°)                                                                                                                                     & \nc{-1.00}             & \nc{-0.20}             & \nc{0.60}              & \nc{-0.40}             \\ \hline
CSI (5.0m, 10.0°)                                                                                                                                     & \nc{-1.00}             & \nc{-0.80}             & \nc{-1.00}             & \nc{-0.20}             \\ \hline
EWB (5.0m, 10.0°)                                                                                                                                     & \nc{-1.00}             & \nc{-0.20}             & \nc{0.40}              & \nc{-0.40}             \\ \hline
Local SFM (0.25m, 2.0°)                                                                                                                               &                   & \nc{0.20}              & \nc{0.80}              & \nc{0.60}              \\ \hline
Global SFM (0.25m, 2.0°)                                                                                                                              & \cc{-0.32}             & \nc{-0.40}             & \nc{-0.80}             & \cc{-0.21}             \\ \hline
\rowcolor[HTML]{EFEFEF} 
                                                                                                                                                      & \multicolumn{4}{c|}{\cellcolor[HTML]{EFEFEF}GangnamStation\_B2}               \\ \hline
BDI (5.0m, 10.0°)                                                                                                                                     & \nc{0.40}              & \nc{1.00}              & \nc{0.80}              & \nc{0.20}              \\ \hline
CSI (5.0m, 10.0°)                                                                                                                                     & \nc{0.40}              & \nc{0.80}              & \nc{0.80}              & \nc{1.00}              \\ \hline
EWB (5.0m, 10.0°)                                                                                                                                     & \nc{0.40}             & \cc{0.95}              & \nc{0.40}              & \nc{0.40}              \\ \hline
Local SFM (0.25m, 2.0°)                                                                                                                               &                   & \nc{1.00}              & \nc{-0.40}             & \nc{-0.80}             \\ \hline
Global SFM (0.25m, 2.0°)                                                                                                                              & \nc{1.00}              & \nc{0.80}              & \cc{0.32}              & \nc{0.00}              \\ \hline
\rowcolor[HTML]{EFEFEF} 
                                                                                                                                                      & \multicolumn{4}{c|}{\cellcolor[HTML]{EFEFEF}RobotCar}                         \\ \hline
BDI (5.0m, 10.0°)                                                                                                                                     & \nc{-0.40}             & \nc{0.80}              & \nc{0.80}              & \nc{0.60}              \\ \hline
CSI (5.0m, 10.0°)                                                                                                                                     & \nc{-0.40}             & \nc{0.80}              & \nc{0.80}              & \cc{0.63}              \\ \hline
EWB (5.0m, 10.0°)                                                                                                                                     & \nc{-0.40}             & \nc{0.80}              & \nc{0.80}              & \nc{0.60}             \\ \hline
Local SFM (0.25m, 2.0°)                                                                                                                               &                   & \nc{0.80}              & \cc{0.95}              & \nc{1.00}              \\ \hline
Global SFM (0.25m, 2.0°)                                                                                                                              & \nc{0.00}              & \nc{1.00}              & \nc{1.00}              & \nc{1.00}              \\ \hline
\end{tabular}

%% file: IJCV_plots/table_recall_pearson.tex
\begin{tabular}{|l|
>{\columncolor[HTML]{FFFFFF}}c |
>{\columncolor[HTML]{FFFFFF}}c |
>{\columncolor[HTML]{FFFFFF}}c |
>{\columncolor[HTML]{FFFFFF}}c |}
\hline
\multicolumn{1}{|c|}{\cellcolor[HTML]{EFEFEF}}                                                                                                        & \multicolumn{4}{c|}{\cellcolor[HTML]{EFEFEF}Pearson correlation coefficient}                                                              \\ \cline{2-5} 
\multicolumn{1}{|c|}{\cellcolor[HTML]{EFEFEF}}                                                                                                        & \multicolumn{4}{c|}{\cellcolor[HTML]{EFEFEF}Baidu-mall}                                                                                   \\ \cline{2-5} 
\multicolumn{1}{|c|}{\multirow{-3}{*}{\cellcolor[HTML]{EFEFEF}\begin{tabular}[c]{@{}c@{}}Place recognition\\ recall vs. localized (\%)\end{tabular}}} & \cellcolor[HTML]{EFEFEF}AP-GeM & \cellcolor[HTML]{EFEFEF}NetVLAD & \cellcolor[HTML]{EFEFEF}DenseVLAD & \cellcolor[HTML]{EFEFEF}DELG-GLDv2 \\ \hline
BDI (5.0m, 10.0°)                                                                                                                                     & \nc{-0.87}                          & \nc{-0.88}                           & \nc{-0.92}                             & \nc{-0.87}                              \\ \hline
CSI (5.0m, 10.0°)                                                                                                                                     & \nc{-0.92}                          & \nc{-0.91}                          & \nc{-0.93}                             & \nc{-0.90}                              \\ \hline
EWB (5.0m, 10.0°)                                                                                                                                     & \nc{-0.93}                          & \nc{-0.91}                           & \nc{-0.94}                             & \nc{-0.90}                              \\ \hline
Local SFM (0.25m, 2.0°)                                                                                                                               & \nc{1.00}                          & \nc{1.00}                            & \nc{1.00}                              & \nc{1.00}                               \\ \hline
Global SFM (0.25m, 2.0°)                                                                                                                              & \nc{0.99}                           & \nc{1.00}                            & \nc{1.00}                              & \nc{1.00}                               \\ \hline
\cellcolor[HTML]{EFEFEF}                                                                                                                              & \multicolumn{4}{c|}{\cellcolor[HTML]{EFEFEF}GangnamStation\_B2}                                                                           \\ \hline
BDI (5.0m, 10.0°)                                                                                                                                     & \nc{-0.95}                          & \nc{-0.85}                           & \cc{-0.96}                             & \nc{-0.97}                             \\ \hline
CSI (5.0m, 10.0°)                                                                                                                                     & \nc{-0.95}                          & \cc{-0.85}                           & \cc{-0.97}                             & \cc{-0.98}                              \\ \hline
EWB (5.0m, 10.0°)                                                                                                                                     & \cc{-0.95}                          & \nc{-0.88}                           & \cc{-0.97}                             & \nc{-0.98}                              \\ \hline
Local SFM (0.25m, 2.0°)                                                                                                                               & \cc{0.99}                           & \nc{1.00}                            & \cc{0.97}                              & \nc{1.00}                            \\ \hline
Global SFM (0.25m, 2.0°)                                                                                                                              & \nc{0.99}                           & \nc{0.99}                            & \cc{0.98}                              & \nc{0.99}                               \\ \hline
\cellcolor[HTML]{EFEFEF}                                                                                                                              & \multicolumn{4}{c|}{\cellcolor[HTML]{EFEFEF}RobotCar}                                                                                     \\ \hline
BDI (5.0m, 10.0°)                                                                                                                                     & \cc{-0.83}                          & \cc{-0.87}                           & \cc{-0.90}                             & \nc{-0.90}                             \\ \hline
CSI (5.0m, 10.0°)                                                                                                                                     & \cc{-0.85}                          & \cc{-0.91}                           & \cc{-0.94}                             & \nc{-0.88}                              \\ \hline
EWB (5.0m, 10.0°)                                                                                                                                     & \cc{-0.81}                          & \cc{-0.86}                           & \nc{-0.90}                             & \nc{-0.87}                              \\ \hline
Local SFM (0.25m, 2.0°)                                                                                                                               & \ccc{0.99}                           & 1.00                            & 1.00                              & \cc{0.97}                               \\ \hline
Global SFM (0.25m, 2.0°)                                                                                                                              & \nc{1.00}                          & \cc{0.98}                            & \cc{0.99}                              & \cc{0.97}                               \\ \hline
\end{tabular}